\documentclass[final,5p,times,twocolumn, sort&compress]{elsarticle}

\usepackage{amsmath}
\usepackage{moreverb,url}
\usepackage{color, colortbl}
\usepackage[table]{xcolor}
\usepackage{multirow}

\definecolor{Gray}{gray}{0.9}
\definecolor{Gray2}{rgb}{0.1,0.1,0.1}
\newcommand{\myParagraph}[1]{\vspace{1mm}\noindent\textbf{#1}}
\usepackage{siunitx}
\DeclareMathOperator*{\argmin}{arg\,min}
\usepackage[colorlinks,bookmarksopen,bookmarksnumbered,citecolor=red,urlcolor=red]{hyperref}

\newcommand\BibTeX{{\rmfamily B\kern-.05em \textsc{i\kern-.025em b}\kern-.08em
T\kern-.1667em\lower.7ex\hbox{E}\kern-.125emX}}

\usepackage{eso-pic}
\usepackage{transparent}
 
\AddToShipoutPictureBG{
  \AtPageLowerLeft{%
    \raisebox{20pt}{\makebox[\paperwidth]{\begin{minipage}{21cm}\centering \transparent{0.4}
      \centering Author version of the manuscript published as: 
      Stepanova, K.; Rozlivek, J.; Puciow, F.; Krsek, P.; Pajdla, T. \& Hoffmann, M. (2022),
       'Automatic self-contained calibration of an industrial dual-arm robot with cameras using self-contact, planar constraints, and self-observation', Robotics and Computer-Integrated Manufacturing 73, 102550. (C) Elsevier. OPEN ACCESS.
    \end{minipage}}}%
  }
}

\begin{document}

\begin{frontmatter}

\title{Automatic self-contained calibration of an industrial dual-arm robot with cameras using self-contact, planar constraints, and self-observation}

\author[1,2,*]{Karla Stepanova}
\author[1,*]{Jakub Rozlivek}
\author[1]{Frantisek Puciow}
\author[2]{Pavel Krsek}
\author[2]{Tomas Pajdla}
\author[1]{Matej Hoffmann\texorpdfstring{\corref{cor1}}{}}\ead{matej.hoffmann@fel.cvut.cz} 

\address[1]{Department of Cybernetics, Faculty of Electrical Engineering, Czech Technical University in Prague, Czech Republic}
\address[2]{Czech Institute of Informatics, Robotics, and Cybernetics, Czech Technical University in Prague, Czech Republic}
\address[*]{Both authors contributed equally.}
\cortext[cor1]{Corresponding author}

\begin{abstract}
We present a robot kinematic calibration method that combines complementary calibration approaches: self-contact, planar constraints, and self-observation. We analyze the estimation of the end effector parameters, joint offsets of the manipulators, and calibration of the complete kinematic chain (DH parameters). The results are compared with ground truth measurements provided by a laser tracker.  Our main findings are: (1) When applying the complementary calibration approaches in isolation, the self-contact approach yields the best and most stable results. (2) All combinations of more than one approach were always superior to using any single approach in terms of calibration errors and the observability of the estimated parameters. Combining more approaches delivers robot parameters that better generalize to the workspace parts not used for the calibration. (3) Sequential calibration, i.e.\ calibrating cameras first and then robot kinematics, is more effective than simultaneous calibration of all parameters. In real experiments, we employ two industrial manipulators mounted on a common base. The manipulators are equipped with force/torque sensors at their wrists, with two cameras attached to the robot base, and with special end effectors with fiducial markers. We collect a new comprehensive dataset for robot kinematic calibration and make it publicly available. The dataset and its analysis provide quantitative and qualitative insights that go beyond the specific manipulators used in this work and  apply to self-contained robot kinematic calibration in general. 
\end{abstract}


\end{frontmatter}

\section{Introduction}
Accurate calibration is essential for the performance of every robot. Traditional calibration procedures involve some form of external measuring apparatus and become impractical if the robot itself or the site where it is deployed change frequently---the current trend in automation with the shift from mass to small batch production. Furthermore, collaborative and social robots often employ cheaper and more elastic materials, making them less accurate and prone to frequent re-calibration. At the same time, advances in sensor technology make affordable but increasingly accurate devices such as standard or RGB-D cameras, tactile, force, or inertial sensors available; often, robots come already equipped with a selection of them. These factors together constitute the need and the opportunity to perform an automated self-contained calibration relying on redundant information in these sensory streams originating from the robot platform itself. 

\begin{figure}
\centering
\includegraphics[width=0.23\textwidth]{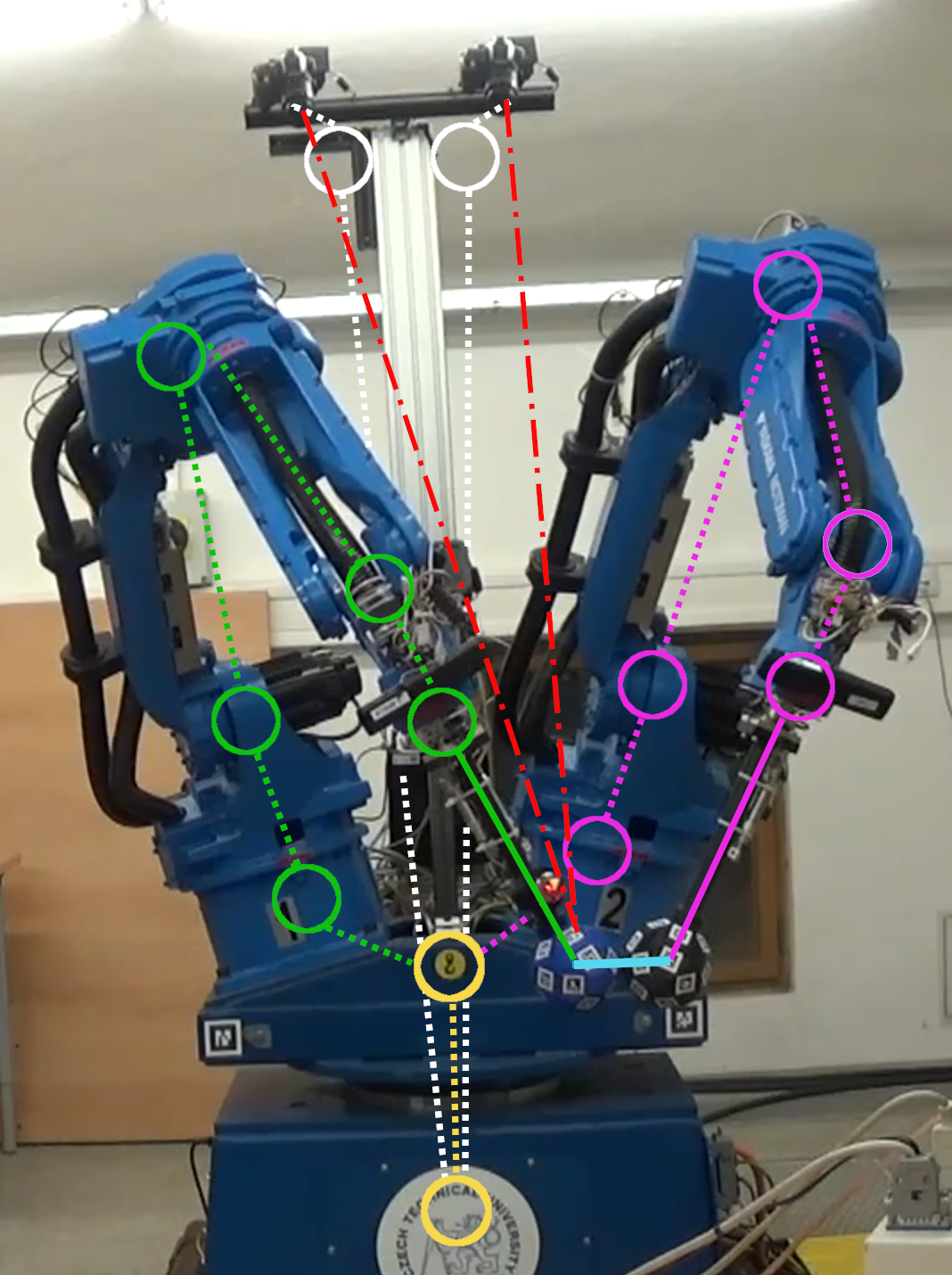}
\includegraphics[width=0.23\textwidth]{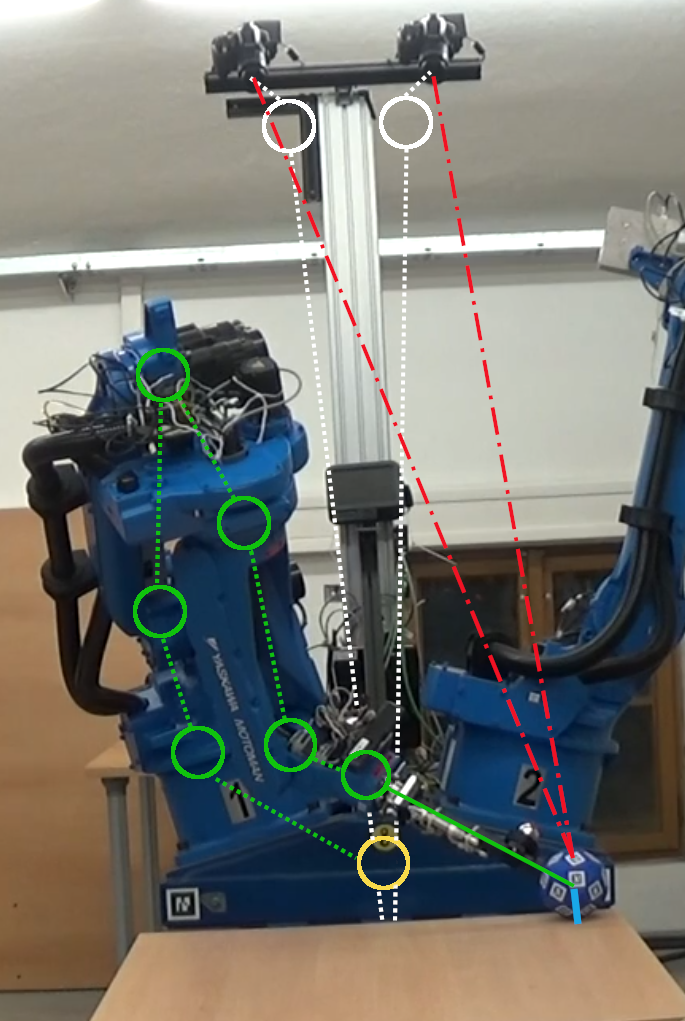}
\caption{\label{fig:self_touch_setup}Setups for automatic calibration of the dual-arm robot with graphs of the kinematic chains. Setup for self-touch calibration (left) and calibration when a horizontal plane is touched (right). All chains originate in a common base frame (bottom yellow circle). The left and right arm chains are drawn in purple and green, respectively. The eye chains are drawn in white. Red lines denote reprojection into the cameras. Cyan indicates the distance between end effector centers (one diameter) or between the end effector and a plane (end effector radius). }
\end{figure}

We present a robot kinematic calibration method, which combines complementary calibration approaches: self-contact, planar constraints, and self-observation. There are two main approaches to robot calibration \cite{Hollerbach2016}: (i) \textit{open-loop} approaches, where the manipulator is not in mechanical (physical) contact with the environment and an external metrology system is used to measure robot pose components, and (ii) \textit{closed-loop} approaches, where physical constraints on robot pose components substitute for external measurements. However, in their standard formulations, both approaches require a setup that is external to the robot itself. This work aims to extend these well-established tools to \textit{automatic self-contained robot calibration} \cite{Birbach2015} to include camera reprojection errors, constraints arising from robot self-contact, or simultaneous calibration of multiple kinematic chains.

In this work, we employ two industrial manipulators mounted on a common base with force/torque sensors at their wrists, two cameras overlooking the workspace, and special end effectors with fiducial markers. Using this setup, we study different kinematic calibration methods: self-contact, planar constraints, and self-observation. These can be employed separately or in combination (simultaneously or sequentially). Fig.~\ref{fig:self_touch_setup} shows an example: a self-contact configuration provides constraints that can be exploited to calibrate the kinematic chain of one or both arms. At the same time, the end effectors are observed by the cameras, providing additional data (equations) for calibrating the manipulators' kinematics and the camera orientation parameters. 

We evaluate the performance of individual approaches on the calibration of: (i) parameters of the custom end effector, (ii) joint offsets of the complete kinematic chain of one arm, and (iii) the full Denavit-Hartenberg (DH) representation of the platform. Calibration using an external measurement device (laser tracker) is performed and serves together with the nominal manipulators' kinematic parameters as a reference. The calibration performance of individual methods is compared on an independent testing dataset which covers a significant portion of the robot workspace.

We present four main contributions. First, we provide a thorough experimental comparison of geometric kinematic calibration using self-contained approaches---self-contact, planar constraints, and self-observation---with kinematic calibration based on an external laser tracker. We demonstrate the viability of the self-contained approaches. Second, we design ways how the individual methods can be combined into a single cost function. The merits of such a synergistic approach leading to better identifiability of the calibrated parameters and generalization to the part of the robot workspace, which was not used for parameter calibration, are empirically evaluated. Third, we compare simultaneous (all parameters at once) with sequential calibration. Fourth, we collect a new comprehensive dataset for robot kinematic calibration and make it publicly available. This dataset and its analysis provide quantitative and qualitative insights that go beyond the specific manipulators used in this work and apply to self-contained robot kinematic calibration in general.

To our knowledge, this work is the first that bridges the realm of very accurate industrial calibration using external metrology with compact approaches relying on sensors on the robot---typical for humanoid or social robotics. 
The innovation this work brings is that it expands the portfolio of traditional calibration methods by self-contained approaches such as self-contact or self-observation. While traditionally industrial robots did not have means to detect contacts, this is changing now with the advent of collaborative robots where sensitive contact detection is necessary for safe human-robot collaboration. Furthermore, torque sensing within the robot structure or joint/torque sensors at the flange facilitate dexterous manipulation and interaction with the environment. At the same time, integrated vision sensors (cameras, RGB-D sensors) are becoming popular. Here we provide a framework that makes it possible to exploit all available calibration methods and sensory streams simultaneously and combine them in an optimal way, including weighting by their respective accuracy.

This article is structured as follows. Related  work is reviewed in Sec.~\ref{section:RW}.  Section~\ref{sec:setup} describes Experimental setup and data acquisition: Robot  (Sec.~\ref{subsec:robot_setup_description}) and Laser tracker (Sec.~\ref{subsec:laser_setup}) setup, robot control using force feedback (Sec.~\ref{subsec:robot_control}), robot and end effector dimensions (Sec.~\ref{subsec:robot_dimensions}), camera calibration (Sec.~\ref{section:camera_calib}), and description of individual acquired datasets (Sec.~\ref{section:dataset_description}). The optimization problem formulation for different combinations of kinematic chains is described in Sec.~\ref{sec:multichain}. Experimental results for end effector, manipulator offsets, and all DH parameters calibration (including laser tracker reference) are presented in Sec.~\ref{section:results}. Finally, we present the conclusion (Sec.~\ref{section:conclusion}) and discussion, including future work (Sec.~\ref{section:discussion}). The accompanying video illustrating the experiments is available at \url{https://youtu.be/LMwINqA1t9w}; the dataset is available from~\cite{dataset_our}.

\section{Related work}
\label{section:RW}

Standard robot kinematic calibration employs different mechanical systems---for example, measurement arms \cite{ginani2011theoretical}---or contactless measurement systems like laser trackers \cite{ha2008kinematic,nguyen2013new,Newman2000, nubiola2013absolute}. Cameras are used less often: for example, stereo cameras combined with a calibration sphere \cite{svaco2014calibration} or a camera together with a laser pointer at the end effector \cite{hu2012kinematic}. All of these systems require specialized external equipment.
Our platform was previously calibrated using two different methods: (1) Redundant parallel Calibration and measuring Machine (RedCaM) by Bene{\v{s}} et al.~\cite{benevs2007experiments}, Volech et al.~\cite{volech2013concepts}, and (2) Leica laser tracker. Petr{\'\i}k and Smutn{\'y} \cite{petrik_2014_comparison} reviewed the precision of these methods using a linear probe sensor. 

Since we aim at automatic self-contained multisensorial calibration---that is, calibration using sensors on the robot, involving multiple sensory modalities like vision and touch---we next focus on reviewing the state of the art in calibration methods that do not rely on external metrology systems. We will also pay special attention to humanoid-like setups that offer the richest possibilities for self-contained calibration. 

{\em Calibration by self-observation.}~Cameras mounted on a robot can be used to calibrate the robot by closing the calibration loop through self-observation of its end effectors. The theory for this approach is laid out in \cite{bennett1991autonomous} for a stereo camera system observing a robot arm. The manipulator's kinematics and extrinsic and intrinsic camera parameters are calibrated. Self-observation has been applied to humanoid robots viewing their hands with fiducial markers using online methods to calibrate kinematics relying on gradient descent by Hersch et al.~\cite{Hersch2008} and recursive least squares estimation by Martinez-Cantin et al.~\cite{Martinez-Cantin2010}. Fiducial markers can sometimes be avoided when the robot wrist, hand, or fingertip are identified directly in the image \cite{Fanello2014,Vicente2016}.
The work of Birbach et al.~\cite{Birbach2015} on the humanoid Agile Justin will be discussed in more detail below.

{\em Calibration using physical constraints.}~The next family of approaches exploits physical contacts of the end effector with the environment, such as fixing the end effector to the ground \cite{Bennett1988} or using more complex setups \cite{Joubair2015,Koval2017,xu2018calibration}. Some form of force sensing on the part of the manipulator is required.  
Kinematic calibration using plane constraints (with known or unknown parameters of the plane) was explored by Ikits and Hollerbach~\cite{Ikits1997}; they proposed a new approach focusing on the proper definition of the base and end link frames and evaluated primarily in simulation. Zhuang et al.~\cite{Zhuang1999} explored multiple variants of plane constraints and the option with/without known plane parameters and demonstrated their results on a PUMA 560 robot. In particular, they showed that a single-plane constraint does not necessarily guarantee that all kinematic parameters of the robot will be observable. On the other hand, a multiple-plane constraint should be a remedy to this problem. They show that the data collected from 3 planar constraints is equivalent to the data collected from a point measurement device provided that: 1) all three planes are mutually non-parallel; 2) the identification Jacobian of the unconstrained system is nonsingular; and 3) the measured points from each individual plane do not lie on a line on that plane. Joubair and Bonev~\cite{Joubair2015b} showed how multi-planar constraints can significantly improve the accuracy of calibration of a six-axis serial robot. Zenha et al.~\cite{zenha2018incremental} had the simulated iCub humanoid touch three known planes, employing adaptive parameter estimation (Extended Kalman Filter) for kinematic calibration. Khusainov et al.~\cite{Khusainov2017} exploit a specific type of mechanical coupling as they fix the end effector of a manipulator to the legs of a humanoid robot that is being calibrated. A point and distance constraint can also be obtained from visual sensing \cite{wang2020point}.

{\em Calibration by self-contact (self-touch).}~Self-contact constitutes a specific, less common way of kinematic loop closure available only to humanoid-like or dual arm setups. Additionally, corresponding sensory and motor equipment such that this self-contact can be performed in a controlled manner is needed. One possibility is to utilize artificial electronic skins covering specific areas or complete robot bodies (see \cite{Bartolozzi2016, Dahiya2019} for recent overviews). 
A tactile array may be used for contact detection. If accurate spatial calibration of the skin is available, then additional components of the self-contact configuration---where contact occurs on each of the two intersecting chains---can be measured.
 Roncone et al.~\cite{Roncone_ICRA_2014} performed kinematic calibration on the iCub using autonomous self-touch---index finger on the contralateral forearm; Li et al.~\cite{QiangLi2015} employed a dual KUKA arm setup with a sensorized ``finger'' and a tactile array on the other manipulator.
Mittendorfer and Cheng~\cite{Mittendorfer2011} also exploit artificial skin to learn models of robots, but their methods primarily utilize the signals from accelerometers embedded in their multimodal skin. Alternatively, if controlled self-contact can be established but the exact position is not measured---typically when using force/torque sensing---such constraints can also be employed for calibration, as will be demonstrated in this work.

{\em Self-contained multisensorial calibration.} There are only a few approaches that exploit ``multisensorial'' (or ``multimodal'') input for self-contained calibration. Birbach et al.~\cite{Birbach2015} calibrated the humanoid robot Justin observing its wrist. Sensors were fused by minimizing a single cost function that aggregates the errors obtained by comparing the discrepancies between simulated projections (left and right camera images, Kinect image, Kinect disparity) and the wrist position from forward kinematics. An inertial term from an IMU in the head was also considered. It is claimed that while pair-wise calibration can lead to inconsistencies, calibrating everything together  in a ``mutually supportive way'' is the most efficient. Limoyo et al.~\cite{limoyo2018self} used contact constraints from sliding on a surface together with RGB-D camera information to formulate a self-calibration problem for a mobile manipulator to estimate camera extrinsic parameters and manipulator joint angle biases. The former part is also experimentally verified. 
Stepanova et al.~\cite{Stepanova2019} systematically studied on the simulated iCub humanoid robot how self-observation, self-contact, and their combination can be used for self-calibration and evaluated the relative performance of these approaches by varying the number of poses, initial parameter perturbation and measurement noise. They found that employing multiple kinematic chains (``self-observation'' and ``self-touch'') is superior in terms of optimization results as well as observability.

Largely orthogonal to the calibration types mentioned above, observability and identifiability of the system and speed of optimization convergence can be improved by i) combination of geometric and parametric approaches to kinematic identification~\cite{Boby2021}, (ii) improving the error model by incorporating impact of strain wave gearing errors~\cite{Jiang2021} and unifying various error sources~\cite{Fu2020}, or iii) improving the optimization method~\cite{Luo2021}. 

In summary, existing works on self-contained automatic calibration have typically focused on a single approach---relying on self-observation, physical constraints, or self-contact. Birbach et al.~\cite{Birbach2015} combined multiple sensors. 
However, first, essentially only ``self-observation'' chains closed through different vision-like sensors in the robot head were used. Second, only the case where all chains were combined using a single cost function was considered.
In \cite{limoyo2018self}, self-observation and contact information are combined, but the results have a proof-of-concept character. Our work is inspired by the simulation study of Stepanova et al.~\cite{Stepanova2019} but presents a completely new setting and results on a different platform. 
Using a dual-arm industrial robot with force/torque sensing and cameras, we present several formulations of the optimization problem for geometric kinematic calibration and empirical verification using (i) self-contact, (ii) planar constraints, and (iii) self-observation. The pros and cons of the individual methods and their synergies are assessed and compared to an independent calibration using an industrial quality laser tracker.

\section{Experimental setup and data acquisition}
\label{sec:setup}

In this section, we introduce the experimental setup: the robot platform and its control, dimensions, and camera calibration. Then we present the process of data acquisition and the structure of the collected datasets. 

\subsection{Robot setup description}
\label{subsec:robot_setup_description}
\begin{figure}[!h]
\centering
\begin{minipage}[c]{0.4\textwidth}
    \includegraphics[width=1\textwidth]{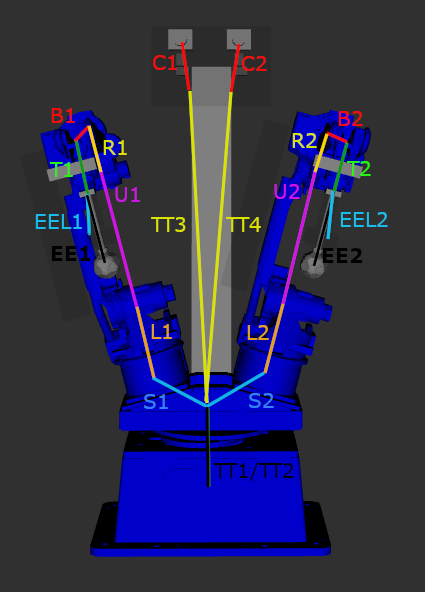}
  \end{minipage}\hfill
  \begin{minipage}[c]{0.5\textwidth}
    \caption{Setup of the whole dual arm robot: visualisation of individual links and their location on the robot. DH parameters of these links are listed in Table \ref{tab:merged_dh}. }
    \label{fig:setupMotoman}
  \end{minipage}
\end{figure}

For our experiments, we used a robotic platform developed in the CloPeMa project \cite{clopema_robot} -- Fig.~\ref{fig:self_touch_setup} and Fig.~\ref{fig:setupMotoman}. It consisted of two industrial manipulators Yaskawa-Motoman MA1400 installed on top of a Yaskawa R750 robotic turntable, which allows rotation of the two manipulators around the vertical axis, a control unit, and two computers connected to a local network. Two Nikon D3100 cameras with Nikkor 18-55 AF-S DX VR lens were mounted side-by-side on the turntable over the robot base, moving along with the turntable.

Manipulators were equipped with ATI Industrial Automation Mini45 6-axis force/torque (F/T) sensors, placed between the last link of the manipulator and the end effector. 
All the different parts of the robot system were integrated into and operated with Robot Operating System (ROS) \cite{ros}. MoveIt! planning framework was used to control the robot.

\begin{figure*}[!h]
\centering
\makebox[\textwidth]{
\includegraphics[width=0.4\textwidth]{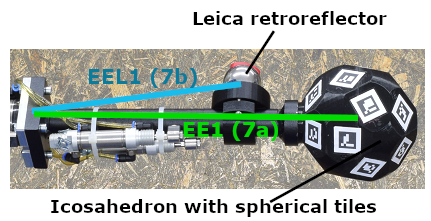}
\includegraphics[width=0.33\textwidth]{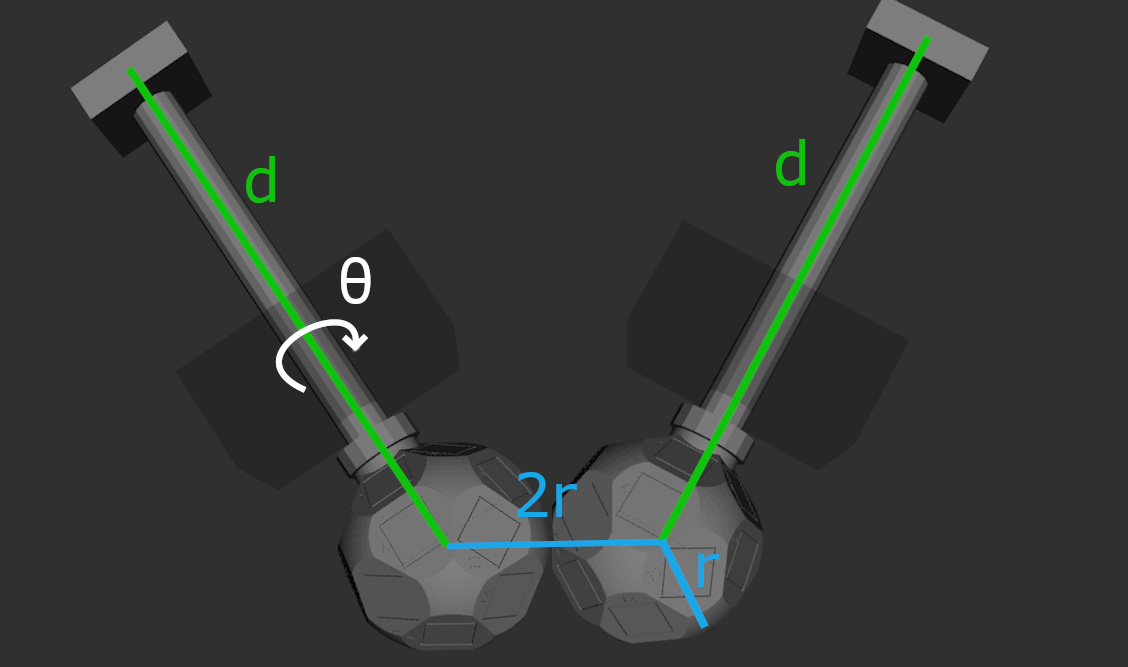}
\includegraphics[width=0.195\textwidth,angle=90]{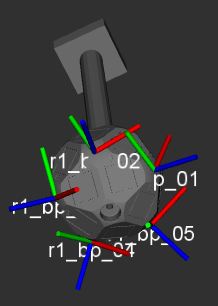}
}
\caption[End effector]{Custom end effector with icosahedron and laser tracker retroreflector (left), its collision model in self-contact configuration ($d$ and $o$ are DH parameters of end effector and $r$ is the radius of icosahedrons) (center), and spherical tiles used for contacts with their reference frames (right). \label{fig:ee}}
\end{figure*}

We further equipped the robot with custom-made end effectors, one on each robot arm,  that can be used to achieve self-contact and, at the same time, visual self-observation by the two cameras. To this end, we designed and 3D printed icosahedron shapes  (Figure~\ref{fig:ee} (left)) that have 20 flat faces, where fiducial markers can be placed, interleaved with 10 spherical surface regions that can be used for self-contact. The end effectors (further also referred to as icosahedron) were then attached to a steel rod. The collision model (Fig.~\ref{fig:ee}, center) for the end effector was added to the robot model. For calibration employing self-contact or contact with a plane, the end effectors are treated as spheres (see Fig.~\ref{fig:ee}, center). The spherical surfaces with corresponding reference frames are shown in Fig.~\ref{fig:ee}, right.

For visual self-observation, having 20 evenly spaced ArUco markers should ensure that at least 3 of them are always seen by each camera unless another part of the robot occludes the view.
We used the OpenCV ArUco module to detect the markers in images.

\subsection{Laser tracker setup}
\label{subsec:laser_setup}
For acquiring reference parameter values and a baseline for self-contained calibration, we used an external ``laser tracker'', i.e.\ Leica Geosystems Absolute Tracker AT402, which is a portable 3D absolute tracker for fully guided measurement processes \cite{Leica}. The tracker collects the 3D coordinates of points in the coordinate system of the tracker. 
The measurement has a resolution of 0.1 microns. The tracker was placed approximately 4~m from the manipulator. Thus, the typical $U_{xyz}$ uncertainty of the measurement was $\pm 7.5+3\times 4$ microns \cite{Leica_patent,Leica}.

The retroreflector was attached to the robot approximately 25 cm from the last joint. During data collection, the distance from the laser tracker was between 2.5 and 4.5 m. The retroreflector and its collision model are shown in Figure~\ref{fig:ee} (left and center).

\subsection{Robot control for contact configurations using force feedback}
\label{subsec:robot_control}
Driving the two manipulator arms into physical contact is necessary for exploiting self-contact in robot calibration. A point in the workspace was chosen where the end effectors should touch. Then, the configuration of each arm and the movement trajectory were obtained using MoveIt!. Each contact consists of three or four phases depending on the experiment. In the first phase, the robot right arm moves at a speed of $0.7\si{\meter\per\second}$ to a point close to the desired pose. In the self-contact experiment, this is followed by an analogous movement of the left arm. Then the right arm starts moving to an anticipated contact point (in fact, a small negative distance was used) with the left arm or a plane at a speed of $0.1\si{\meter\per\second}$ until the collision is detected by F/T sensors.  The contact thresholds were determined empirically. 
Once the end effector gets into contact, the arm stops moving and cameras are triggered to take photographs. Joint angles are recorded. Afterwards, the right arm is slowly ($0.1\si{\meter\per\second}$) moving to the former position (from the end of the first phase). Then the new target is selected and the whole process is repeated. For laser tracker experiments, the deceleration phase is skipped and the robot arm chosen moves directly at a speed of $0.7\si{\meter\per\second}$ in free space to predefined configurations which were sampled in a way that a significant part of the robot joint space is covered.

\subsection{Robot and end effector dimensions and representation}
\label{subsec:robot_dimensions}
A parametric model of the whole robot including the custom-made end effectors was created. The robot's complete kinematic model was described by Denavit-Hartenberg (DH) notation, including cameras, end effectors, and laser tracker retroreflector. 

\myParagraph{Initialization of parameters.}  
To obtain the initial model, we used the nominal parameters of the robot provided by the manufacturer, previously acquired transformation matrices describing the mounting of the robot on the base, CAD model and manual measurements of the custom end effector and laser tracker retroreflector placement (the link EEL1/2 (7b) represents the transformation to the retroreflector from the last robot joint -- see Fig.~\ref{fig:setupMotoman}, Fig.~\ref{fig:ee}, and Table~\ref{tab:merged_dh}), and manual measurement of the cameras' attachment.
Using the Denavit-Hartenberg (DH) notation \cite{hartenberg1955kinematic}, Table~\ref{tab:merged_dh} shows parameters of the manipulators, including the custom end effector or the retroreflector placement. Table~\ref{tab:mounting_dh_cam} shows the two links to every camera.

\begin{table*}[!h]
\centering
\begin{tabular}{c|c|cccc|c|cccc}
&\multicolumn{5}{c|}{Manipulator 1 (right arm)}&\multicolumn{5}{c}{Manipulator 2 (left arm)}\\
Link&Link name&\bfseries a [m] & \bfseries d [m] & \bfseries $\boldsymbol{\alpha}$ [rad] & \bfseries $\boldsymbol{o}$ [rad] & Link name &
\bfseries a [m] & \bfseries d [m] & \bfseries $\boldsymbol{\alpha}$ [rad] & \bfseries $\boldsymbol{o}$ [rad]\\ \hline
0 & TT1 & \cellcolor{gray!30} 0  &\cellcolor{gray!30} -0.263 &\cellcolor{gray!30} $0.262$ &\cellcolor{gray!30} $-1.571$ & TT2 &\cellcolor{gray!30} 0  & \cellcolor{gray!30} -0.263 & \cellcolor{gray!30} $-0.262$ & \cellcolor{gray!30} $-1.571$\\
1&S1&\cellcolor{gray!30}0.150 & \cellcolor{gray!30} 1.416 &\cellcolor{gray!30} $-1.571$ & \cellcolor{gray!30} 0  &S2&\cellcolor{gray!30} 0.150 &\cellcolor{gray!30} 1.416 &\cellcolor{gray!30} $-1.571$ & \cellcolor{gray!30} $0 $\\
2&L1&0.614 &\cellcolor{gray!30} 0  & $3.142$ & $-1.571$ &L2& 0.614 & \cellcolor{gray!30}0  & $3.142$ & $-1.571$\\
3&U1&0.200 & 0  & $-1.571$ & 0  & U2& 0.200 & 0  & $-1.571$ & 0 \\
4&R1&0  & -0.640 & $1.571$ & 0  & R2 & 0  & -0.640 & $1.571$ & 0 \\
5&B1&0.030 & 0  & $1.571$ & $-1.571$ & B2 & 0.030 & 0  & $1.571$ & $-1.571$\\
6&T1&0  & \cellcolor{gray!30}0.200 & 0  & \cellcolor{gray!30}0  & T2 & 0  &\cellcolor{gray!30} 0.200 & 0  &\cellcolor{gray!30} 0 \\
7a&EE1&\cellcolor{gray!30}0  & 0.350 &\cellcolor{gray!30} 0  & 0  & EE2 & \cellcolor{gray!30}0 & 0.350 & \cellcolor{gray!30}0  & 0 \\
\end{tabular}
\vspace{2em}
\begin{tabular}{c|c|cccc|c|cccc}
&\multicolumn{5}{c|}{Laser tracker: Manipulator 1 (right arm)}&\multicolumn{5}{c}{Laser tracker: Manipulator 2 (left arm)}\\
Link&Link name&\bfseries a [m] & \bfseries d [m] & \bfseries $\boldsymbol{\alpha}$ [rad] & \bfseries $\boldsymbol{o}$ [rad] & Link name &
\bfseries a [m] & \bfseries d [m] & \bfseries $\boldsymbol{\alpha}$ [rad] & \bfseries $\boldsymbol{o}$ [rad]\\
\hline
6&T1&\cellcolor{gray!30}0  & \cellcolor{gray!30}0.200 & \cellcolor{gray!30}0  & \cellcolor{gray!30}0  & T2 & \cellcolor{gray!30}0  &\cellcolor{gray!30} 0.200 & \cellcolor{gray!30}0 &\cellcolor{gray!30} 0\\
7b&EEL1&0.02 & 0.250 & \cellcolor{gray!30}0 & $1.571$ & EEL2 & 0.020 & 0.250 &\cellcolor{gray!30} 0 & $1.571$\\
\end{tabular}
\caption{Complete initial DH parameter description of both arm kinematic chains. 7a is the last link to the icosahedron end effector; 7b is the link to the laser tracker retroreflector. The parameters with gray shading were not calibrated unless otherwise stated (when calibrating using the laser tracker, we do not calibrate the $6$th link). The individual links are visualized in Fig.~\ref{fig:setupMotoman}.}
\label{tab:merged_dh}
\end{table*}
\begin{table*}[!h]
\centering
\begin{tabular}{c|c|cccc|c|cccc}
&\multicolumn{5}{c|}{Camera 1}&\multicolumn{5}{c}{Camera 2}\\
Link& Link name &\bfseries a [m] & \bfseries d [m] & \bfseries $\boldsymbol{\alpha}$ [rad] & \bfseries $\boldsymbol{o}$ [rad] & Link name &
\bfseries a [m] & \bfseries d [m] & \bfseries $\boldsymbol{\alpha}$ [rad] & \bfseries $\boldsymbol{o}$ [rad]\\
\hline
0& TT3 &0.2315 & 1.8034  & -2.5086 & -2.7753 & TT4 &0.2315 & 1.8602  & 2.5486 & -0.0860\\
1& C1 &\cellcolor{gray!30}0 & -0.5670 & \cellcolor{gray!30}0 & 0.2863 & C2 &\cellcolor{gray!30} 0 & -0.4982 & \cellcolor{gray!30}0 & 3.0618
\end{tabular}
\caption{Initial DH parameters of camera chains. Parameters with gray shading were not subject to calibration unless otherwise stated.}
\label{tab:mounting_dh_cam}
\end{table*}

The manipulator has 6 actuated rotational joints, denoted ${S}$, ${L}$, ${U}$, ${R}$, ${B}$ and ${T}$ (\textit{in italics}), connected by links which are denoted S, L, U, R, B, and T in the order from the turntable to the end effector. Joint ${S}$ connects the turntable with link S of the robot; joint ${L}$ connects link S with link L, and so on -- see Fig.~\ref{fig:setupMotoman}. We added one last link to every manipulator's kinematic chain. These links represent the transformation to the end effector (icosahedron) (EE1, EE2) or retroreflector (EEL1, EEL2). Links between the origin and the turntables (TT1 and TT2) were added as well.
The camera chain was also expressed using DH representation and is composed of 2 links: the link between the origin and the camera turntable (TT3, TT4) and the link to the camera entrance pupil (C1, C2). These last links (EE1, EE2, EEL1, EEL2, C1, C2) are not connected by an actual joint, and thus their joint angle was set to zero. These joints may still have a non-zero offset $o$ though. 

We consider the complete kinematic system of the robot platform to be composed of four individual kinematic chains described by the DH parameters: (1) Right arm (TT1, S1, L1, U1, R1, B1, T1, EE1), (2) Left arm (TT2, S2, L2, U2, R2, B2, T2, EE2), (3) Right camera (TT3, C1) and (4) Left camera (TT4, C2). Additionally, we consider kinematic chains for laser tracker measurement (5) Right arm (TT1, S1, L1, U1, R1, B1, T1, EEL1), and (6) Left arm (TT2, S2, L2, U2, R2, B2, T2, EEL2). All the kinematic chains start in the base frame. The transformation from the base frame to the first joint, i.e.\ to the turntable joint, is identity. For calibration using the laser tracker, we replaced the icosahedron end effector links with  links ending in the retroreflectors (EEL1, EEL2) (see Fig.~\ref{fig:setupMotoman}, Table~\ref{tab:merged_dh} for robot DH parameters, and Table~\ref{tab:mounting_dh_cam} for camera DH parameters).
The left and right arm chains finish with the end effectors; the last frame is in the center of each icosahedron or in the laser tracker retroreflector. Note that since the mounting of the two manipulators on the turntable is not identical, the first 4-tuple of DH parameters is also not the same; we recognize two distinct turntable links in the arm chains and two more turntable links in the camera chains.  All the chains have the same rotation of the first joint, but all other joints and parameters are independent. 

\subsection{Camera calibration}
\label{section:camera_calib}
The calibration of the intrinsic parameters of the cameras was carried out with a dot pattern.
The dataset used for calibration was composed of 22 pattern images.
Each of the captured images had a different position and orientation w.r.t. cameras. Calibration of the camera matrix $K$ and distortion coefficients vector $\mathbf{d}$ was performed using OpenCV camera calibration function \textit{calibrateCamera}~\cite{opencv_library}, and the following pinhole camera model extended with radial and tangential distortion coefficients:
\begin{equation}
\begin{split}
x_c' = x_c/z_c,& \qquad y_c' = y_c/z_c, \qquad r = \sqrt{x_c'^{2}+y_c'^{2}}\\
x_c'' = & x_c'(1 + d_{1}r^{2} + d_{2}r^{4} + d_{3}r^{6}) +\\  &2d_{4}x_c'y_c' + d_{5}(r^{2} + 2x_c'^{2})\\
y_c'' = & y_c'(1 + d_{1}r^{2} + d_{2}r^{4} + d_{3}r^{6}) +\\& d_{4}(r^{2} + 2y_c'^{2}) + 2d_{5}x_c'y_c'\\
\begin{bmatrix}u\\v\\1\end{bmatrix} = & K\begin{bmatrix}x_c''\\y_c''\\1\end{bmatrix}
\end{split}
\label{eq:projection}
\end{equation}
where $[x_c,y_c,z_c]$ is a 3D point in camera frame, $[u,v]$ are the image coordinates.

New camera matrices $K$ and the distortion coefficients $\mathbf{d}_R$ and $\mathbf{d}_L$  for right and left camera, respectively, were found as:
\begin{gather}
K_R = \begin{bmatrix}
 8185.397 &  0.000 &  2009.318\\  
 0.000 &  8170.401 &  2963.960\\
 0.000 &  0.000 &  1.000
\end{bmatrix},  \nonumber\\
K_L = \begin{bmatrix}
 8110.478 & 0.000 &  1949.921\\
 0.000 & 8098.218 & 2991.727\\
 0.000 &  0.000 &  1.000
\end{bmatrix},
\label{eq:cam_params}
\end{gather}
\begin{equation}
\begin{split}
\mathbf{d}_R &= 
[\begin{matrix}-0.020602 & -0.205606 & -0.001819 \end{matrix} \\
 &\qquad\qquad \begin{matrix} -0.000820 &  0.718890 \end{matrix}] \nonumber
\end{split}
\end{equation}
\begin{equation}
\begin{split}
\mathbf{d}_L &= 
[\begin{matrix}-0.022546 & -0.213094 & -0.000684 \end{matrix} \\
 &\qquad\qquad \begin{matrix} -0.000512 & 0.662333 \end{matrix}]
\end{split}
\label{eq:cam_params_dist}
\end{equation}

The distortion coefficients are presented here in the order as they are returned from OpenCV camera calibration function \textit{calibrateCamera}~\cite{opencv_library}.

\subsection{Data acquisition and description}
\label{section:dataset_description}
An accompanying video illustrating the experiments is available at \url{https://youtu.be/LMwINqA1t9w}. There were 5 distinct datasets collected: (1) self-contact/self-touch experiment (both robot end effectors get into contact), (2) contact with a lower and upper horizontal plane, (3) contact with a vertical plane (``wall''), (4) repeatability measurement, and (5) laser tracker dataset. 
Datasets were acquired using a simple GUI applet \cite{puciow_thesis}.
Laser tracker data were also recorded for parts of self-contact and planar contact datasets (not used in this paper but included in the published dataset) \cite{dataset_our}.

\myParagraph{Individual datasets.} The whole dataset $\boldsymbol{D}^{whole}$ is a set of individual datasets:
\begin{equation*}
\boldsymbol{D}^{whole}=\{\boldsymbol{D}^{st}, \boldsymbol{D}^{hp}, \boldsymbol{D}^{vp}, \boldsymbol{D}^{lt}\}, 
\label{eq:dataset}
\end{equation*}
where $\boldsymbol{D}^{st}$, $\boldsymbol{D}^{hp}$, $\boldsymbol{D}^{vp}$, and $\boldsymbol{D}^{lt}$ are datasets collected during self-contact/self-touch experiment (st), contact with horizontal planes (hp), contact with a vertical plane (vp), and laser tracker experiments (lt), respectively. 

The world coordinate system is a right-handed Cartesian coordinate system with axes denoted by $x$, $y$, $z$. Units are meters; the origin is in the center of the robot base on the floor, the $y$-axis points behind the robot, and $z$-axis points up.  
Euler angles refer to the orientation of the end effector with respect to a coordinate system $x'$, $y'$, $z'$ attached to a moving body (end effector). In our notation, $\alpha$, $\beta$, and $\gamma$ represent the first, second, and third rotation, respectively. The rotation axis is indicated by the subscript ($x$, $y$, $z$ and  $x'$, $y'$, $z'$ denote the fixed and current axes, respectively).

\begin{table*}[!h]
    \centering
    \begin{tabular}{p{\dimexpr 0.13\linewidth-2\tabcolsep}|p{\dimexpr 0.35\linewidth-2\tabcolsep}p{\dimexpr 0.35\linewidth-2\tabcolsep}p{\dimexpr 0.17\linewidth-2\tabcolsep}}
         Dataset & Contact positions & Orientations & No. datapoints \\
         \hline
         Self-touch ($\boldsymbol{D}^{st}$) & 4$\times$4$\times$2 grid in the $xyz$-box with $x \in [-0.3, 0.2]\si{\meter}$, $y \in [-1.1, -0.6]\si{\meter}$ and $z = \{0.8,\,1\}\si{\meter}$ & 9 combinations of right and left arm orientations -- see Fig.~\ref{fig:datasetSelftouch} & 566 datapoints (10 not logged) \\ \hline
         Horizontal planes ($\boldsymbol{D}^{hp}$) & 5$\times$5 grid in the $xy$-plane with $x \in [-0.4, 0.3]\si{\meter}$, $y \in [-1.35, -0.65]\si{\meter}$, and $z = \{0.67,\,85\}\si{\meter}$ & 5~orientations in Euler angles $(\alpha_x,\beta_y,\gamma_z)$: \{$(0, 185, 52)$, $(0, 180, -20)$, $(0, 180, 124)$, $(0, 180, -164)$, $(0, 180,-92)\}$ & 454 datapoints (48 not logged) \\
         \hline
         Vertical plane ($\boldsymbol{D}^{vp}$) & 5$\times$5 grid in the $yz$-plane with $y \in [-1.2,-0.7]\text{ m}$, $z = [0.7,1.1]\text{ m}$, and $x = 0.05\text{ m}$ & 5~orientations in Euler angles $(\alpha_z,\beta_y,\gamma_z)$: \{$(180, 90, 0)$, $(108, 90, 0)$, $(36, 90, 0)$, $(-36, 90, 0)$, $(-108, 90, 0)\}$ & 248 datapoints (2 not logged)  \\
         \hline
         \multirow{3}{2cm}{Repeatability measurement} & \begin{tabular}[t]{@{}p{0.5cm}|l@{}}\multirow{2}{*}{st} & $[x,y,z]^T= [0.225, -0.85, 0.8]^T\si{\meter}$ \\ & $[x,y,z]^T=[-0.025, -0.85, 1]^T\si{\meter}$\end{tabular} &
         \begin{tabular}[t]{@{}p{1.3cm}l@{}} right arm & $(\alpha_z,\beta_y,\gamma_z): (-108, 90, 0)$ \\left arm & $(\alpha_z,\beta_y,\gamma_z): (0, -90, 0)$  \end{tabular} & 20 datapoints \\ \cline{2-4} &
         \begin{tabular}[t]{@{}p{0.5cm}|l@{}} \multirow{2}{*}{hp} & $[x,y,z]^T= [0.3,-0.825,0.67]^T\si{\meter}$ \\ & $[x,y,z]^T= [-0.4,-0.825,0.67]^T\si{\meter}$ \end{tabular} & \begin{tabular}[t]{@{}p{1.3cm}l@{}} \multirow{2}{*}{right arm} & \multirow{2}{*}{$(\alpha_x,\beta_y,\gamma_z): (0, 185, 52)$} \\ & \end{tabular} & 20 datapoints\\ \cline{2-4} & \begin{tabular}[t]{@{}p{0.5cm}|l@{}} \multirow{2}{*}{vp} & $[x,y,z]^T= [0.17,-0.825,1.1]^T\si{\meter}$ \\ & $[x,y,z]^T=[0.17,-0.95,0.7]^T\si{\meter}$\end{tabular} &   \begin{tabular}[t]{@{}p{1.3cm}l@{}} \multirow{2}{*}{right arm} & \multirow{2}{*}{$(\alpha_z,\beta_y,\gamma_z): (-108, 90, 0)$} \\ & \end{tabular} & 20 datapoints \\
         \hline
         Laser tracker ($\boldsymbol{D}^{lt}$) & whole configuration space of the manipulator was sampled  (see text for details)& - & 586 datapoints (99 not logged)
    \end{tabular}
    \caption{Dataset overview.}
    \label{tab:datasets}
\end{table*}

\begin{figure}[!h]
\makebox[0.5\textwidth]{
\includegraphics[width=0.35\textwidth]{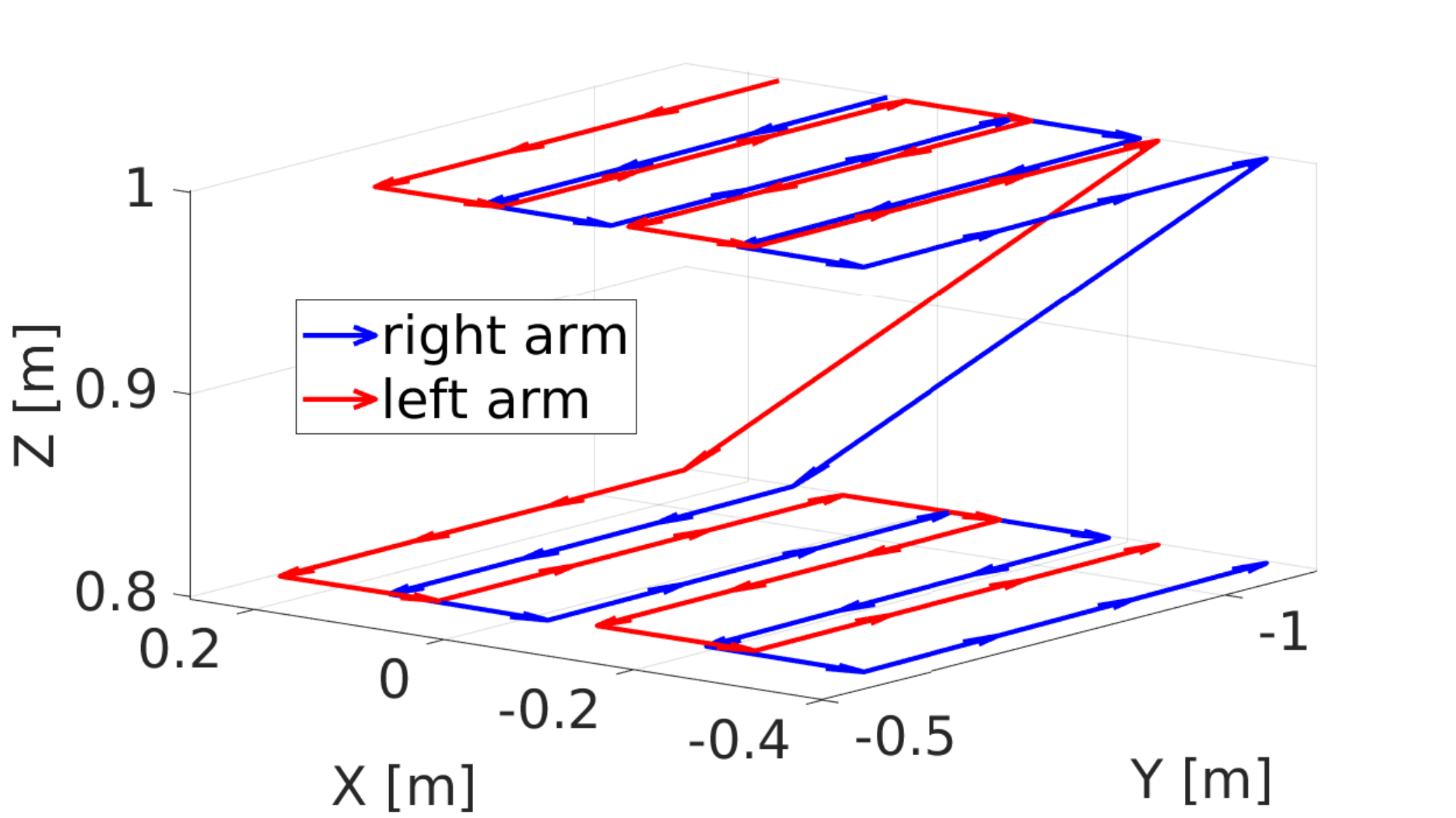}}
\caption{Positions of end effector centers during the self-contact experiment. The end of each arrow denotes a position of end effector center in individual poses (red -- left arm, blue -- right arm). The distance between left and right arm position is given by two times the radius of the end effectors in contact.}
\label{fig:datasetSelftouch}
\end{figure}

\begin{figure}[!h]
\centering
\makebox[0.5\textwidth]{
\includegraphics[width=0.25\textwidth]{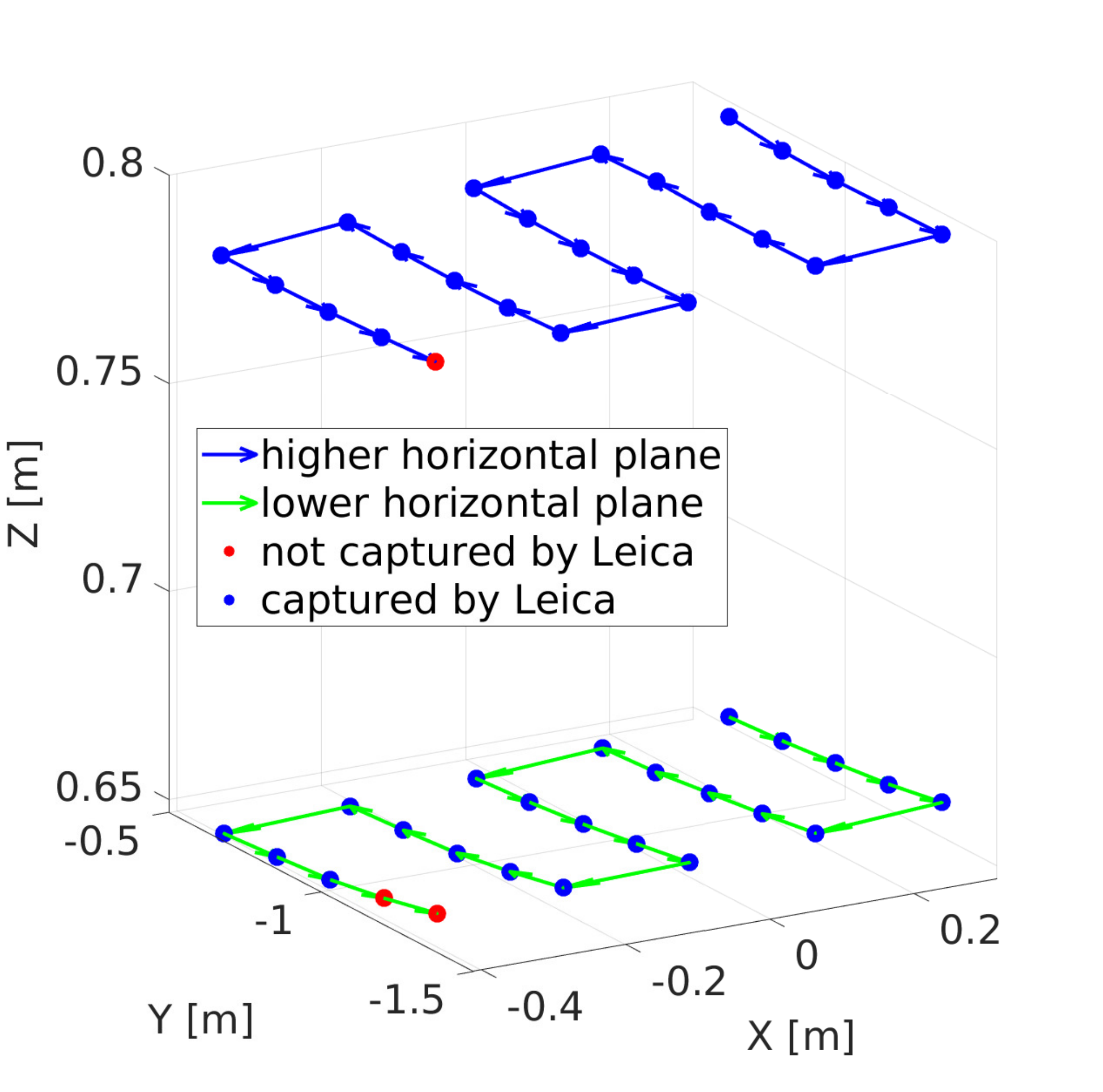}
\includegraphics[width=0.25\textwidth]{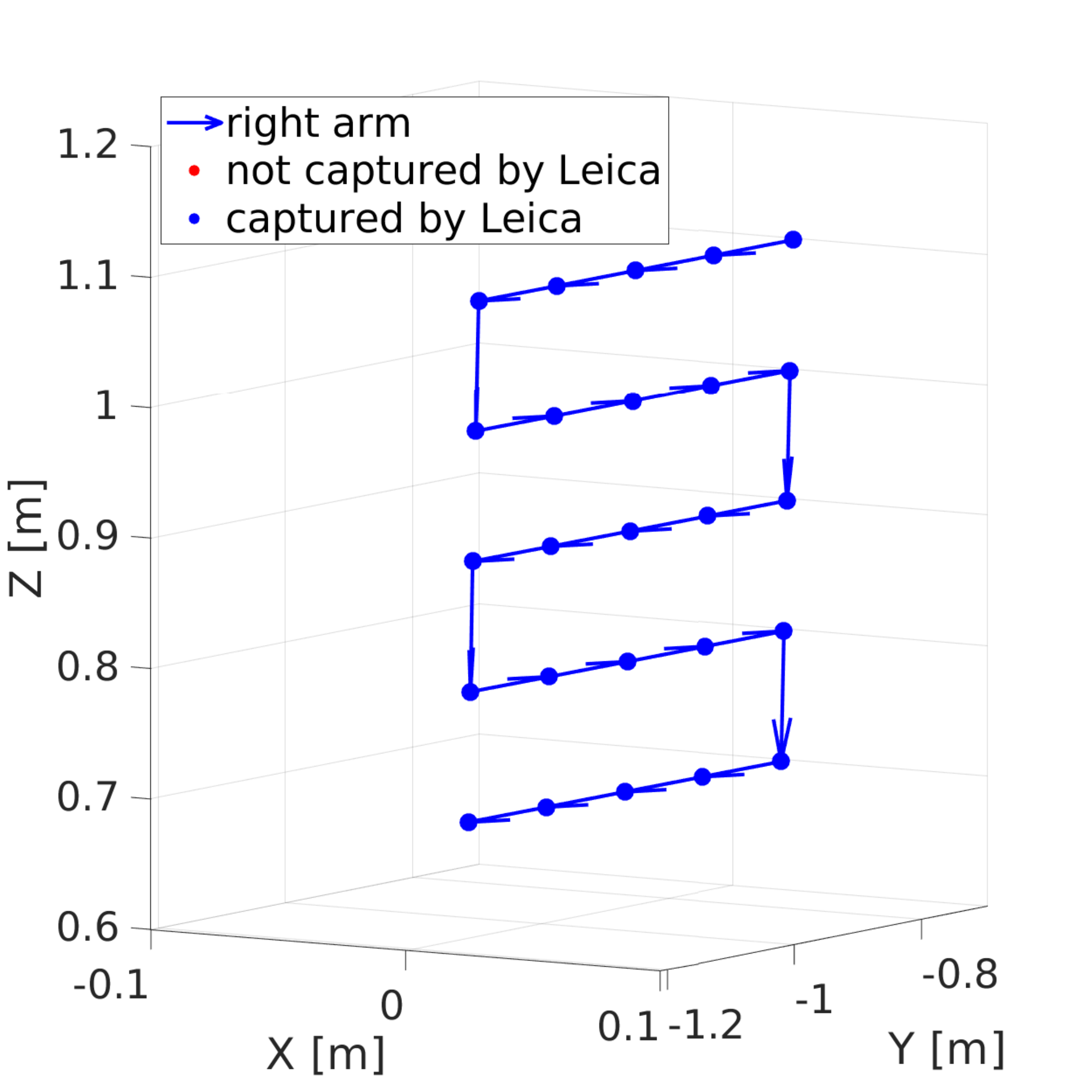}
}
\caption[Horizontal plane grid]{Positions of end effector centers in the horizontal plane contact datasets (left, higher plane in blue, lower plane in green) and vertical plane contact dataset (right) with the information whether the pose was logged by laser tracker (blue) or not (red).}
\label{fig:datasetPlanar}
\end{figure}

\begin{figure}[!h]
\centering
\includegraphics[width=0.5\textwidth]{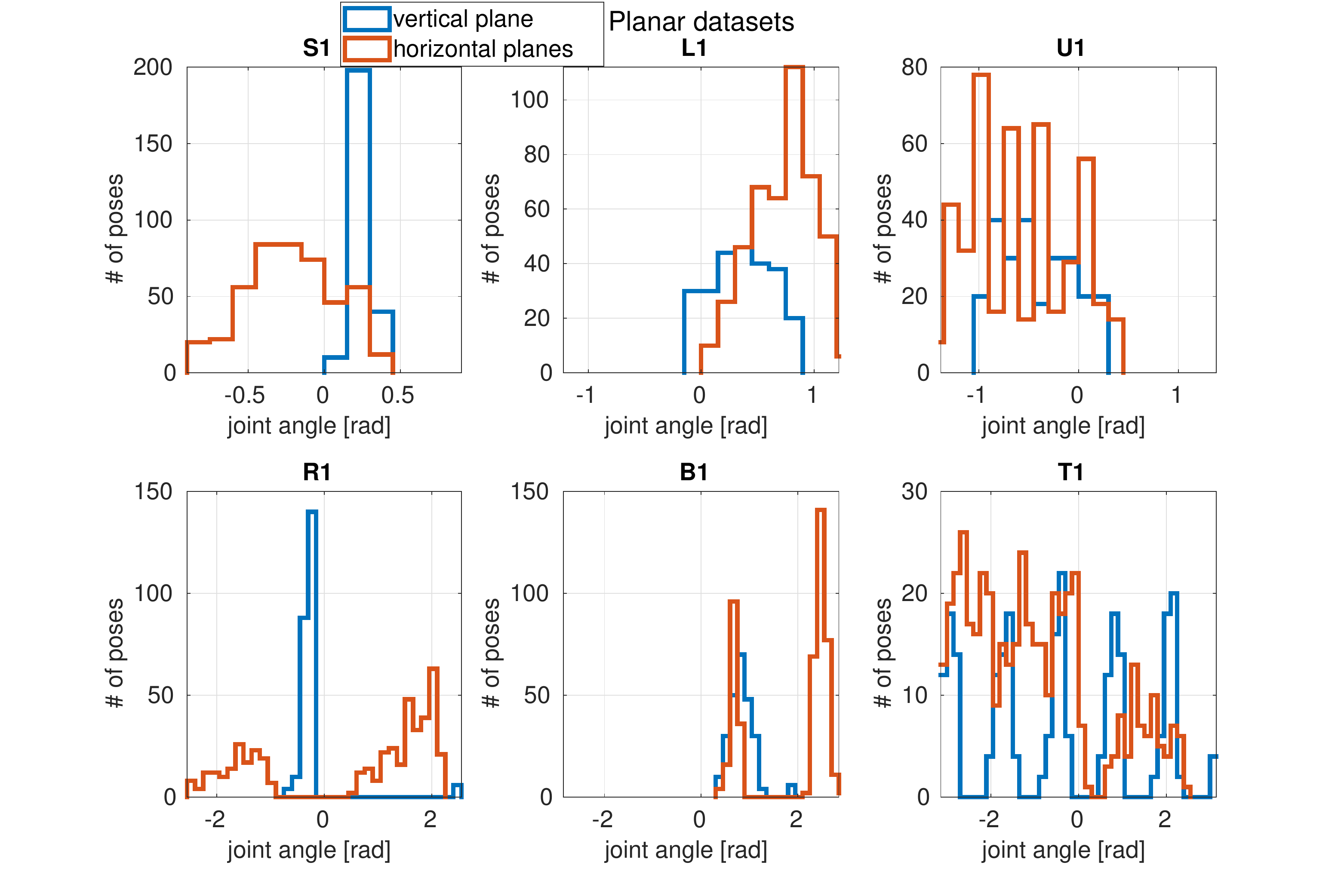}
\includegraphics[width=0.5\textwidth]{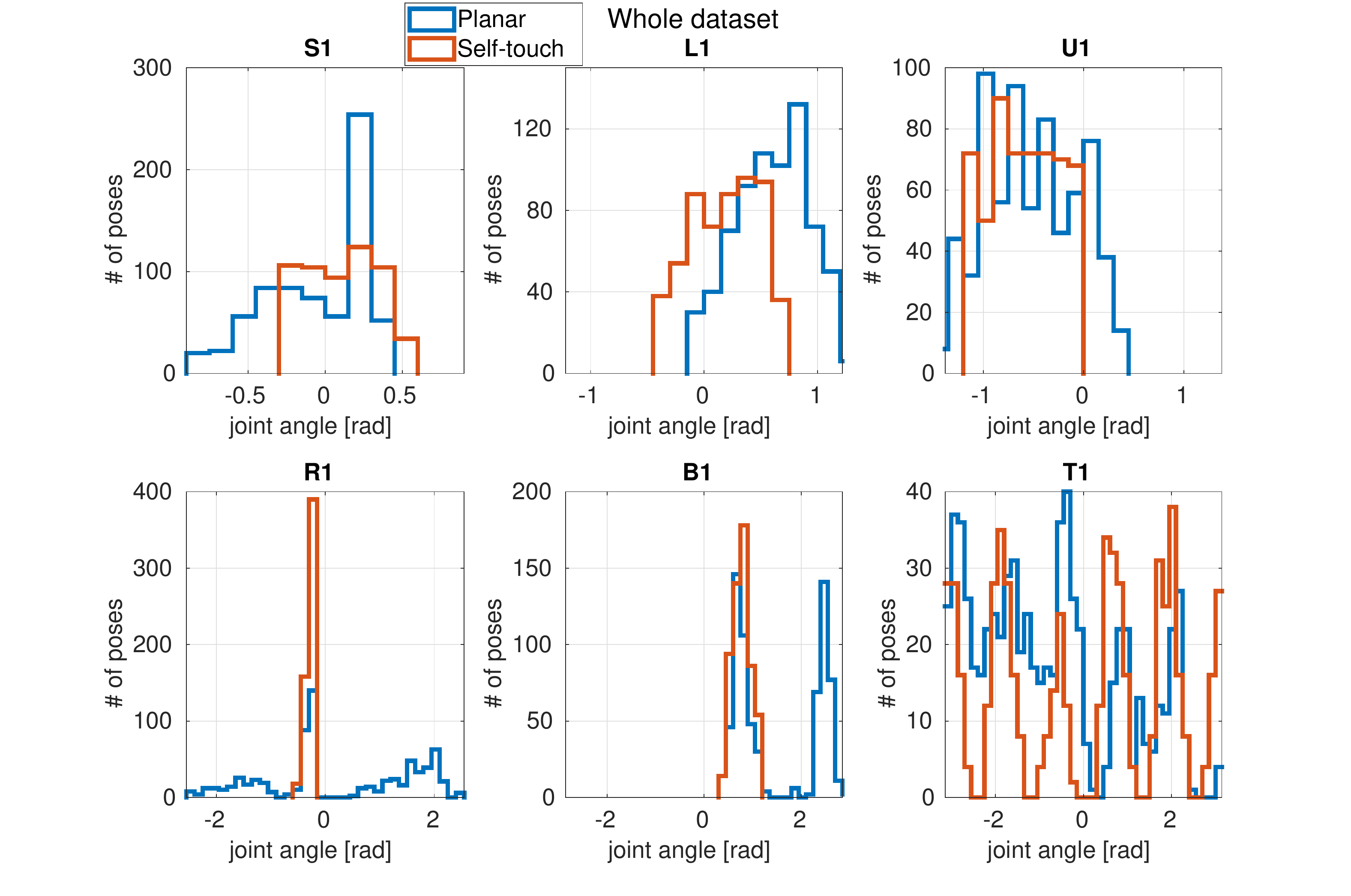}
\caption[Joint distribution]{\label{fig:joints}Distributions of robot joint angles (S1,L1,U1,R1,B1,T1) for measured poses for the right arm across different experiments (only the right arm was evaluated in all experiments). ``Planar datasets'' subfigure compares the vertical plane contact in blue with the horizontal plane contact in red. ``Whole dataset'' subfigure compares the self-touch experiment in red with the combination of all planar setups (2 horizontal + 1 vertical plane) in blue.}
\label{fig:distJoints}
\end{figure}

Contact points of end effectors in datasets $\boldsymbol{D}^{st}$, $\boldsymbol{D}^{hp}$, and $\boldsymbol{D}^{vp}$ were planned on grids in the manipulators' workspace, as shown in Fig.~\ref{fig:datasetSelftouch} and Fig.~\ref{fig:datasetPlanar} for the self-contact and contact with planes, respectively.
Several end effector orientations were tested for every position. A detailed description of individual datasets with contact points and orientations is provided in Table~\ref{tab:datasets}. Every experiment was repeated twice; a few configurations could not be reached due to robot motion planner failure. In all cases, photographs of one or both icosahedrons in every pose were taken and added to the dataset.
In addition, for every setup (self-touch, horizontal and vertical plane), 20 repetitions in 2 different positions were performed to evaluate the measurements' repeatability (Repeatability measurement dataset). 
For kinematic calibration, the distribution of robot joint angles is important. It is visualized in Fig.~\ref{fig:distJoints} for the different experiments.

\begin{figure}[!h]
\centering
\begin{minipage}[c]{0.27\textwidth}
    \includegraphics[width=\textwidth]{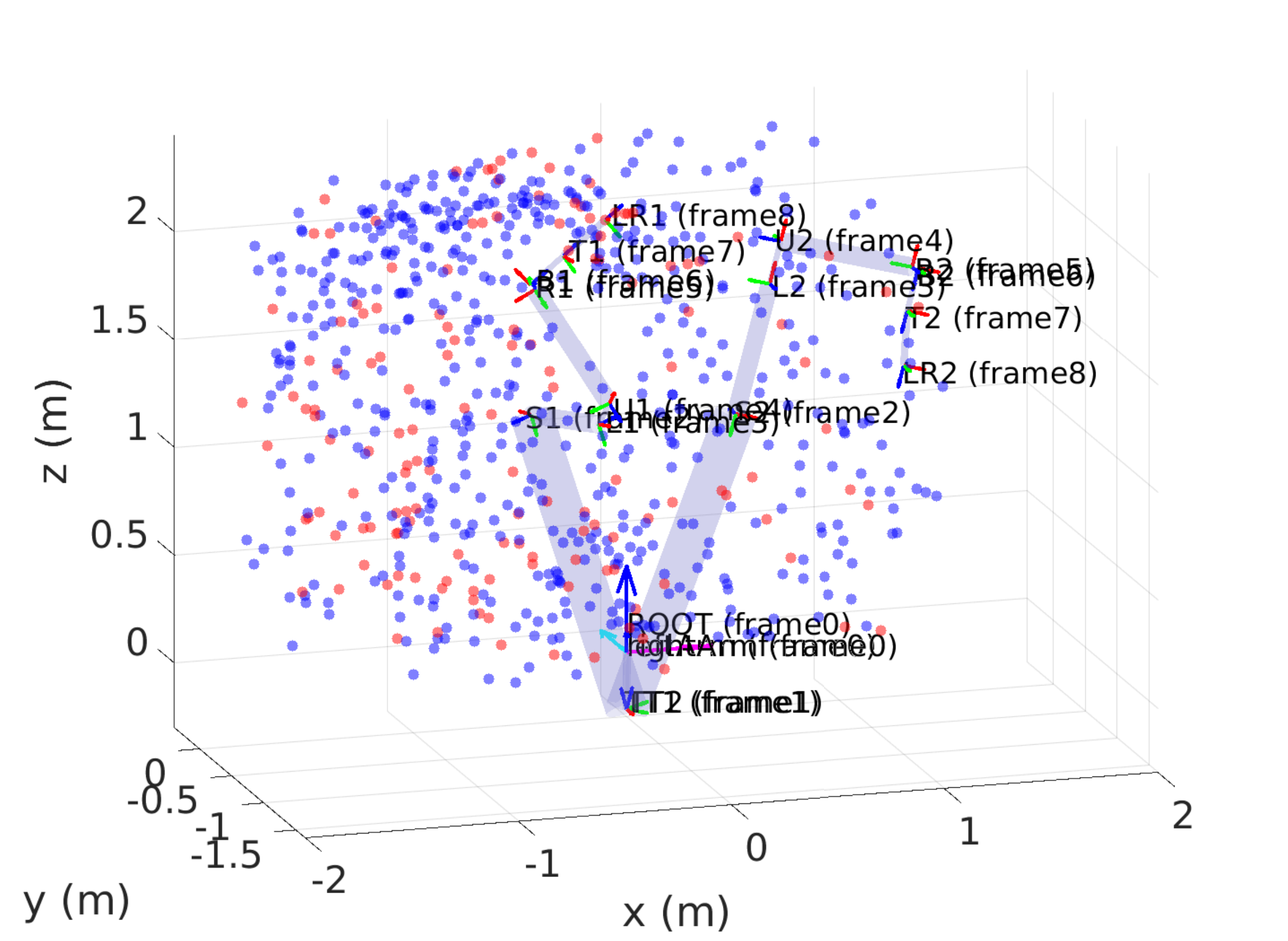}
  \end{minipage}\hfill
  \begin{minipage}[c]{0.2\textwidth}
    \caption[Laser tracker dataset grid]{Positions of the retroreflector in the Laser tracker dataset (right arm) with information whether the pose was additionally logged by the tracker or not (blue -- captured by the tracker (586 poses), red -- not captured by the tracker (99 poses) due to the poor visibility of the retroreflector).} \label{fig:datasetLeica}
  \end{minipage}
\end{figure}

\begin{figure}[!h]
\centering
\includegraphics[width=0.5\textwidth]{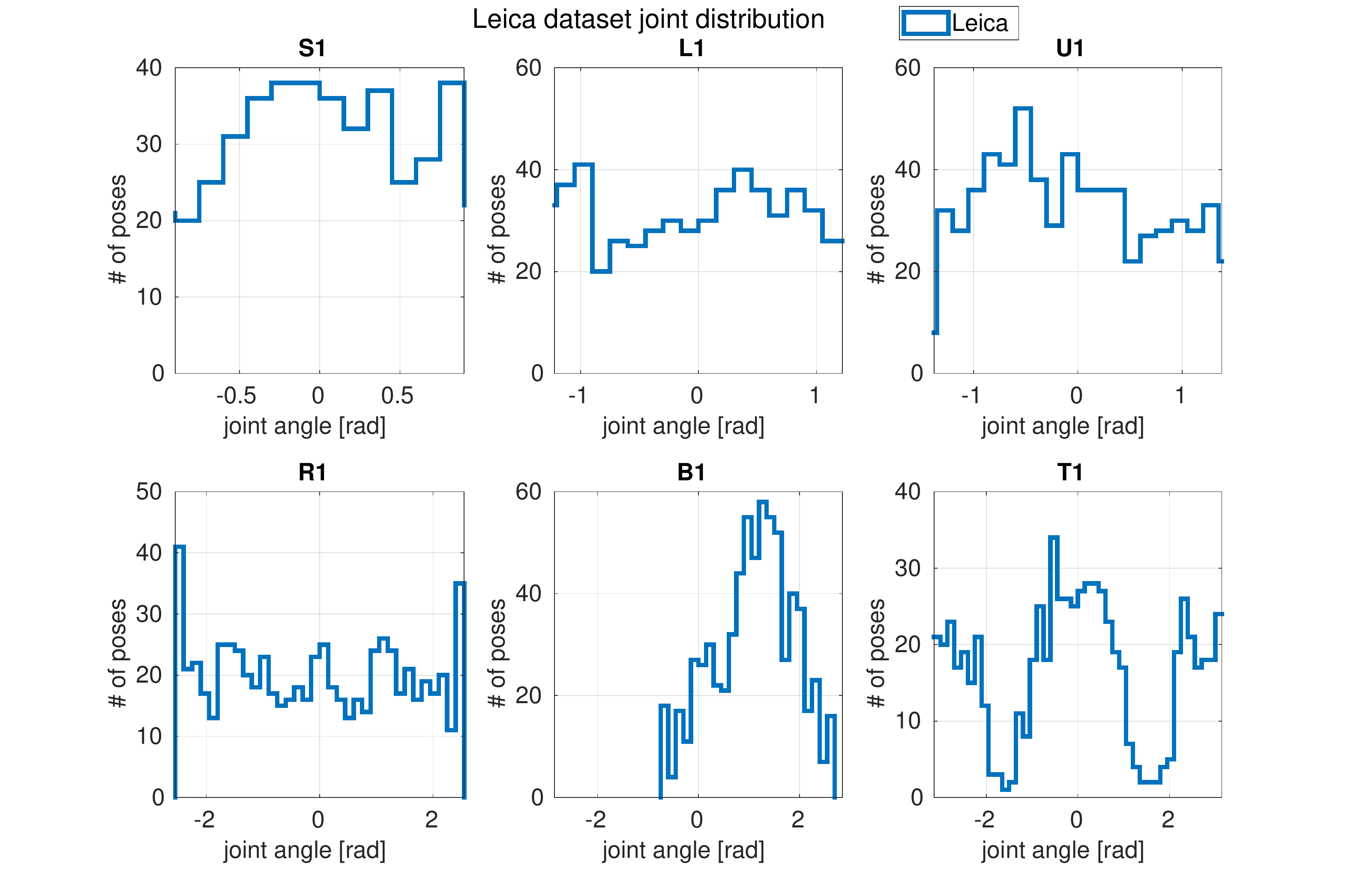}
\caption[Joint distribution]{\label{fig:joints_leica}Distribution of joint angles for Laser tracker dataset (right arm).}
\label{fig:distJointsLeica}
\end{figure}

\noindent\textbf{Laser tracker experiment ($\boldsymbol{D}^{lt}$).} We sampled the whole range of the first 4 joints of the manipulator and added uniform noise to the values to cover the whole range of joint angles. The joint angles of the last two joints were set so that the retroreflector faced the laser tracker. Configurations which would be in collision with the robot or the surrounding environment were excluded. 
This resulted in 685 configurations for the right arm, out of which 586 poses were actually recorded by the laser tracker ((see Fig.~\ref{fig:datasetLeica}; Fig.~\ref{fig:distJointsLeica} for joint space distribution)).

\myParagraph{Dataset structure.} The whole dataset $\boldsymbol{D}^{whole}$ contains $M$ dataset points (in fact, they are row vectors): $\boldsymbol{D}^{whole}  = \{\boldsymbol{D}_1,\dots,\boldsymbol{D}_M\}$. Each dataset point $\boldsymbol{D}_j$ of the dataset $\boldsymbol{D}^{whole}$ consists of the assumed pose of right and left icosahedron centers $\boldsymbol{x}^{I,r}$ and $\boldsymbol{x}^{I,l}$, respectively (computed from forward kinematics); the joint configuration of the robot $\boldsymbol{\theta} = \{\theta^{ra}_1,...,\theta^{ra}_N,\theta^{la}_1,...,\theta^{la}_N\}$ ($ra$ and $la$ denoting right and left arm, respectively); coordinates of every marker ($K$ being the number of markers) in each of the cameras $\boldsymbol{u} =\{\boldsymbol{u}^{rc}_1,...,\boldsymbol{u}^{rc}_K, \boldsymbol{u}^{lc}_1,...,\boldsymbol{u}^{lc}_K\}$ ($rc$ and $lc$ denoting right camera and left camera, respectively); position of laser tracker ball retroreflector in the tracker coordinate system including uncertainties U95~\cite{Leica} of position detection and the tracker measurement timestamp $\boldsymbol{L} = \{\mathbf{x}^L, u^L, t^L\}$. We also saved for reference the magnitude of force measured by both force sensors before ($F_b$) and during ($F_a$) the contact for each arm $\boldsymbol{F} = \{F^{ra}_b$, $F^{ra}_a$, $F^{la}_b$, $F^{la}_a\}$; and the names of saved camera images $n^r, n^l$. Individual dataset points are organized in a matrix, where each line $i$ of the matrix corresponds to the individual dataset point:

\begin{equation}
\begin{split}
& \boldsymbol{D}^{whole}  = [\boldsymbol{D}_1,\dots,\boldsymbol{D}_M]^T, \mathrm{where: }\\
& \boldsymbol{D}_i  =[\boldsymbol{x}^{I,r}_i, \boldsymbol{x}^{I,l}_i, \boldsymbol{\theta}_i, \boldsymbol{F}_i, \boldsymbol{L}_i, n^r, n^l,\boldsymbol{u}_i ]^T.
\end{split}
\label{eq:dataset2}
\end{equation}

The whole dataset $\boldsymbol{D}^{whole}$ contains $M = 1268$ logged poses with 23022 marker reprojections in total (12371 from the right camera and 10651 from the left camera), which makes approx. 25 marker reprojection per pose for self-contact experiment and 11 marker reprojections per pose for planar constraints. There are also 586 poses acquired by the laser tracker---this dataset does not contain marker reprojections. The reprojections are sorted from the lowest to the highest marker ID for every camera image. If a marker was not found in the image, its coordinates are denoted as $(NaN, NaN)$. Similarly, if data were not measured or captured by Leica tracker, all corresponding  values are filled with $NaN$. 

For optimization, we transformed the dataset so that one line would relate to one data point. That is, for a single robot configuration with multiple marker reprojections, the corresponding datapoint is unfolded into multiple rows---1 per marker detected. A number defining the robot configuration ($i$ from the original dataset $\boldsymbol{D}^{whole}$) is repeated on every line. Thus, one line consists of a number defining the robot pose, a face number of the detected marker, index of the arm (1 for right or 2 for left), index of the camera (1 for right or 2 for left), position of the marker center in the camera $\boldsymbol{u} = (u, v)$ in [px], and the current robot joint configuration (turntable, S1, L1, U1, R1, B1, T1, S2, L2, U2, R2, B2, T2) in [rad], position of the tracker ball retroreflector in the tracker coordinate system, and uncertainty U95 of the measurement. 

The dataset and its description can be downloaded from \cite{dataset_our}. For the positions including LEICA measurements, the appropriate csv file with ($x$,$y$,$z$) positions detected by the laser tracker scanner are available.

\section{Multi-chain robot calibration}
\label{sec:multichain}

In multi-chain robot calibration we estimate parameter vector ${\boldsymbol{\phi}} =\{
[a_1,...,a_n], [d_1,...,d_n], [\alpha_1,...,\alpha_n], [o_1,...,o_n]\}$ with $k \in N$, where $N = \{1,..,n \}$ is a set of indices identifying individual links; $a_k$, $d_k$ and $\alpha_k$ are the first three parameters of the DH formulation of link $k$; $o_k$ is the offset that specifies the positioning of the encoders on the joints with respect to the DH representation. This is sometimes referred to as static geometric calibration. We often estimate a subset of these parameters only,  assuming that the others are known. This subset can for example consist of a subset of links $N' \subset N$ (e.g., only parameters of one arm are to be calibrated) or a subset of the link parameters. Here we focus on offsets in the revolute joints $\mathbf{o}$ (sometimes dubbed ``daily calibration'' \cite{Nickels2003}). 

Let $\boldsymbol{D} \subset \boldsymbol{D}^{whole}$ denote the set of robot configurations (dataset points) used for optimization: 
$$\boldsymbol{D}_i = [m_i, c_i, \boldsymbol{u}_i, \boldsymbol{\theta}_i]$$
where $i \in \{1,..,M'\}$ is an index identifying one particular dataset point, $M'$ is the number of dataset points used for optimization, $m_i$ is face number of the detected marker, $c_i$ is the index of the used camera, $\boldsymbol{u}_i = (u_i, v_i)$ is the position of the marker center in the camera, and  $\boldsymbol{\theta}_i$ is the current robot joint configuration (joint angles from joint encoders for the given robot configuration). 

Estimation of the parameter vector $\boldsymbol{\phi}$ is done by optimizing a given objective function $\boldsymbol{f}(\boldsymbol{\phi}, \boldsymbol{D}, \boldsymbol{\zeta})$:
\begin{align}
\boldsymbol{\phi}^* =& \argmin_{\boldsymbol{\phi}} \boldsymbol{f}(\boldsymbol{\phi}, \boldsymbol{D}, \boldsymbol{\zeta}), \\
\boldsymbol{f}(\boldsymbol{\phi}, \boldsymbol{D}, \boldsymbol{\zeta})
=&\|\boldsymbol{g}(\boldsymbol{\phi}, \boldsymbol{D}, \boldsymbol{\zeta})\|^2 = \sum_{i=1}^{M'} g(\boldsymbol{\phi}, \boldsymbol{D}_i, \boldsymbol{\zeta})^2,
\label{eq:optimization}
\end{align}
where $M'$ is the number of robot configurations and corresponding end effector positions used for calibration (hereafter often referred to as ``poses'' for short), ${\boldsymbol{\phi}}$ is a given parameter estimate, dataset point $\boldsymbol{D}_i$ includes  joint angles ${\boldsymbol{\theta}}_i$ from joint encoders for the given robot configuration, and constant vector $\boldsymbol{\zeta}$ defines all other necessary parameters such as camera calibration, fixed transformations, fixed DH parameters, or other properties of the robot. For chains involving cameras, the function ${g}(\boldsymbol{\phi}, \boldsymbol{D}_i, \boldsymbol{\zeta})$ refers to reprojection error as described in the next section while for the chains including contact, it corresponds to the distance between a real (observed) end effector position ${\boldsymbol{p}}_i^r$ and an estimated end effector position ${\boldsymbol{p}}_i^e$ computed using forward kinematic function for a given parameter estimate ${\boldsymbol{\phi}}$ and joint angles from joint encoders ${\boldsymbol{\theta}}_i$. For the case of planar constraints, the shortest distance of end effector position (icosahedron center) to the parameterized plane is minimized.

We study different combinations of intersecting chains and their performance in calibrating one another. Specific form of the function $g(\boldsymbol{\phi}, \boldsymbol{D}_i, \boldsymbol{\zeta})$ for individual considered chains and their combinations is discussed in the following subsections. At the end of every subsection, we also state how many components of the pose can be measured using the different methods---similarly to the analysis for standard open-loop and closed-loop calibration approaches in \cite{Hollerbach2016}.

For problem representation, optimization, and some visualization, we employed the multisensorial calibration toolbox \cite{multirobot-github}.

\subsection{Self-contact -- two arms chain (LA-RA)}
\label{section:self_contact}

\begin{figure}[!h]
\centering
\begin{minipage}[c]{0.4\textwidth}
    \includegraphics[width=1\textwidth]{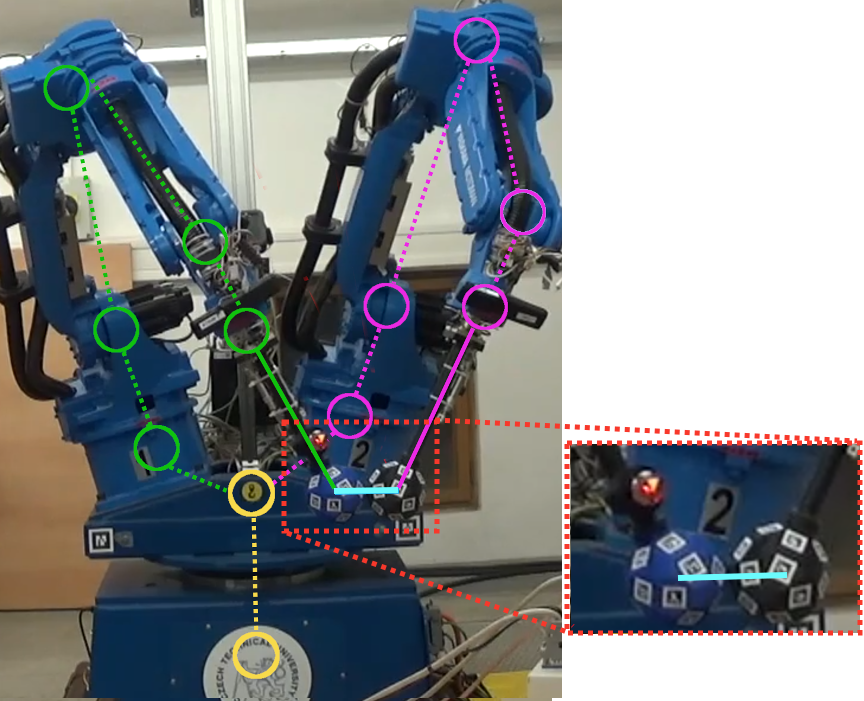}
  \end{minipage}\hfill
  \begin{minipage}[c]{0.5\textwidth}
    \caption{Self-contact experiment. The left and right arm chains are drawn in purple and green, respectively. Cyan indicates the distance between end effector centers.} \label{fig:SelfTouchSetup}
  \end{minipage}
\end{figure}

\begin{figure}
    \centering
    \includegraphics[width = .24\textwidth]{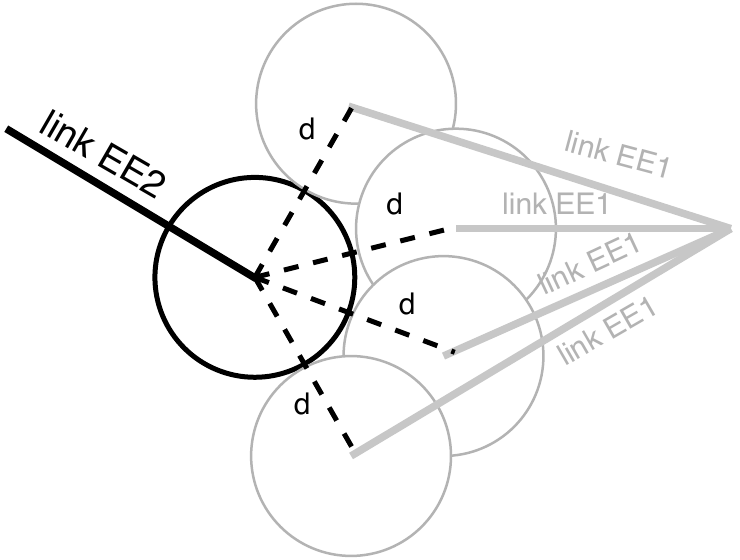}
    \includegraphics[width = .24\textwidth]{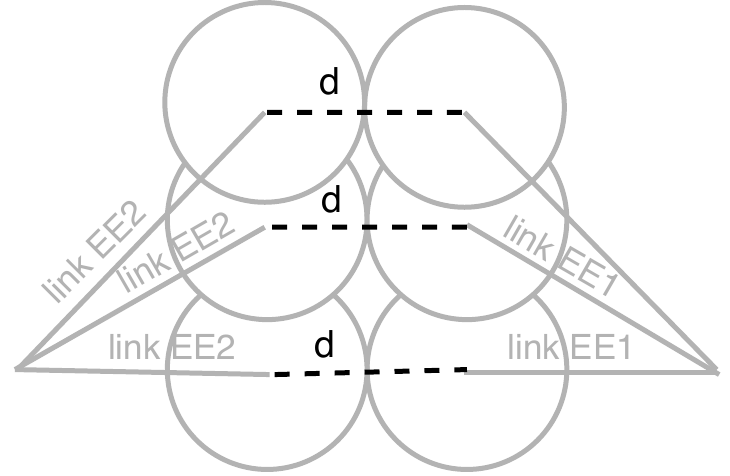}
    \caption{Visualisation of self-contact scenario. When end effectors are in contact, they are separated by distance $d$. This can happen in multiple configurations. For simplicity, we visualize the case when only end effector link length is estimated for one arm (left) or both arms (right), with the additional assumption that the link length is the  same for both arms.}
    \label{fig:st_drawing}
\end{figure}

This corresponds to the self-contact scenario in which contact occurs directly between the end effectors of the Left Arm (LA) and Right Arm (RA). As described in  Section~\ref{subsec:robot_setup_description}, for contact, the end effectors can be treated as spheres and the contact can occur between any of the 10 spherical tiles. The newly established kinematic chain for the upper body includes both arms; the head and eyes are excluded (see Fig.~\ref{fig:SelfTouchSetup}). To optimize parameters describing this chain, we minimize the distance between estimated positions in the 3D space of left and right arm end effectors (see Fig.~\ref{fig:st_drawing}). 

In this case, the parameter vector ${\boldsymbol{\phi}}$ consists of the following parameters: ${\boldsymbol{\phi}} = \{{\boldsymbol{\phi}}^{ra},{\boldsymbol{\phi}}^{la}\}$, where ${\boldsymbol{\phi}}^{ra}$ and $\boldsymbol{\phi}^{la}$ are DH parameters being calibrated corresponding to the robot right and left arm, respectively. The objective function to be optimized is:
\begin{equation}
\begin{split}
\boldsymbol{g^{st}}(\boldsymbol{\phi}, \boldsymbol{D}^{st,u}, \zeta) =& [c(\boldsymbol{\phi}, \boldsymbol{D}_1, \boldsymbol{\zeta}) - q(\boldsymbol{\zeta}) ,...,\\
&c(\boldsymbol{\phi}, \boldsymbol{D}_M, \boldsymbol{\zeta}) - q(\boldsymbol{\zeta})]
\end{split}
\label{eq:objST}
\end{equation}
where the function $c(\boldsymbol{\phi}, \boldsymbol{D}_i, \boldsymbol{\zeta})$ computes the distance of the end effector centers in the configuration given by the dataset point $\boldsymbol{D}_i$, where $\boldsymbol{D}_i \in \boldsymbol{D}^{st,u}$, $\boldsymbol{D}^{st,u} \subset \boldsymbol{D}^{st}$ is a subset of the dataset points with a particular robot configuration (see Dataset structure in Sec.~\ref{section:dataset_description} for details).  The distance of the end effector centers, marked as $q(\boldsymbol{\zeta})$, is equal to one icosahedron diameter $2r$, because both end effectors have identical shape. For the icosahedron diameter, we took the value from CAD model of $2r = 116$~mm and kept it fixed~\cite{puciow_thesis}.

As can be seen, the objective function contains a set of constraints on the distances between the end effector positions. These constraints can be written for the case of self-contact as follows: Let $(x^r_i, y^r_i,z^r_i)$ and $(x^l_i, y^l_i, z^l_i)$, be the centre of right and left arm end effector computed from forward kinematics for data point $i$ with the given joint configuration $\boldsymbol{\theta}_i$, respectively. Then for each data point $\boldsymbol{D}_i$ in the dataset, we have the following constraint (left side of the equation corresponds to $c(\boldsymbol{\phi}, \boldsymbol{D}_i, \boldsymbol{\zeta})$ and right side to $q(\boldsymbol{\zeta})$ in Eq.~\ref{eq:objST}):
\begin{equation}
    \sqrt{(x^l_i - x^r_i)^2 + (y^l_i - y^r_i)^2 + (z^l_i - z^r_i)^2} = 2r
\end{equation}

According to \cite{Hollerbach2016}, this corresponds to restricting 1 position parameter. The parameters of the touched surface ($2r$)
are in this case known---same as for planar constraints with known plane parameters.

\subsection{Planar constraints -- one arm chain in contact with a plane (LA/RA)}
\label{section:planar_contact}

\begin{figure}[!h]
\centering
\begin{minipage}[c]{0.5\textwidth}
    \includegraphics[height=0.58\textwidth]{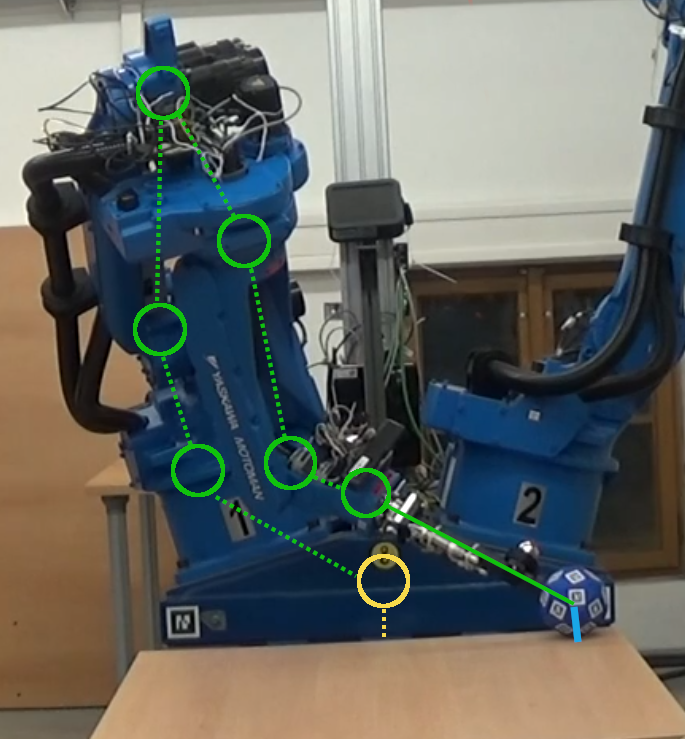}
        \includegraphics[height=0.58\textwidth]{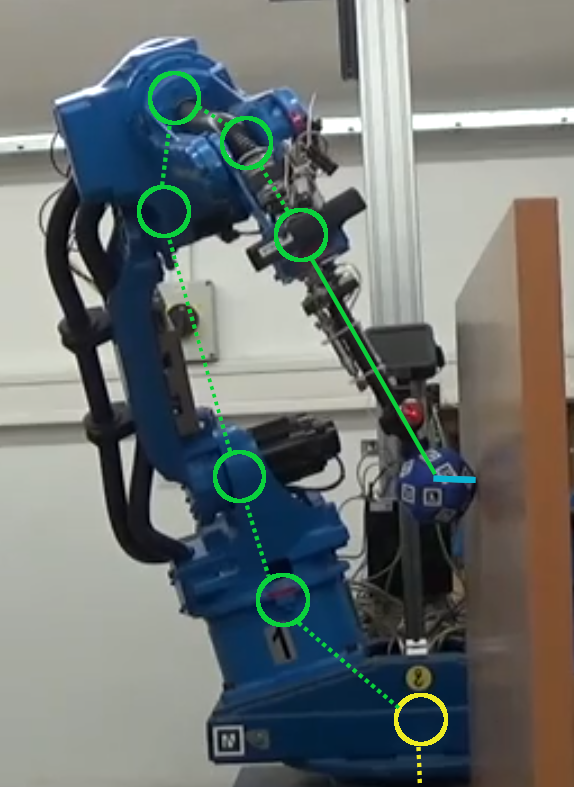}
  \end{minipage}\hfill
  \begin{minipage}[c]{0.5\textwidth}
    \caption{\label{fig:planar_touch_setup_no_cameras} Contact with planar constraints: horizontal plane (left), vertical plane (right).}
  \end{minipage}
\end{figure}

\begin{figure}
    \centering
    \includegraphics[width = .5\textwidth]{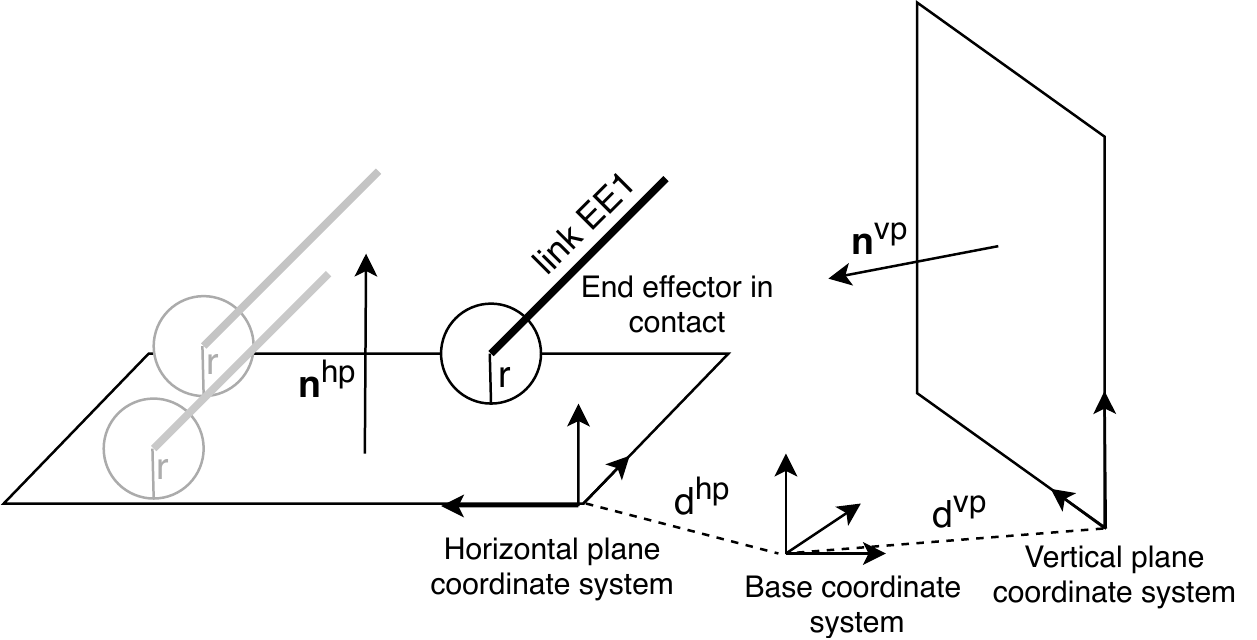}
    \caption{Visualisation of planes with their corresponding parameters ($\boldsymbol{n}^{hp}$, $d^{hp}$ and $\boldsymbol{n}^{vp}$, $d^{vp}$) and coordinate frames. End effector with radius $r$ in contact with horizontal plane in multiple places is shown.}
    \label{fig:plane_drawing}
\end{figure}

This corresponds to the scenario where the Left (LA) or Right (RA) robotic arm is getting into contact with a plane (see Fig.~\ref{fig:planar_touch_setup_no_cameras}). In this type of optimization problem, we can distinguish formulations including single-plane or multiple-plane constraints \cite{ Zhuang1999, Joubair2015b}. The classical formulations of the problem use either a general equation of the constraint plane or plane normals \cite{Ikits1997}.
The general equation of a plane is:
\begin{equation}
\label{eq:plane}
ax + by + cz + d = 0,
\end{equation}
where $\mathbf{n} = (a, b, c)$ is a plane normal vector. The parameters of the plane can be known in advance (as in \cite{Ikits1997}, \cite{zenha2018incremental} or \cite{Joubair2015} where calibration cube is used), or unknown (as in our case or \cite{Zhuang1999}). The planes and their corresponding parameters are visualized in Fig.~\ref{fig:plane_drawing}.

When the parameters of the plane are unknown, the parameter vector $\boldsymbol{\phi}$ consists of the following parameters: $\boldsymbol{\phi} = \{\boldsymbol{\phi}^{ra/la}, \mathbf{n}, d\}$, where $\boldsymbol{\phi}^{ra/la}$ are the DH parameters of the robot arm in contact, $\mathbf{n}$ is a plane normal, and $d$ is the distance of the plane from the origin. We formulated the objective function as the distances between contacts and a single or multiple fitted planes:
\begin{multline}
\boldsymbol{g}^{p}(\boldsymbol{\phi}^p, \boldsymbol{D}^p, \boldsymbol{\zeta}) = \left[ \begin{matrix}\mathbf{c}(\boldsymbol{\phi}^{hp1}, \boldsymbol{D}^{hp1}, \boldsymbol{\zeta}) - r ,\end{matrix}\right. \\
        \left.\begin{matrix}  \; \mathbf{c}(\boldsymbol{\phi}^{hp2}, \boldsymbol{D}^{hp2},\boldsymbol{\zeta}) - r, \mathbf{c}(\boldsymbol{\phi}^{vp}, \boldsymbol{D}^{vp}, \boldsymbol{\zeta}) - r\end{matrix}
        \right],\ \ \ \ \ \ \ \ \ \ \ \ 
        \label{eq:objplanarCon}
\end{multline}
where $\boldsymbol{D}^p \subset \boldsymbol{D}^{whole}$, $\boldsymbol{D}^p =\{\boldsymbol{D}^{hp1}, \boldsymbol{D}^{hp2}, \boldsymbol{D}^{vp}\}$ is a set of datasets where contacts between the end effector and lower horizontal plane ($\boldsymbol{D}^{hp1}$), higher horizontal plane ($\boldsymbol{D}^{hp2}$) or vertical plane ($\boldsymbol{D}^{vp}$) were performed. The set $\boldsymbol{\phi}^p =\{\boldsymbol{\phi}^{hp1}, \boldsymbol{\phi}^{hp2}, \boldsymbol{\phi}^{vp}\}$ is a set of parameters for lower horizontal plane ($\boldsymbol{\phi}^{hp1}$), higher horizontal plane ($\boldsymbol{\phi}^{hp2}$), and vertical plane ($\boldsymbol{\phi}^{vp}$), respectively. The vector $\mathbf{c}(\mathbf{\boldsymbol{\phi}}^j, \boldsymbol{D}^j, \boldsymbol{\zeta})$ is a vector of distances between individual end effector positions and the given plane j for each datapoint $\boldsymbol{D}^j_i$ from the given dataset $\boldsymbol{D}^j$. The distance is computed using plane normals and corresponding plane coordinates as follows: $c(\boldsymbol{\phi}^j, \boldsymbol{D}^j_i, \boldsymbol{\zeta}) = ||\boldsymbol{n}^j\boldsymbol{p}^j_i(\boldsymbol{\phi}^{j,ra/la}) + d ||$.
Point $\mathbf{p}^j_i$ is the centre of the end effector computed by forward kinematics from dataset point $\boldsymbol{D}^j_i$; $\boldsymbol{\phi}^{j,ra/la}$ is the estimated parameter vector corresponding to the DH parameters of the touching arm; $\mathbf{n}^j$ $ = [a\ b\ c]$ is the plane normal; $d^j$ is the distance of the plane from the origin. These plane parameters ($\mathbf{n}^j$ and $d^j$) are estimated at each iteration of the optimization process based on current point coordinates estimates by SVD method as described below. The $\boldsymbol{\zeta}$ wraps up all other necessary parameters. 

To acquire parameters of the fitted plane in each iteration, the measured points are converted to homogeneous coordinates and their centre of gravity is computed and subtracted from all points. Afterwards, Matlab function \emph{SVD} is called. The singular vector corresponding to the smallest singular value is set as a normal of the plane.  
Parameter $d$ in Equation \ref{eq:plane} is calculated from: \begin{equation}
d = -ax_0 - by_0 - cz_0,
\end{equation} where $ (x_0, y_0, z_0) $ are coordinates of the points center.

Let $\boldsymbol{n} = [a,b,c]$ be a plane normal, $d$ the distance of the plane from the origin, $(x^r_i, y^r_i, z^r_i)$ the centre of the right arm end effector computed from forward kinematics for data point $\boldsymbol{D}_i$ with the given joint configuration $\theta_i$, and $r$ the radius of the icosahedron.  Then for each data point $\boldsymbol{D}_i$ in the dataset $\boldsymbol{D}^{hp/vp}$ we get the following constraint (left side of the equation corresponds to $\mathbf{c}(\boldsymbol{\phi}^{vp/hp}, \boldsymbol{D}^{vp/hp}, \boldsymbol{\zeta})$ in Eq.~\ref{eq:objplanarCon}):
\begin{equation}
    \sqrt{(a x^r_i + b y^r_i + c z^r_i + d)^2} = r
\end{equation}
As per \cite{Hollerbach2016}, this type of calibration corresponds to the restriction of 1 position parameter. Compared to the self-contact setup, in the case of planes with unknown parameters (our case), new calibration parameters (parameters of the plane $\boldsymbol{n}$ and $d$) have to be added.

\subsection{Self-observation by cameras (LA-LEye, LA-REye, RA-LEye, RA-REye)}
\label{section:self_observation}
\begin{figure}[!h]
\centering
\begin{minipage}[c]{0.25\textwidth}
    \includegraphics[width=1\textwidth]{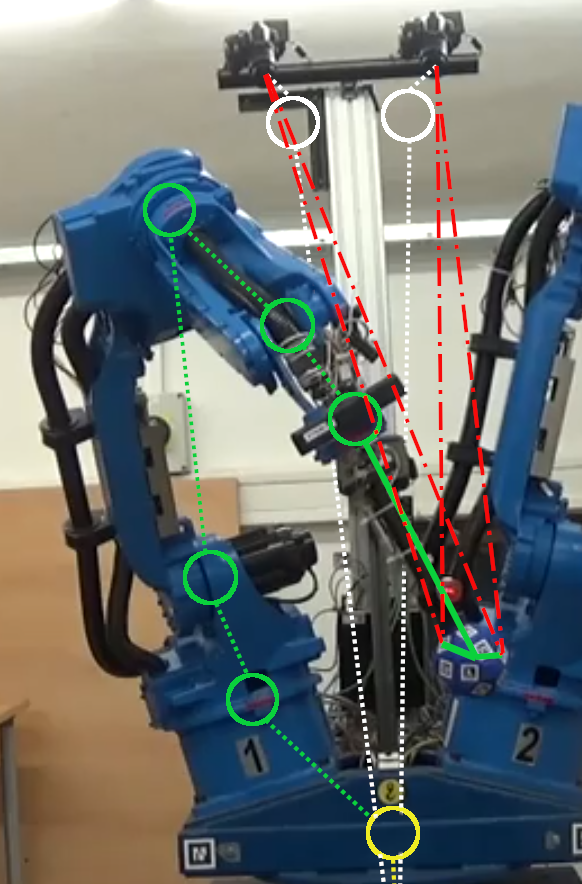}
  \end{minipage}\hfill
  \begin{minipage}[c]{0.23\textwidth}
    \caption{\label{fig:self_observation_setup} Self-observation chains: Red denotes reprojection from individual AruCo markers to left and right camera.}
  \end{minipage}
\end{figure}

This corresponds to the scenario where we observe the Left Arm (LA) or Right Arm (RA) end effector with AruCo markers via Left Camera (LEye) or Right Camera (REye) (see Fig.~\ref{fig:self_observation_setup}). We calibrate: (i) extrinsic parameters of the cameras (in our case as DH links) while assuming the robot DH parameters to be known, or (ii) the whole kinematic chain of the robot arm simultaneously with camera extrinsic parameters. In this case, the optimization is done by minimizing the reprojection error between the observed AruCo markers' positions in the camera and the estimated position using the current estimated kinematic model.

To obtain projected marker positions (which determine end effector position) in each of the robot cameras, we apply the calibrated camera model. The camera was precalibrated using a standard camera intrinsic calibration (see Section~\ref{section:camera_calib}). The extrinsic parameters (in our case expressed in DH parameters) of the camera were precalibrated based on reprojections of AruCo markers using the fixed nominal parameters of the arm. First, we have to transform marker positions to the camera frame: 

$$\begin{bmatrix}x_c&y_c&z_c&1\end{bmatrix}^T = \boldsymbol{T}_{m}^{camera} \cdot \begin{bmatrix}0 &0 &0 & 1\end{bmatrix}^T,$$
where $[x_c,y_c,z_c,1]^T$ are homogeneous coordinates of marker position in the frame of the given camera and $\boldsymbol{T}_{m}^{camera}$ is a transformation from the marker $M_m$ to the given camera achieved through standard approach for forward kinematics using DH parameters. 
Afterwards, we apply a standard pinhole camera model extended with radial and tangential distortion coefficients and transform the 3D point in camera frame ($[x_c,y_c,z_c]$) into image coordinates $[u,v]$ (2D plane of the camera) (see camera model in Sec.~\ref{section:camera_calib}). 
The actual marker position means the center of the ArUco marker. The OpenCV function \textit{calibrateCamera} provides calibration with resolution of whole pixels. ArUco marker detection is done by the  OpenCV function \textit{cv2.aruco.detectMarkers}, which provides the coordinates of all four marker corners. From these, the center of the marker is calculated as an intersection of the diagonals connecting the marker corners. 
The error resulting from this assumption is smaller than the calibration error~\cite{puciow_thesis}. 

The parameter vector ${\boldsymbol{\phi}}$ consists of the following parameters: ${\boldsymbol{\phi}} = \{{\boldsymbol{\phi}}^{ra/la},\boldsymbol{\phi}^{rc/lc}\}$, where ${\boldsymbol{\phi}}^{ra}$, $\boldsymbol{\phi}^{la}$, $\boldsymbol{\phi}^{rc}$, and $\boldsymbol{\phi}^{lc}$ are calibrated DH parameters corresponding to the right arm, left arm, right camera, and left camera, respectively. The objective function is formulated as the distance between projected markers and their pixel coordinates in the images:

\begin{multline}
\boldsymbol{g}^{so}(\boldsymbol{\phi}, \boldsymbol{D}^{so}, \boldsymbol{\zeta}) = \left[ \begin{matrix}p(\boldsymbol{\phi}, \boldsymbol{D}_1, \boldsymbol{\zeta}) - {z}(\boldsymbol{D}_1), ...,\end{matrix}\right. \\
        \left.\begin{matrix}  p(\boldsymbol{\phi}, \boldsymbol{D}_M', {\zeta}) - {z}(\boldsymbol{D}_M')\end{matrix}
        \right]
\label{eq:objSO}
\end{multline} 
where $p(\boldsymbol{\phi}, \boldsymbol{D}_i, \boldsymbol{\zeta})$ is the reprojection of marker $m_i$ from dataset point $\boldsymbol{D}_i$ where $i \in \{1,...,M'\}$, $M'$ is the length of the dataset $\boldsymbol{D}$. 
The ${z}(\boldsymbol{D}_i)$ is the actual marker position in the camera image.

This approach does not require information from both eyes and enables us to estimate only one side of the robot body (e.g., parameters of the left arm and left camera). For example, the estimated parameter vector $\boldsymbol{\phi}$ in the case of the kinematic chain connecting left arm and left camera consists of the following parameters: ${\boldsymbol{\phi}} = \{{\boldsymbol{\phi}}^l,{\boldsymbol{\phi}}^{lc}\}$, where  ${\boldsymbol{\phi}}^l$ and ${\boldsymbol{\phi}}^{lc}$ are parameters corresponding to the robot left arm and to the left camera, respectively. 

The conditions in the objective function can be expressed as follows: Let $\boldsymbol{x}^{m}$ be a centre of origin of marker $m_i$ (obtained by forward kinematics for the given DH parameters estimate), $\boldsymbol{x}^c$ be a marker position in the given camera frame, $\boldsymbol{T}^{EEF}_{m_i}$ be a transformation from marker $m_i$ to the arm end effector (transforms to individual ArUco markers can be found at the dataset page at \cite{dataset_our}) and $\boldsymbol{T}^{camera}_{EEF}$ transformation from the end effector to the given camera frame, $p(\boldsymbol{x^c})$ a reprojection of 3D point in camera frame to image coordinates $[u,v]$ (see Eq.~\ref{eq:projection}):
$$\begin{bmatrix}u\\v\\1\end{bmatrix} = p(\boldsymbol{x}^c) = p( \boldsymbol{T}_{EEF}^{camera} \boldsymbol{T}_{m}^{EEF} \cdot \boldsymbol{x}^{m}).$$

Then, for each data point $i$ (corresponding to position of marker $m_i$ in the configuration $\boldsymbol{\theta}_i$), reprojection of marker $m_i$ to camera frame $[u_i,v_i]$, and each measured marker position in the camera image $(u^{m}_i,v^{m}_i)$ we get two equations:
\begin{equation}
    \begin{split}
&u_i = u^{m}_i,\\
&v_i = v^{m}_i.
\end{split}
\end{equation}

According to \cite{Hollerbach2016}, this type of calibration corresponds to the restriction of 2 position parameters. Still, we have to add camera DH parameters (see Table~\ref{tab:mounting_dh_cam}) to parameters being calibrated to enable the reprojection of markers to the camera frame.

\subsection{Combining multiple chains (LA-RA-LEye, LA-RA-LEye-REye)}
\label{section:multichain_calibration}

\begin{figure}[!h]
\centering
\begin{minipage}[c]{0.25\textwidth}
    \includegraphics[width=1\textwidth]{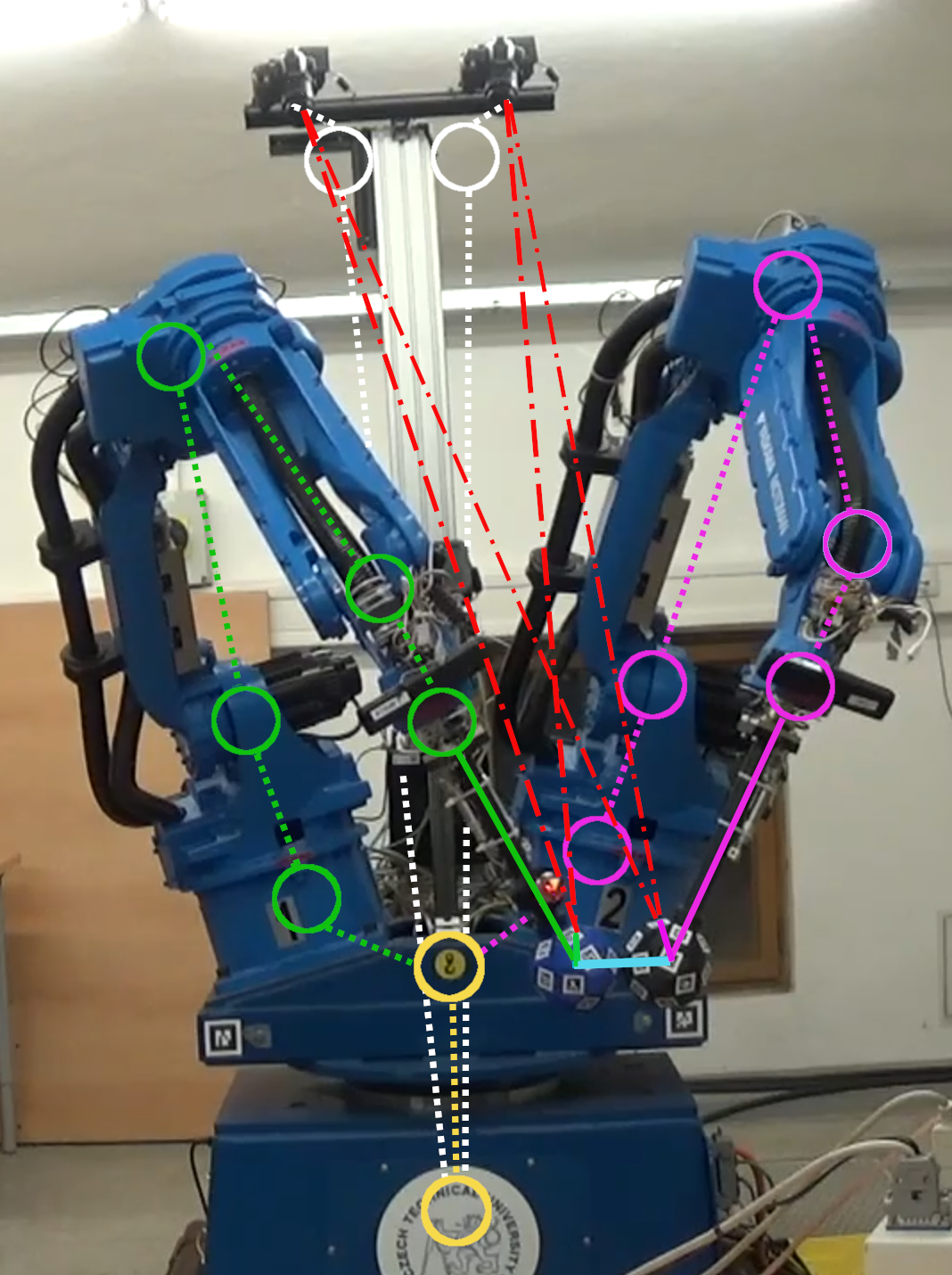}
  \end{minipage}\hfill
  \begin{minipage}[c]{0.23\textwidth}
    \caption[Calibration using different chains]{Illustration of  calibration combining multiple chains: self-contact and self-observation. All chains originate in a common base frame (bottom yellow circle). The left and right arm chains are drawn in purple and green, respectively. The eye chains are drawn in white. Red lines denote reprojection into the cameras. The cyan mark indicates the distance between end effector centers (one diameter).} \label{fig:calibrationMultiChains}
  \end{minipage}
\end{figure}

In order to estimate all kinematic parameters of the robot, we can take advantage of combining some or all of the above-mentioned kinematic chains. For example, in the case that we combine LA-RA, LA-LEye and LA-REye chains together into LA-RA-LReye (see Fig.~\ref{fig:calibrationMultiChains}), the estimated parameter vector $\boldsymbol{\phi}$ consists of the following parameters: ${\boldsymbol{\phi}} = \{{\boldsymbol{\phi}}^r,{\boldsymbol{\phi}}^l,{\boldsymbol{\phi}}^{rc},{\boldsymbol{\phi}}^{lc}\}$, where ${\boldsymbol{\phi}}^l$, ${\boldsymbol{\phi}}^r$, ${\boldsymbol{\phi}}^{rc}$,  and ${\boldsymbol{\phi}}^{lc}$ are parameters corresponding to the left arm, right arm, right camera, and left camera, respectively. Similarly, contact of right arm with a horizontal and vertical plane can be combined with self-observation by right camera, resulting in the parameter vector $\boldsymbol{\phi}$: ${\boldsymbol{\phi}} = \{{\boldsymbol{\phi}}^r,{\boldsymbol{\phi}}^{rc},\boldsymbol{n}^{hp},d^{hp}, \boldsymbol{n}^{vp},d^{vp}\}$, where $\boldsymbol{n}^{hp}$ and $d^{hp}$ are parameters defining the horizontal plane, and $\boldsymbol{n}^{vp}$ and $d^{vp}$ are parameters defining the vertical plane.

The overall objective function can be generally defined as (depending on which datasets and criteria we want to use for calibration): 
\begin{multline}
\label{eq:objCombined}
\boldsymbol{g}(\boldsymbol{\phi}, D, \boldsymbol{\zeta}) = \left[ \begin{matrix}\boldsymbol{k}^{st}\odot\boldsymbol{g}^{st}(\boldsymbol{\phi}, \boldsymbol{D}^{st}, \boldsymbol{\zeta}),\boldsymbol{k}^{p}\odot\boldsymbol{g}^p(\boldsymbol{\phi}, \boldsymbol{D}^p, \boldsymbol{\zeta}),\end{matrix}\right. \\
        \left.\begin{matrix}  \boldsymbol{k}^{so}\odot\boldsymbol{g}^{so}(\boldsymbol{\phi}, \boldsymbol{D}^{so}, \boldsymbol{\zeta})\end{matrix}
        \right],
\end{multline}
where $\boldsymbol{D}^{st}$, $\boldsymbol{D}^{p} = \{ \boldsymbol{D}^{hp1}, \boldsymbol{D}^{hp2}, \boldsymbol{D}^{vp}\}$ and $\boldsymbol{D}^{so}=\{\boldsymbol{D}^{st}, \boldsymbol{D}^{p}\}$ are datasets for self-touch, planar constraints optimization, and self-observation, respectively. 
Parameters $\boldsymbol{k}^{st}, \boldsymbol{k}^{p}$, $\boldsymbol{k}^{so}$ are scale factors to reflect the different uncertainty/reliability of the components, the number of measurements per configuration, and transformations from distance errors given in meters with the reprojection errors in pixels. Symbol $\odot$ marks a Hadamard product: i.e. ($\boldsymbol{k}^{st}\odot\boldsymbol{g}^{st})_i = k^{st}_i\cdot g^{st}_i$.
The value of these parameters is set independently for each pose:
$k^{st}_i = \eta^{st}_i \cdot p^{st}\cdot\mu_i$, $k^{p}_i = \eta^{p}_i \cdotp^{p}\cdot\mu_i$, and $k^{so}_i = \eta^{so}_i \cdot p^{so}$,
where $\eta^{st}, \eta^{p}$, $\eta^{so}$ reflect the reliability of the measurement (e.g., $\eta_i = \sigma_{i}^{-\frac{1}{2}}$, where $\sigma_i$ is the uncertainty of the measurement in the given pose). 
In this work, $\eta=1$ was used for all approaches. The parameter $p_i$ reflects the fact that there are multiple markers detected by cameras for the given contact configuration. Therefore, in the case of (planar constraints / self-touch) contact and self-observation combination $p^{p}=10$, $p^{st}=20$ (there are two icosahedrons in contact and on average 20 marker detections per contact event), and $p^{so}=1$.
The coefficient $\mu_i$ is determined from intrinsic parameters of cameras
($60\deg$ horizontal view angle, image size $4000\times6000\si{px}$) and distance $d_i$ of the end effector from the camera: $\mu_{i,x} = 4000\si{px}/(d_i (\pi/3))$, $\mu_{i,y} = 6000\si{px}/(d_i (\pi/3))$. For simplicity, $\mu_{i,x} = \mu_{i,y}$ was used. Figure \ref{fig:calibrationMultiChains} shows connections of different calibration chains and constraints (e.g., distance between end effectors during a self-contact or distance between end effector and a plane for a plane constraint). 

\subsection{Calibration using laser tracker}
\label{section:leica_calibration}

In this scenario, the robot arm with attached retroreflector is moving in free space to the configurations selected by sampling a joint space in the way that the retroreflector is facing the laser tracker (more details under Laser tracker experiment in Section~\ref{section:dataset_description}). The distance between the position of the retroreflector acquired by the laser tracker and the position computed from current robot arm DH parameters (plus current joint angle values and using forward kinematics) is minimized.

The parameter vector ${\boldsymbol{\phi}}$ consists of the following parameters: ${\boldsymbol{\phi}} = \{{\boldsymbol{\phi}}^{ra/la}, \boldsymbol{R}, \boldsymbol{T}\}$, where ${\boldsymbol{\phi}}^{ra/la}$ are DH parameters corresponding to the right/left robot arm (with the link EEL1/EEL2 to the retroreflector -- see Table \ref{tab:merged_dh}), $\boldsymbol{R}$ and $\boldsymbol{T}$ are rotation and translation matrices defining laser tracker position w.r.t. the robot base frame.

The objective function is formulated as the error of distances:
\begin{equation}
\boldsymbol{g}^L(\boldsymbol{\phi}, D, \boldsymbol{\zeta}) =\begin{bmatrix}p(\boldsymbol{\phi}, \boldsymbol{D}_1, \boldsymbol{\zeta}), ...,\; p(\boldsymbol{\phi}, \boldsymbol{D}_M, \boldsymbol{\zeta})\end{bmatrix},
\label{eq:obj_func_leica}
\end{equation}
where the function $p(\boldsymbol{\phi}, \boldsymbol{D}_i, \boldsymbol{\zeta}) = ||\boldsymbol{x}_i^L-\boldsymbol{x}_i^r||$ computes the distance of the transformed point $\mathbf{x}_i^L$ from laser tracker and the point $\mathbf{x}_i^r$ gained from forward kinematics and current estimate of robot DH parameters in the configuration given by the dataset point $\boldsymbol{D}_i$.

To calibrate the robot DH parameters, we minimize the distance between these 2 sets of 3D points (set $\mathbf{X}^L$ contains points from laser tracker in its coordinate system and set $\mathbf{X}^r$ includes points in the base coordinate system computed from a joint configuration using forward kinematics and the current estimate of robot DH parameters) using an iterative approach. 

In each iteration of the optimization process we:
\begin{enumerate}
    \item recompute the estimate of robot DH parameters
    \item recompute rotation and translation matrix defining laser tracker position w.r.t. base frame. 
    The relation between corresponding points in sets is generally:
\begin{equation}
\label{eq:fitsets}
\mathbf{x}^r_{i} = \boldsymbol{R}\mathbf{x}^L_{i}+\boldsymbol{T}+N_i,
\end{equation}
where $\boldsymbol{R}$ and $\boldsymbol{T}$ are rotation and translation matrices defining laser tracker position w.r.t. the robot base frame, $N_i$ is noise for the $i$-th datapoint. We used an algorithm introduced by Arun et al.~\cite{Arun1987} for finding least-squares solution of $\boldsymbol{R}$ and $\boldsymbol{T}$. 
It is a non-iterative algorithm using the singular value decomposition of a 3$\times$3 matrix. 
\end{enumerate}

This is a standard open-loop calibration method in which 3 components of the pose are measured: the 3D position, not orientation, is acquired from the laser tracking system \cite{Hollerbach2016}. The transform ($\boldsymbol{R}$, $\boldsymbol{T}$) relating the robot base frame to the laser tracker frame of reference has to be added to calibration. Since we express rotation by rotation vector, this corresponds to adding $6$ parameters to calibration.

\subsection{Non-linear least squares optimization}
For solving the optimization problem, the Levenberg-Marquardt iterative algorithm was employed. This is a standard choice for kinematic calibration (e.g., \cite{bennett1991autonomous,Hollerbach1996,Birbach2015}) and combines the advantages of gradient descent and Gauss-Newton algorithms, which can be also applied to this problem. For implementation, we used the Matlab Optimization Toolbox and the nonlinear least-squares solver \emph{lsqnonlin} with the Levenberg-Marquardt option and parameters \emph{typicalx} and \emph{scaleproblem}.

\subsection{Observability and identifiability of parameters}
\label{section:multichain_observability}

According to \cite{sun2008}, the observability index measures the quality of the dataset based on the identification Jacobian matrix \textbf{J}, which represents the sensitivity of minimized values to the change of individual parameters. Borm and Menq~\cite{borm1989} proposed a measure $O_1$; Driels and Pathre~\cite{driels1990} proposed $O_2$; Nahvi and Hollerbach proposed measures $O_3$ \cite{Nahvi1994} and $O_4$ \cite{nahvi1996}. All these measures can be computed from the singular value decomposition of \textbf{J}. They are defined as:
\begin{gather}
O_1 = \frac{(\sigma_1\sigma_2\ldots\sigma_m)^{1/m}}{\sqrt{n}}, \quad O_2 = \frac{\sigma_{min}}{\sigma_{max}},\nonumber \\\quad
O_3 = \sigma_{min}, \quad O_4 = \frac{\sigma_{min}^2}{\sigma_{max}}, 
\end{gather}
where $\sigma_j$ is $j$-th singular number, $m$ is the number of independent parameters to be identified and $n$ is the number of calibration configurations. 

The identification Jacobian matrix itself shows us the identifiability of individual optimized parameters: $\textbf{J}(i,j) =\frac{\partial X_i}{\partial \phi_j}$, where $X_i$ is a distance (Eq.~\ref{eq:objST} and Eq.~\ref{eq:objplanarCon}) or a reprojection error (Eq.~\ref{eq:objSO}) and $\phi_j$ is the parameter to be estimated. If a matrix column related to a parameter consists only of zeros, the parameter is unidentifiable which leads to an unobservable problem (the minimal singular number is zero). According to \cite{Hollerbach2016}, an unidentifiable parameter means that the experimental setup does not allow it to be identified, not that it is intrinsically unidentifiable. The identifiability can be improved by adding additional sensors to the setup as well as by extending the amount of poses in the dataset. In our analysis, we compare observability indices $O_1$ (representing the volume of a hyperellipsoid specified by singular numbers) and $O_4$ (noise amplification index which measures both eccentricity of the hyperellipsoid through $O_2$ and size of the hyperellipsoid through $O_3$) (see \cite{Hollerbach2016} for an overview) for individual chains and estimated parameter vectors. 

\subsection{Perturbation of the initial parameters estimate}
\label{section:perturbation}
To evaluate the dependence of the optimization performance on the quality of the initial estimates of the parameters, we perturbed all estimated parameters by a \textit{perturbation factor} $p = \{1,3,10\}$ (in experimental section, we show results for $p = 3$). We perturbed all initial offset values $o_i$ as follows:
\begin{equation}
o^{new}_i = 0.1p\cdot U(-1,1)+ o_i \: [rad],
\end{equation}
where $U(a,b)$ is uniform distribution between $a$ and $b$.
It is reasonable to expect that the remaining DH parameters ($\alpha$, $a$, and $d$) will be in general more accurate as they can be extracted from CAD models and there is no moving part and no encoder involved. Therefore, their perturbation was chosen as follows:
\begin{equation}
\begin{split}
&\alpha: \alpha^{new}_i =0.01p\cdot U(-1,1)+\alpha_i \: [rad],\\
&a, d: \Phi^{new}_i =  0.01p\cdot U(-1,1)+\Phi_i \: [m].\\
\end{split}
\end{equation}

\subsection{Evaluation}
\label{section:evaluation}
Before optimization, we randomly divided the dataset into training and testing data with a ratio of approximately 70 to 30. Training and testing datasets contained distinct sets of poses---multiple observations of markers in a given pose were all added to the same batch. Optimization was based on the training data; testing data were used to evaluate the result. We used root-mean-square (RMS) error to compare the results from multiple optimization runs. It is computed as:
\begin{eqnarray}
    RMS_c = \sqrt{\frac{1}{M}\sum _{i=1}^{M}(k_i^c g^c(\boldsymbol{\phi}, \boldsymbol{D}_i, \boldsymbol{\zeta}))^{2}} = \nonumber \\
    \sqrt{\frac{1}{M}\|\boldsymbol{k^{c}}\odot \boldsymbol{g^{c}}(\boldsymbol{\phi},\boldsymbol{D}^{c},\boldsymbol{\zeta})\|^2},
    \label{eq:rmse}
\end{eqnarray}
where $M$ is the number of observations/measurements, $k^c$ is a scale factor for the given calibration approach $c$ (see Sec.~\ref{section:multichain_calibration}) and $g^c(\boldsymbol{\phi}, \boldsymbol{D}^c_i, \boldsymbol{\zeta})$ is the corresponding objective function (see Sec.~\ref{section:self_contact}, \ref{section:planar_contact}, \ref{section:self_observation} or \ref{section:leica_calibration} for specific form of individual objective functions). Symbol $\odot$ marks a Hadamard product: i.e. ($\boldsymbol{k}^{st}\odot\boldsymbol{g}^{st})_i = k^{st}_i\cdot g^{st}_i$.

\subsection{Accuracy of measuring individual components}
\label{section:accuracy_components}
The accuracy of several components constitutes a lower bound on the overall accuracy. In particular, we estimated or experimentally evaluated the accuracy of the following (\textit{we note in brackets for which chains is the accuracy of the given component relevant}):
\begin{itemize}
    \item \textbf{3D printed parts and their dimensions}: $0.1\si{mm}$ error based on the printer specification; we assume the end effector rod to be straight. (\emph{self-contact (Sec.~\ref{section:self_contact}) and contact with a plane chains (Sec.~\ref{section:planar_contact})})
    \item \textbf{Intrinsic camera calibration}: we evaluated error of intrinsic camera calibration using multiple calibration patterns split to training and testing dataset. The resulting error is $0.73$ mean reprojection error for calibration points in pixels on the testing set. This error is composed of the accuracy of detecting the dot pattern and the calibration itself. (\emph{self-observation chains (Sec.~\ref{section:self_observation})})
    \item \textbf{Pose extraction from ArUco markers}: The calibration object with the Aruco markers was mounted on a linear positioning table. The object was moved along straight lines with a 0.01 mm precision. We captured images of the object in different positions and detected centers of the observed Aruco markers as shown in Fig.~\ref{fig:arucoDetection}. Due to camera reprojection, the markers move along lines in the undistorted image (radial undistortion). We estimate the error of marker detection as distance of the detected markers from interpolated straight line (Fig.~\ref{fig:arucoDetection}). The mean error is around $0.33$ pixels for well visible markers (ID 107, 112, 113) and $0.63$ pixels for other markers with maximal error of $1.5$ pixels. We investigate the error along the line with similar results. Intrinsic camera calibration puts a lower bound on these results. (\emph{self-observation chains (Sec.~\ref{section:self_observation})}) 
    
    \begin{figure}
        \centering
        \includegraphics[width = 0.4\textwidth]{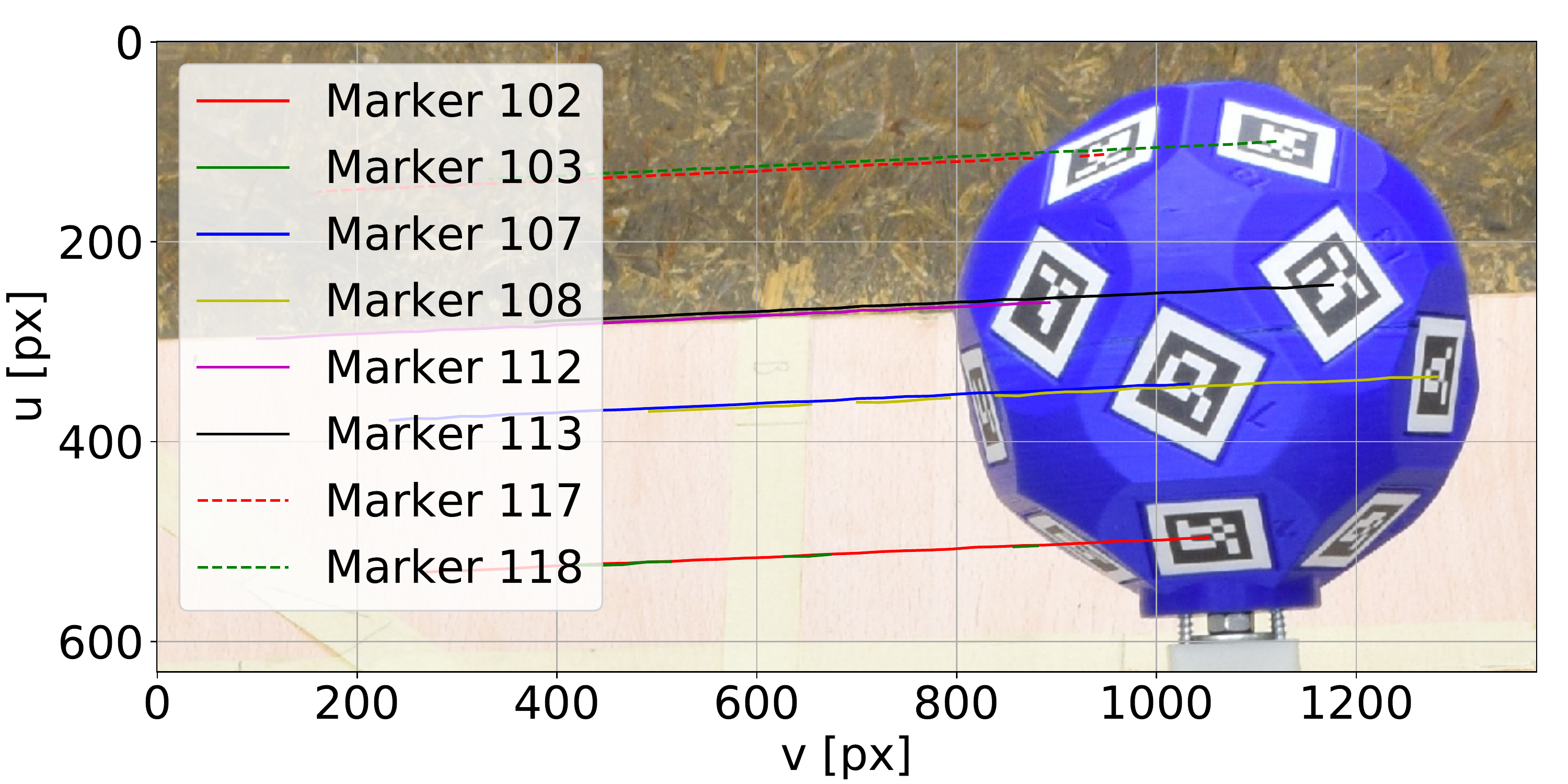}
        \includegraphics[width = 0.5\textwidth]{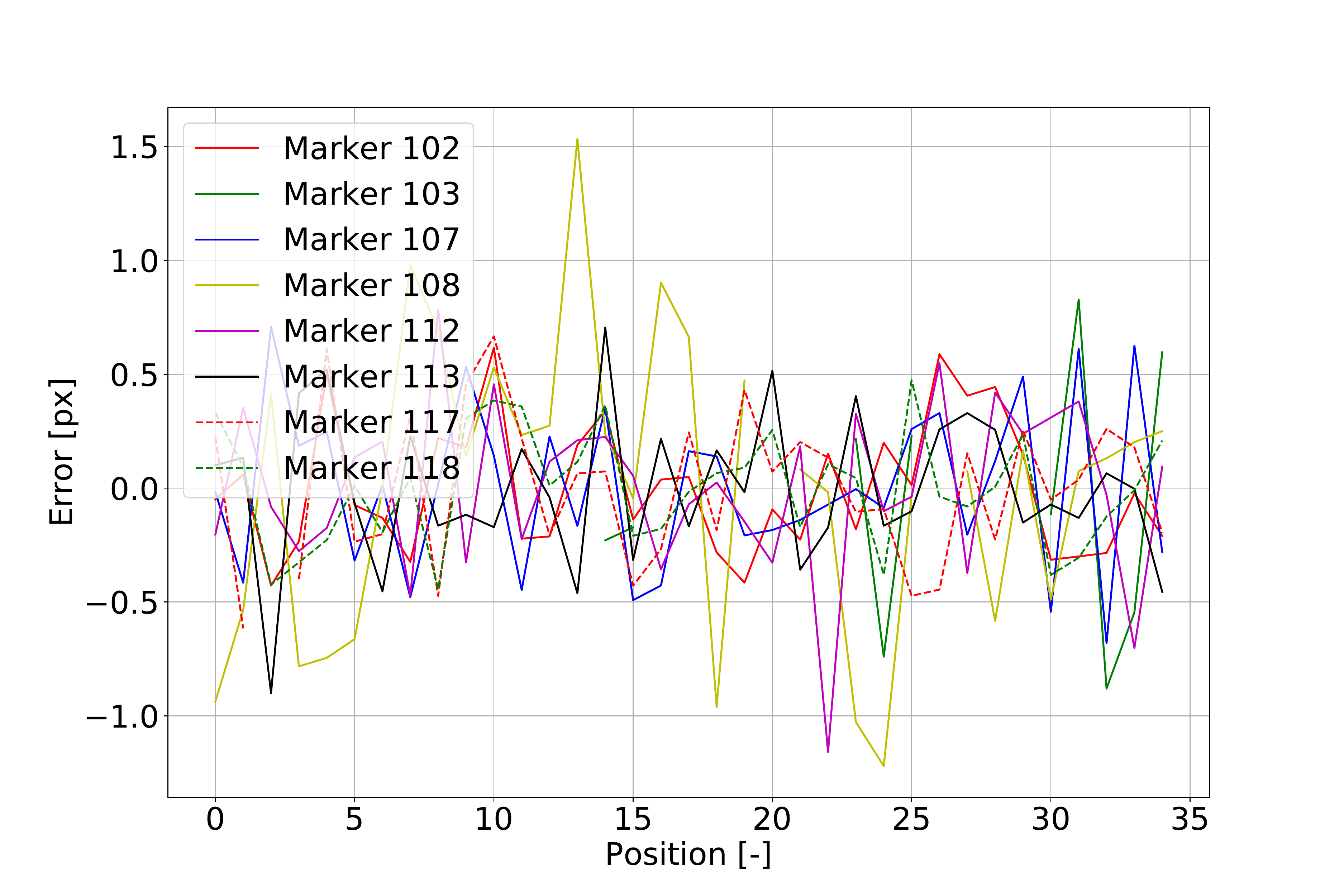}
        \caption{Pose extraction from ArUco markers. (top) Icosahedron moving along a straight line with markers positions reprojected to camera frame ($u$ and $v$ are coordinates of the camera image). (bottom) Errors measured as perpendicular distance from the line for individual markers at specific positions.}
        \label{fig:arucoDetection}
    \end{figure}
    \item \textbf{Repeatability of measurements}: 
    The details of robot movements and how they were stopped once contact was detected are described in Sec.~\ref{subsec:robot_control}. We used the laser tracker to measure the repeatability of these movements, using 20 repetitions of the same movement at two different positions sampled from the grid for planar constraints (Fig.~\ref{fig:datasetPlanar}). These involved only small robot movements (10 cm) and repeatability was high -- Fig.~\ref{fig:repeatability}. For self-touch, larger movements from the robot home position were executed. However, self-contact experiments involved both large movements and small movements similar to those for contact with a plane. Thus, the statistics for Fig.~\ref{fig:repeatability} was combined for the self-touch distribution, resulting in overall lower repeatability.
    Results in $x$-coordinate are shown. Similar distribution and range of errors was observed for $y$ and $z$ coordinate. 
    \emph{(all chains)}

    \item \textbf{Camera projection error propagation}: Cameras measure projections to the image plane, producing higher uncertainty in the z-axis of the image coordinate system (image depth).
    The uncertainties in all three axes can be obtained by error propagation from object position uncertainty. The projection uncertainty can be written as a quadric equation $u^TQu = 1$, where $Q$ is a matrix (4x4 identity matrix assumed here) and $u$ is a 4x1 vector of both cameras' projection pixel coordinates. 
    The projections vector $u$ can be expressed as $u = J_xX$, where $J_x$ is the Jacobian matrix computed from projection equations (Eq.~\ref{eq:projection}) and  $X$ are the projected point coordinates. Combining these two equations, we obtain the equation for position uncertainty $X^T Q_x X = 1$, where $Q_x = J_x^T Q J_x$, which can be interpreted as an ellipsoid. Then, the eigenvalues of $Q_x$ are $a^{-2}$, $b^{-2}$, $c^{-2}$, where $a$, $b$, $c$ are half the principal axes' length in meters. In our case, the mean values are $a = 0.146~\si{mm}$, $b = 0.150~\si{mm}$, $c = 1.330~\si{mm}$, i.e. an error of 1 pixel corresponds to more than 1 mm error in object position in one axis. The comparison of uncertainties can be seen in Fig.~\ref{fig:repeatability}. The length of the last axis of the ellipsoid (image depth) is one order of magnitude higher than the other lengths and the touch uncertainty. (\emph{self-observation chains (Sec.~\ref{section:self_observation})})
    
    \begin{figure}[!h]
    \centering
    \includegraphics[width = 0.48\textwidth]{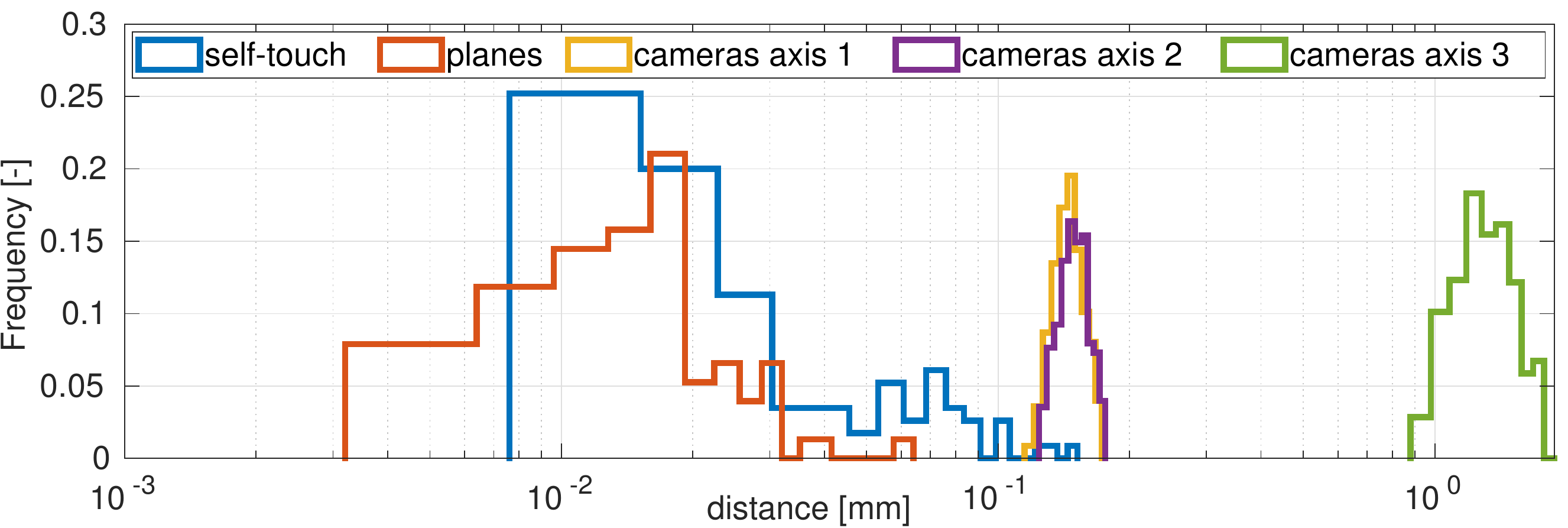}
    \caption{Error distribution of different components. Data for self-touch and planes obtained from repeatability experiments. Camera errors from camera resolution and 3D space projection from robot workspace.}
    \label{fig:repeatability}
\end{figure}{}
\end{itemize}

    Based on the above-mentioned analysis, we can see that the lower bound of error for self-observation will consist of the combination of intrinsic camera calibration error ($e^i$) and ArUco marker detection error ($e^{ArUco}$): $e^{so}$ = $e^i \bigoplus e^{ArUco}$. The measured displacement of the end effector between contacts is influenced by $e^{ArUco}$ and the observed error is systematic. In the worst case, we can estimate the error to be independent on the direction of the impact and for all directions consider the maximum observed error. As can be seen, repeatability measurements show that the accuracy is very high and the errors are below 0.03 mm. Finally, an analysis of camera errors in 3D space reveals their limited accuracy, in particular in one axis corresponding to the depth.

\section{Experimental results}
\label{section:results}
To evaluate and compare the results of kinematic parameters calibration across individual approaches and their combinations, we show results for the right arm of the robot only---this kinematic chain can be calibrated using all the datasets collected. First, we show the results for end effector length and angle offset (Section~\ref{sec:ee_calibration}) without and with camera calibration. Afterward, we show results of ``daily calibration'' (calibrating only joint angle offsets $\boldsymbol{o}$) of the whole right robot arm (including length of the end effector), consisting of: reference values acquired by the laser tracker (Sec.~\ref{section:leica_calibration_results}), calibration by individual approaches without and with camera calibration (Sec.~\ref{sec:calibration_offsets}).  
Finally, the results for all DH calibration are shown in Sec.~\ref{sec:calibration_all_dh} and evaluated on the independent laser tracker dataset.

The following notation/labeling of experiments is used for the individual evaluated approaches: \textit{self-contact/self-touch} (Sec.~\ref{section:self_contact}); \textit{1 horizontal plane, 2 horizontal planes, all planes} (2 horizontal and vertical plane) (Sec.~\ref{section:planar_contact}); \textit{all} (combination of all planes and self-contact calibration, Sec.~\ref{section:multichain_calibration}); \textit{self-observation} (Sec.~\ref{section:self_observation}). 

Each dataset (see Sec.~\ref{section:dataset_description} for details) was split into training and testing part (70:30). The resulting RMS errors (according to Eq.~\ref{eq:rmse}) for both training and testing datasets are shown and compared to the case where nominal parameters are used (\emph{before}). We show resulting RMSE separately for self-contact distances between icosahedrons (\emph{'dist'}), distances to plane (\emph{'plane dist'}), and for camera reprojections in pixels (\emph{'mark.'}) (e.g., Fig.~\ref{fig:rmse_eef}).
In addition, we investigate corrections to the parameter values and their mean and variance.

Calibration methods relying on physical contact---self-contact and planar constraints---can be employed independently or in combination with calibration using cameras (self-observation). We distinguish three possibilities that will be used throughout the rest of the Experimental results section:
\begin{itemize}
    \item \textbf{No cameras.} Only contact-based calibration was employed and cameras were  not used at all.
    \item \textbf{Fixed cameras.} Cameras' DH parameters were precalibrated (see Table~\ref{tab:mounting_dh_cam}) using the contact-based datasets (self-contact and contact with plane) from marker reprojection errors, using nominal values of the robot kinematic parameters. Then,  to calibrate the robot right arm, reprojection of markers to the camera frame is used together with contact information, while the camera extrinsics stay fixed.
    \item \textbf{Camera position calibration (Cam. calib.).} Cameras were  precalibrated in the same way as in \textit{Fixed cameras}. During robot kinematic calibration, reprojection errors are combined with contact-based constraints. Additionally, camera extrinsics (expressed in DH parameters) are subject to calibration as well.
\end{itemize}

\textit{Self-observation (S-O)}, i.e., information from camera reprojection errors, can also be employed independently.
  The datasets used are those featuring contact ($D^{st}$, $D^{hp}$, $D^{whole}$, etc.), but the constraints arising from physical contact are not employed.

Table~\ref{tab:calib_params} provides an overview which parameters can be subject to calibration under the different approaches. The parameters that can be calibrated using the contact-based approaches are marked in black. Fixed cameras enable calibration of the end effector orientation ($o_{EE1}$, in green; contact information alone does not provide any information about orientation -- see Figures \ref{fig:st_drawing} and \ref{fig:plane_drawing}). Camera extrinsic parameters, which may be also subject to calibration, are shown in blue.

\begin{table}[h]
\centering
\begin{tabular}{p{1.8cm}|p{6.2cm}}
Calibration & Calibrated parameters \\ \hline
End effector & $d_{EE1},$  \textcolor{green}{$o_{EE1}$} \\ \hline
Offsets &  $o_{L1},$ $o_{U1},$ $o_{R1},$ $o_{B1},$ $d_{EE1},$ \textcolor{green}{$o_{EE1}$}\\ \hline
Offsets by ext. device & $o_{L1},$ $ o_{U1},$ $ o_{R1},$ $ o_{B1},$ $ a_{EEL},$ $ d_{EEL},$ $ o_{EEL}$ \\ \hline
All DH parameters & $a_{L_1},$ $ d_{L_1},$ $o_{L1}$, $a_{U_1},$ $ d_{U_1},$ $\alpha_{U_1},$ $  o_{U1},$ $a_{R_1},$ $ d_{R_1},$ $\alpha_{R_1}, $ $ o_{R1},$ $a_{B_1},$ $ d_{B_1},$ $\alpha_{B_1},$ $ o_{B1},$ $a_{T1},$ $ \alpha_{T1},$ $d_{EE1},$ \textcolor{green}{$o_{EE1}$} \\ \hline
Camera calibration & \textcolor{blue}{$a_{TT3},$ $d_{TT3},$ $\alpha_{TT3},$ $o_{TT3},$ $d_{C1},$ $o_{C1},$ $a_{TT4},$ $d_{TT4},$ $\alpha_{TT4},$ $o_{TT4},$ $d_{C2},$ $o_{C2}$} 
\end{tabular}
\caption{Overview of calibrated parameters. Parameters in green are added to the corresponding approach under \textit{fixedCameras}. When using \textit{calibCameras}, all the blue parameters are added. See Tables \ref{tab:merged_dh} and \ref{tab:mounting_dh_cam} for details about DH parameters.
}
\label{tab:calib_params}
\end{table}

\subsection{Calibration of end effector length and joint angle offset}
\label{sec:ee_calibration}
First, we evaluated the ability to calibrate the length and joint angle offset of the last link (EE1/EE2) by individual approaches. The diameter of the final part of the custom end effector---assuming it is a sphere in this case as the contact is at the spherical tiles---is known. 
The parameter being calibrated in all cases is the end effector length $d_{EE1}$; with fixed cameras, the orientation, $o_{EE1}$ can be also calibrated.
In the case of planar constraints, the plane parameters $\{\boldsymbol{n}^{hp/vp}, d^{hp/vp}\}$ are estimated in addition.

In the self-contact scenario, we compared the case where we assume that the length and offset for the left arm end effector are known and the case where we calibrate both EE1 and EE2 with the assumption that both end effectors have the same length---this is necessary to avoid compensation of the length of one end effector by the other end effector. In Table~\ref{tab:eefCalib}, we show results for the case when both end effectors lengths are calibrated simultaneously. 

Table~\ref{tab:eefCalib}, the top part, summarizes the corrections to the nominal values of the length of the end effector achieved by individual calibration approaches. The most consistent estimates (mean corrections over 20 repetitions) across individual approaches are achieved with fixed precalibrated cameras, where the range of estimates is from $3.89~\si{mm}$ (all) to $4.84~\si{mm}$ (2 horizontal planes) and from $3.11~\si{mm}$ to $5.95~\si{mm}$ for the self-observation approach trained on three different datasets. 

\begin{table}
    \centering
    \setlength\tabcolsep{2pt}
    \begin{tabular}{c|c c c}
         \multicolumn{4}{c}{Calibrating end effector length -- corrections } \\ 
         $[$mm$]$ & No cameras & Cam. calib. & Fixed cameras \\
         \hline
         1 h. plane & 10 $\pm$ 4 & 12 $\pm$ 2 & 3.32 $\pm$ 0.26 \\
         2 h. planes & 14 $\pm$ 2 & 15 $\pm$ 2  & 4.84 $\pm$ 0.11 \\
         all planes & 13 $\pm$ 3 & 10.2 $\pm$ 0.2 & 4.34 $\pm$ 0.07 \\
         self-contact & 4.31 $\pm$ 0.05 & 4.21 $\pm$ 0.04 & 4.12 $\pm$ 0.04 \\
         all & 4.31 $\pm$ 0.04 & 4.25 $\pm$ 0.04 & 3.89 $\pm$ 0.04 \\
         S-O ($D^{hp}$) &  - & 8 $\pm$ 18 & 3.11 $\pm$ 0.24 \\
         S-O ($D^{planes}$) & - & 10.17 $\pm$ 0.25 & 4.38 $\pm$ 0.11 \\
         S-O ($D^{whole}$) & - & 6.25 $\pm$ 0.07 & 5.95 $\pm$ 0.10\\
    \end{tabular}
    \begin{tabular}{c|c c c}
         \multicolumn{4}{c}{Calibrating end effector orientation -- corrections} \\ 
         $[$rad$] 10^{-2}$ & No cameras & Cam.calib.& Fixed cameras \\
         \hline
         1 h. plane & - & 2.96 $\pm$ 0.11 & 4.25 $\pm$ 0.23 \\
         2 h. planes & - & 3.29 $\pm$ 0.09 & 5.76 $\pm$ 0.21  \\
         all planes & - & 3.28 $\pm$ 0.09 & 3.69 $\pm$ 0.17  \\
         self-contact & - & 7.04 $\pm$ 0.09 & 1.06 $\pm$ 0.13 \\
         all & - &  6.34 $\pm$0.08 & 2.20 $\pm$ 0.08 \\
         S-O ($D^{hp}$) &  - & 3.06 $\pm$ 0.15  & 4.17 $\pm$ 0.38 \\
         S-O ($D^{planes}$) & - & 3.23 $\pm$ 0.13 & 3.60 $\pm$ 0.12 \\
         S-O ($D^{whole}$) & - & 6.40 $\pm$ 0.09 & 2.39 $\pm$ 0.09
    \end{tabular}
    \caption{End effector length calibration (corrections to the manually acquired values). Comparing calibration using plane constraints (h. plane -- 1 horizontal plane, 2 h. planes -- 2 horizontal planes, all planes), self-contact calibration, combination of planar constraints and self-contact (all) with self-observation only (on different datasets -- e.g., S-O ($D^{hp}$) -- self-observation calibration on a dataset from touching a horizontal plane). No cameras / Cam. calib / Fixed cameras -- see text. Mean and standard deviation over 20 repetitions is displayed.}
    \label{tab:eefCalib}
\end{table}

The quality of the individual estimates can also be evaluated with respect to the standard deviation (s.d.) across 20 repetitions. For planar constraints without cameras, we get a very high standard deviation (around $4~\si{mm}$) indicating poor estimates. On the contrary, using the self-contact approach even without cameras results in standard deviation around $0.05~\si{mm}$, which is comparable to the standard deviation achieved when camera information is included (either precalibrated cameras or cameras calibrated together with end effector length). The addition of cameras also improves consistency of end effector length estimates by planar constraints (standard deviation drops from $4$ to $0.2~\si{mm}$ for calibrated cameras and to $0.07~\si{mm}$ for fixed cameras). We also evaluated the calibration using only self-observation information. When we used precalibrated cameras on the whole dataset, we achieve comparable results to other estimates. When we calibrate the cameras together with end effector length, the length of the end effector is overestimated and the standard deviation (based on the used training data) is also significantly higher. 

We compare orientation corrections using only self-observation calibration (with cameras precalibrated on different subsets of datasets) or self-observation combined with other constraints (self-contact, planar constraints) -- see Table~\ref{tab:eefCalib}, bottom part. The corrections to manually measured parameters of the orientation vary between $0-0.06~\si{rad}$ $(0 - 3.4^\circ)$.

\begin{figure*}[!ht]
    \centering
    \includegraphics[width = \textwidth]{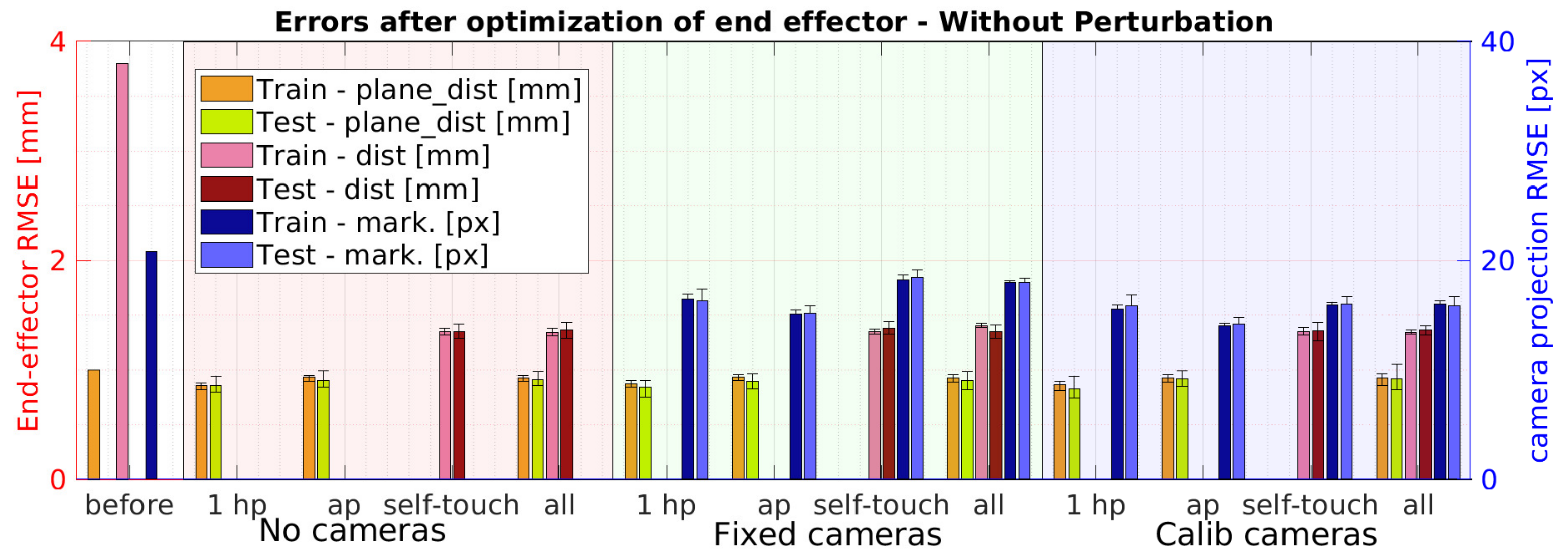}
    \caption{RMS errors -- end effector calibration. Distances in 3D (left y-axis, $mm$); camera reprojection (right y-axis, $px$). \textit{plane\_dist} -- Euclidean distance from a plane, \textit{dist} -- Eucl. distance between end effectors in self-contact, \textit{mark.} -- error of individual markers reprojected to cameras. \textit{No cameras} -- only planar constraints or self-touch; \textit{Fixed cameras} -- precalibrated but fixed cameras; \textit{Calib cameras} -- precalibrated cameras with additional optimization. Error on both training (\textit{Train}) and testing (\textit{Test}) dataset shown. \textit{before} -- nominal parameters, \textit{hp} -- 1 horizontal plane, \textit{ap} -- all planes (2 horizontal, 1 vertical), \textit{all} -- self-contact + planar constraints.} 
    \label{fig:rmse_eef}
\end{figure*}

The resulting RMS errors (see Section~\ref{section:evaluation}) for end effector calibration in$~\si{mm}$ and$~\si{px}$ are shown in Fig.~\ref{fig:rmse_eef}. 
For all variants of the optimization problem, both mean \emph{dist} RMSE and \emph{markers} RMSE decreased compared to the nominal end effector value (Fig.~\ref{fig:rmse_eef}). For the variant \textit{all} and \textit{Fixed cameras}, the distance RMSE decreased from $3.8~\si{mm}$ before calibration to $1.40~\si{mm}$ after calibration and markers RMSE decreased from $20.8~\si{px}$ before to $18.0~\si{px}$. Without cameras, we achieved slightly better results for self-contact distance RMSE ($1.34~\si{mm}$), but in this case the end effector orientation is not calibrated---resulting in higher error in camera reprojection. For the case with calibrated cameras, the resulting RMSE are slightly smaller (mean self-contact distance $1.34~\si{mm}$ and camera reprojection RMSE $16.0~\si{px}$) than for the case with fixed cameras. The calibration resulted in the estimation of end effector length and orientation---the corrections of these parameters are listed in the Table~\ref{tab:eefCalib}.

{}
For the end effector parameters, we do not have any reference value apart from manual measurements (laser tracker calibration cannot be applied due to the retroreflector placement -- see Fig.~\ref{fig:ee}). However, the corrections found using the whole dataset and the combination of all chains (self-touch + planar constraints + cameras) were reliable enough. Consistent adaptation was observed for the length. Hence, for subsequent calibration of the robot arm kinematics, we used the end effector length estimate of $35.4~\si{cm}$. For the orientation, no significant adaptation was arrived at and  the nominal value was kept for the remaining experiments. (see Table~\ref{tab:merged_dh}).

\begin{figure}[!h]
\centering
  \begin{minipage}[c]{0.25\textwidth}
    \includegraphics[width=\textwidth]{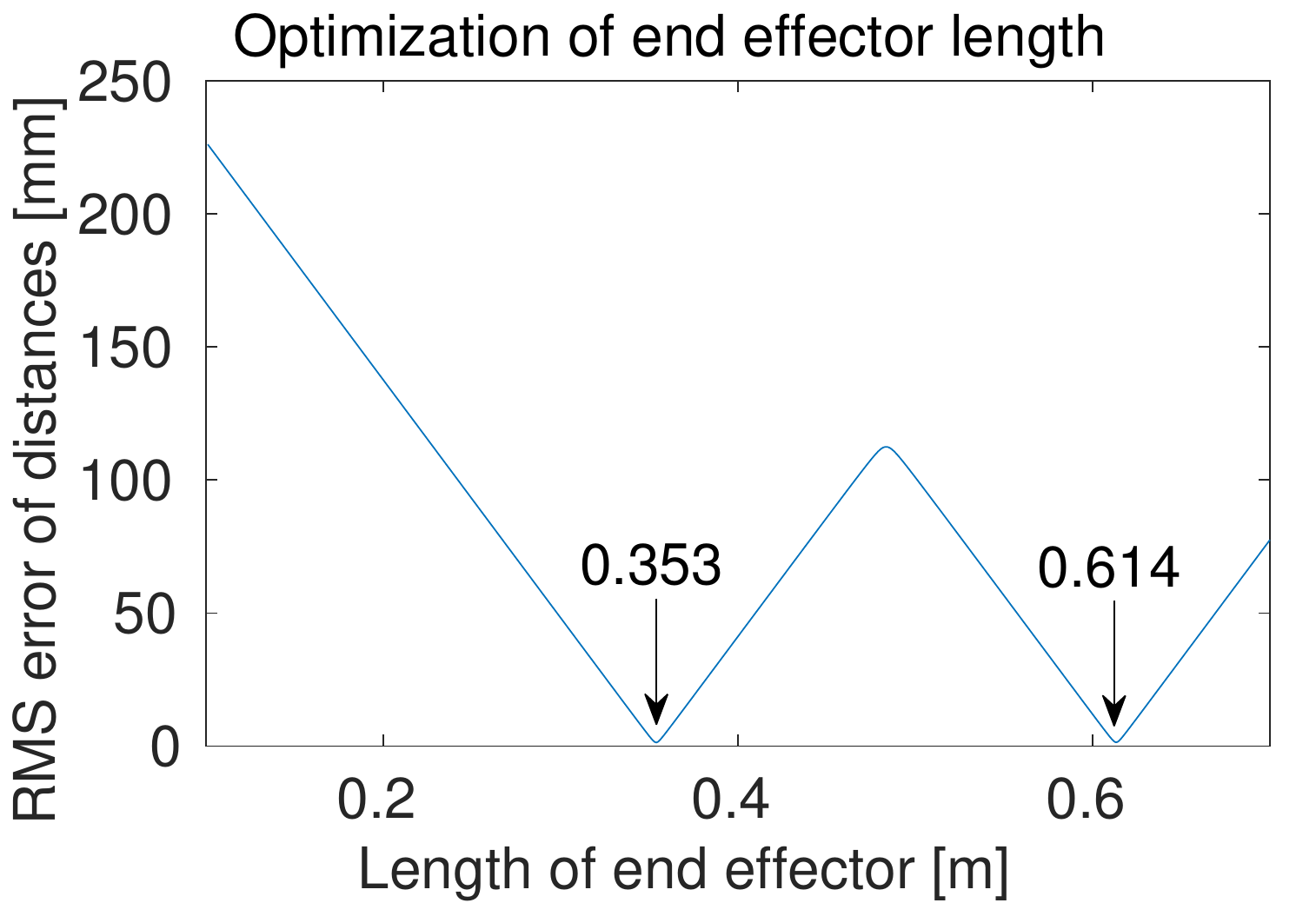} 
  \end{minipage}
  \begin{minipage}[c]{0.15\textwidth}\caption{RMS error on self-contact dataset for different values of end effector length.}
  \label{fig:selftouchCalib}
  \end{minipage}
\end{figure}

To validate the selection of the end effector length---which will be used in what follows as an initial value of this parameter---we performed two additional experiments. First, we systematically varied the initial end effector length parameter before calibration in the range from 10 to 70 mm and evaluated the RMS error after calibration on the self-contact dataset -- see Fig.~\ref{fig:selftouchCalib}. Two solutions were found with minimal RMS error ($35.38~\si{cm}$ and $61.4~\si{cm}$), but one of them is not to be considered, as we do not expect errors bigger than 1 cm from manual measurement (35 \si{cm}). These two solutions can be explained by the nature of the self-contact as there are multiple possibilities arising from the geometrical consideration of self-contact (see Fig.~\ref{fig:st_drawing}, right).

\begin{figure}[!h]
\centering
  \begin{minipage}[c]{0.15\textwidth}
    \includegraphics[height=20 em, width=\textwidth]{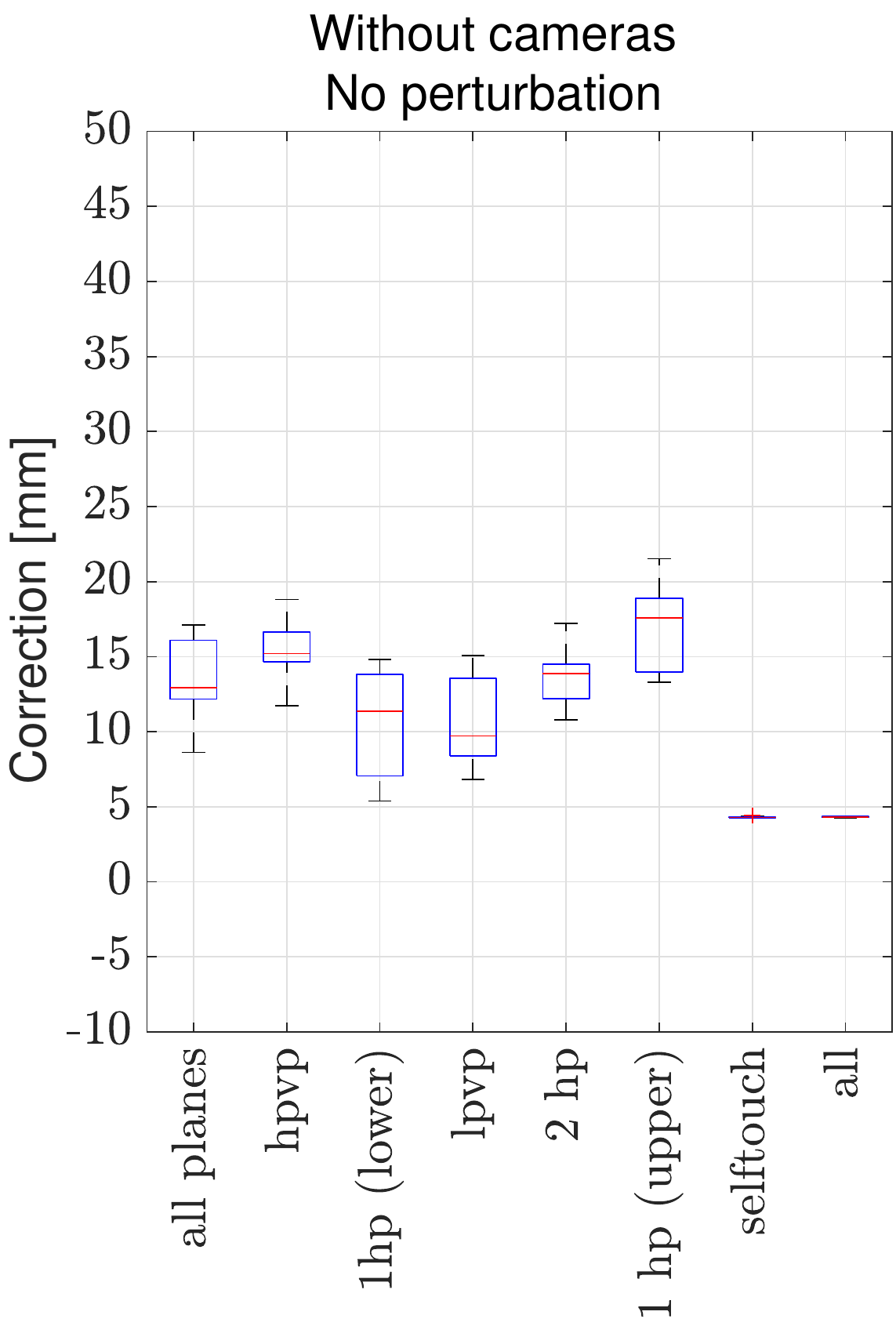} 
  \end{minipage}
  \begin{minipage}[c]{0.15\textwidth}
    \includegraphics[height=20 em, width=\textwidth]{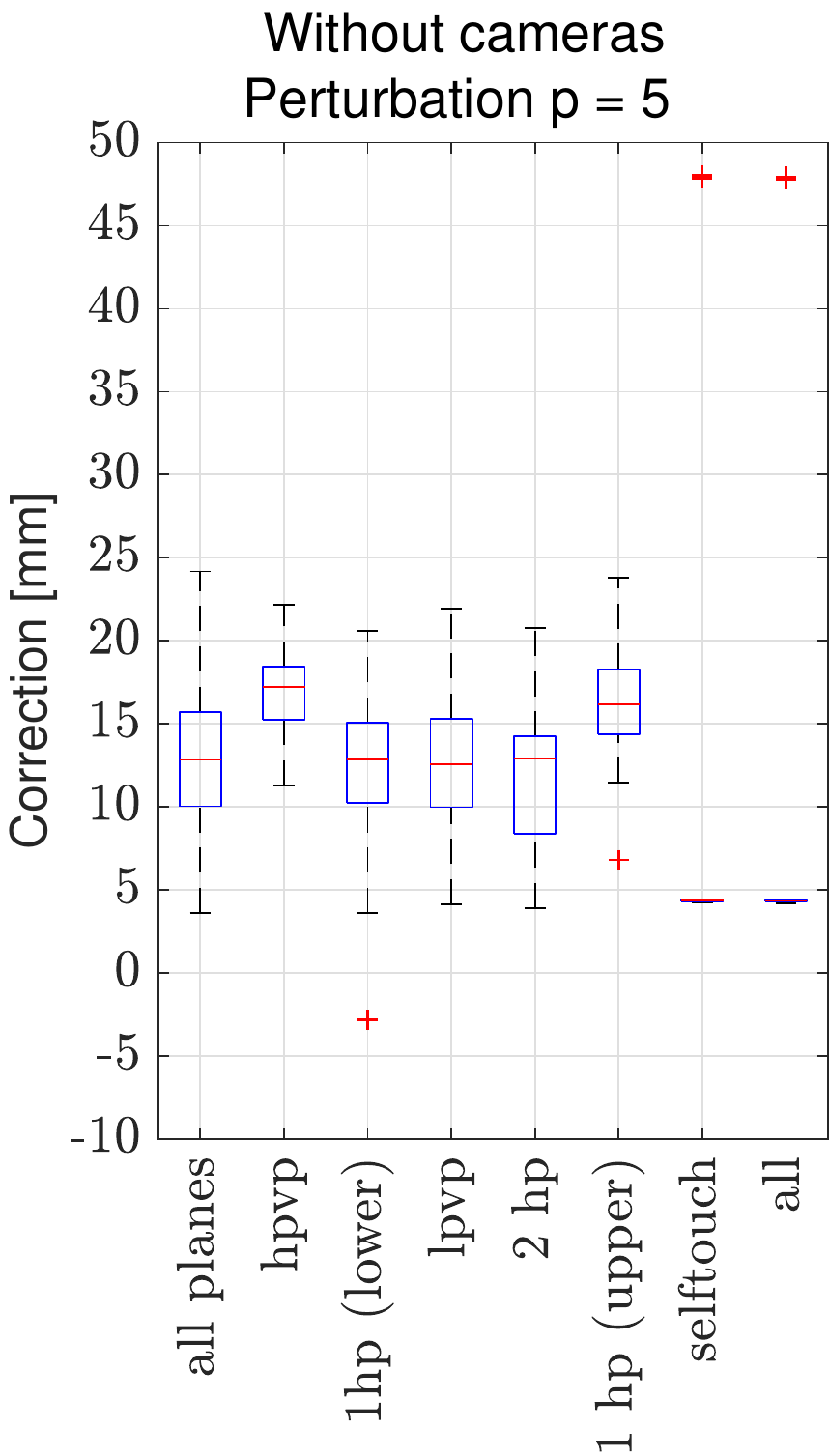} 
  \end{minipage}
  \begin{minipage}[c]{0.15\textwidth}
    \includegraphics[height=20 em, width=\textwidth]{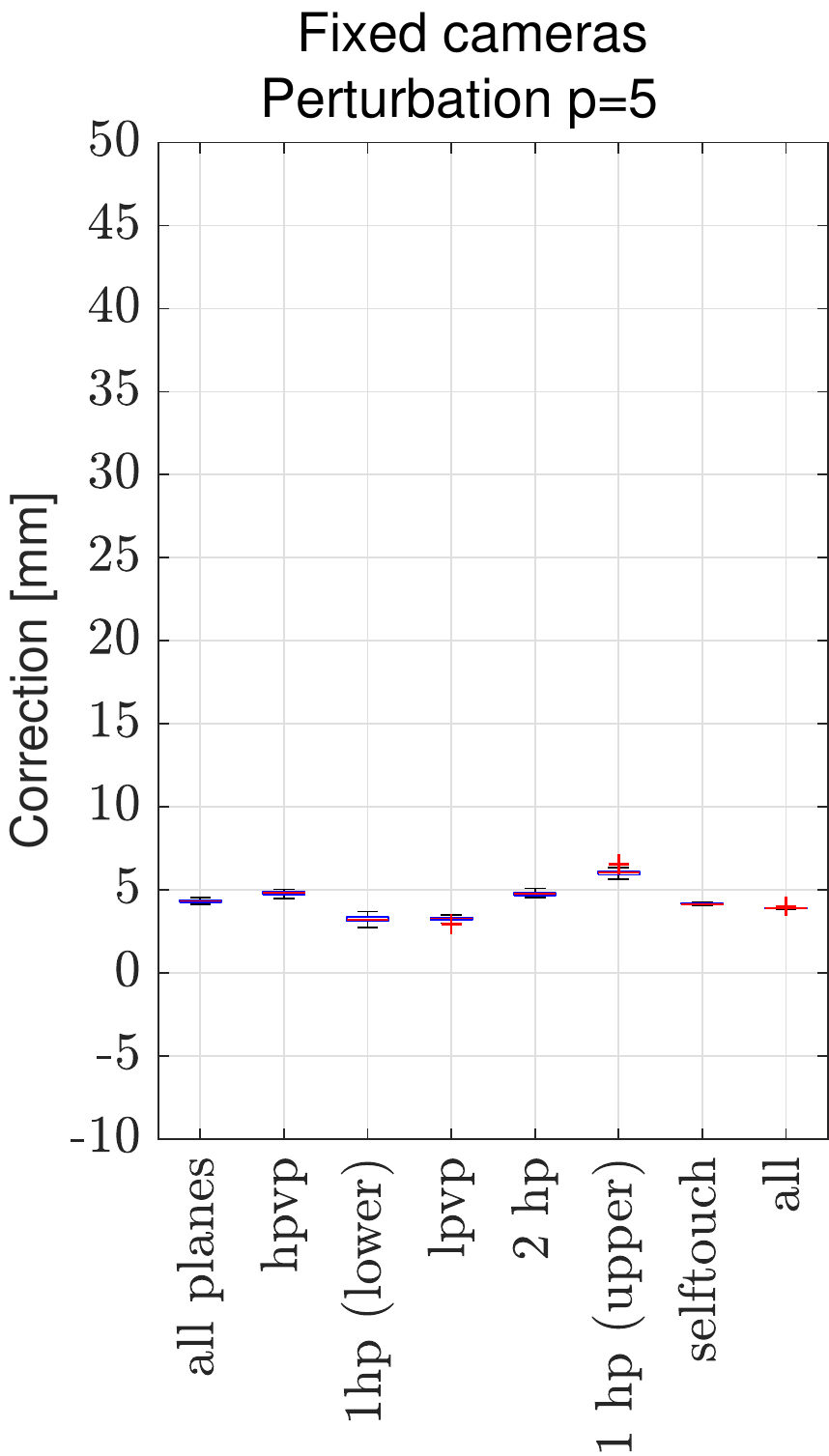} 
  \end{minipage}
  \begin{minipage}[c]{0.5\textwidth}\caption{Corrections of end effector length compared to nominal values: (left) no perturbation + no cameras, (centre) perturbation + no cameras, (right) perturbation + fixed cameras. \textit{all planes} -- 2 horizontal, 1 vertical; \textit{hpvp} -- upper horizontal + vertical plane; \textit{1hp (lower)} -- lower horizontal plane; \textit{lpvp} -- lower horizontal + vertical plane; etc.; \textit{all} -- self-contact + all planes}
  \label{fig:eefPert}
  \end{minipage}
\end{figure}

Second, the sensitivity to perturbation of the initial end effector length for individual approaches was evaluated in the case with/without cameras and compared to the case without perturbation. The resulting corrections can be seen in Fig.~\ref{fig:eefPert}. For the case with perturbation and without cameras (centre), two solutions were found by \emph{self-touch} calibration, corresponding to Fig.~\ref{fig:rmse_eef}. When fixed precalibrated cameras are added, we achieve results with a low standard deviation for all calibration setups; for \emph{self-touch} calibration, only one solution is selected.
For \emph{self-touch} and \emph{all} conditions with fixed cameras, we achieve the same results both for perturbed initial value and non-perturbed.

\subsection{Calibration of robot joint offsets by an external measurement device (reference)}
\label{section:leica_calibration_results}

The whole dataset $D^{tracker}$ collected by the laser tracker (Leica) was used for right arm joint angle offsets calibration to create a reference value to our other calibration approaches (see Sec.~\ref{section:leica_calibration}). Parameters subject to calibration are listed in Tab.~\ref{tab:calib_params}. First, we estimated the tracker position ($R$, $T$) and retroreflector DH parameters ($a$, $d$, and joint angle offset $o$) (link 7b in Table~\ref{tab:merged_dh}). Afterwards, we calibrated all offsets of the robot arm including the retroreflector DH parameters. 

\begin{figure}[!h]
\centering
\begin{minipage}[c]{0.24\textwidth}
    \includegraphics[width= \textwidth]{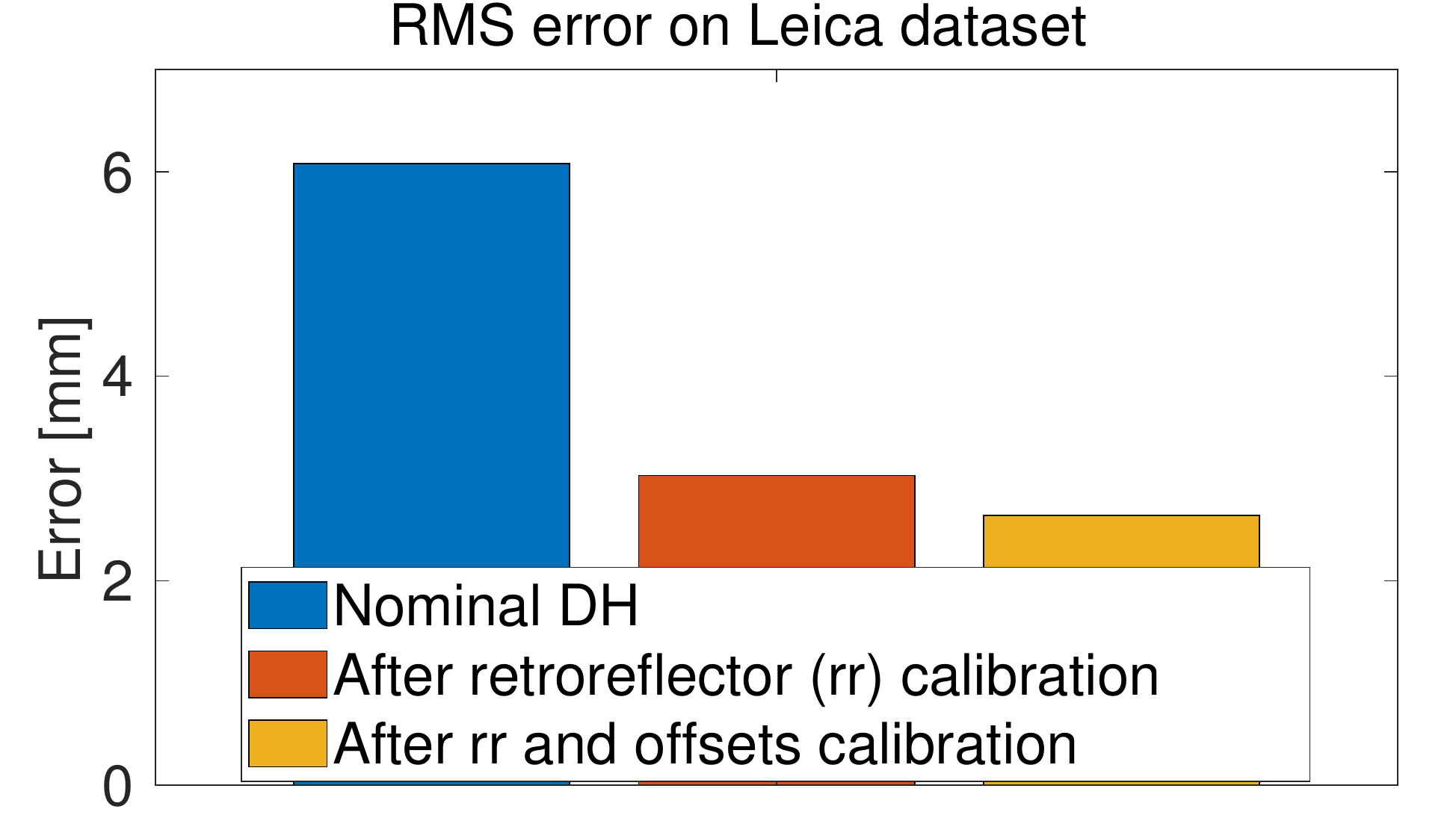}
  \end{minipage}\hfill
    \begin{minipage}[c]{0.24\textwidth}
    \includegraphics[width= \textwidth]{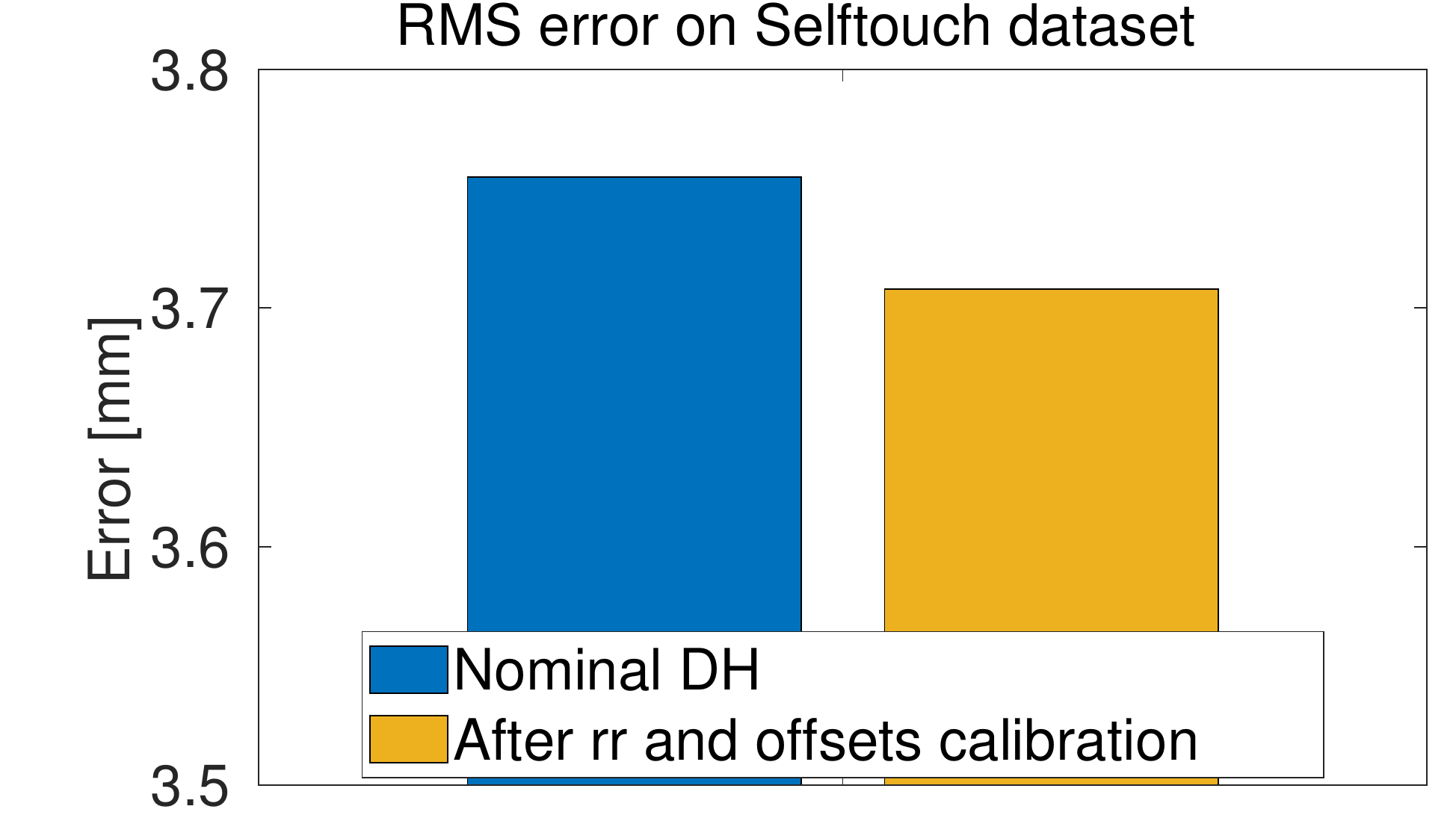} 
  \end{minipage}
  \begin{minipage}[c]{0.5\textwidth}\caption{(left) RMS errors comparison on the laser tracker dataset (\textit{Nominal DH} -- nominal parameters are used), after only retroreflector calibration, and after all right arm offsets including all DH parameters of the tracker retroreflector. (right) RMS errors comparison on self-contact dataset; parameters obtained by laser tracker calibration compared to RMS error achieved by Nominal DH parameters.}
  \label{fig:leicarmsBoxplot}
  \end{minipage}
\end{figure}

\begin{figure}[!h]
\centering
\begin{minipage}[c]{0.23\textwidth}
    \includegraphics[height=9 em]{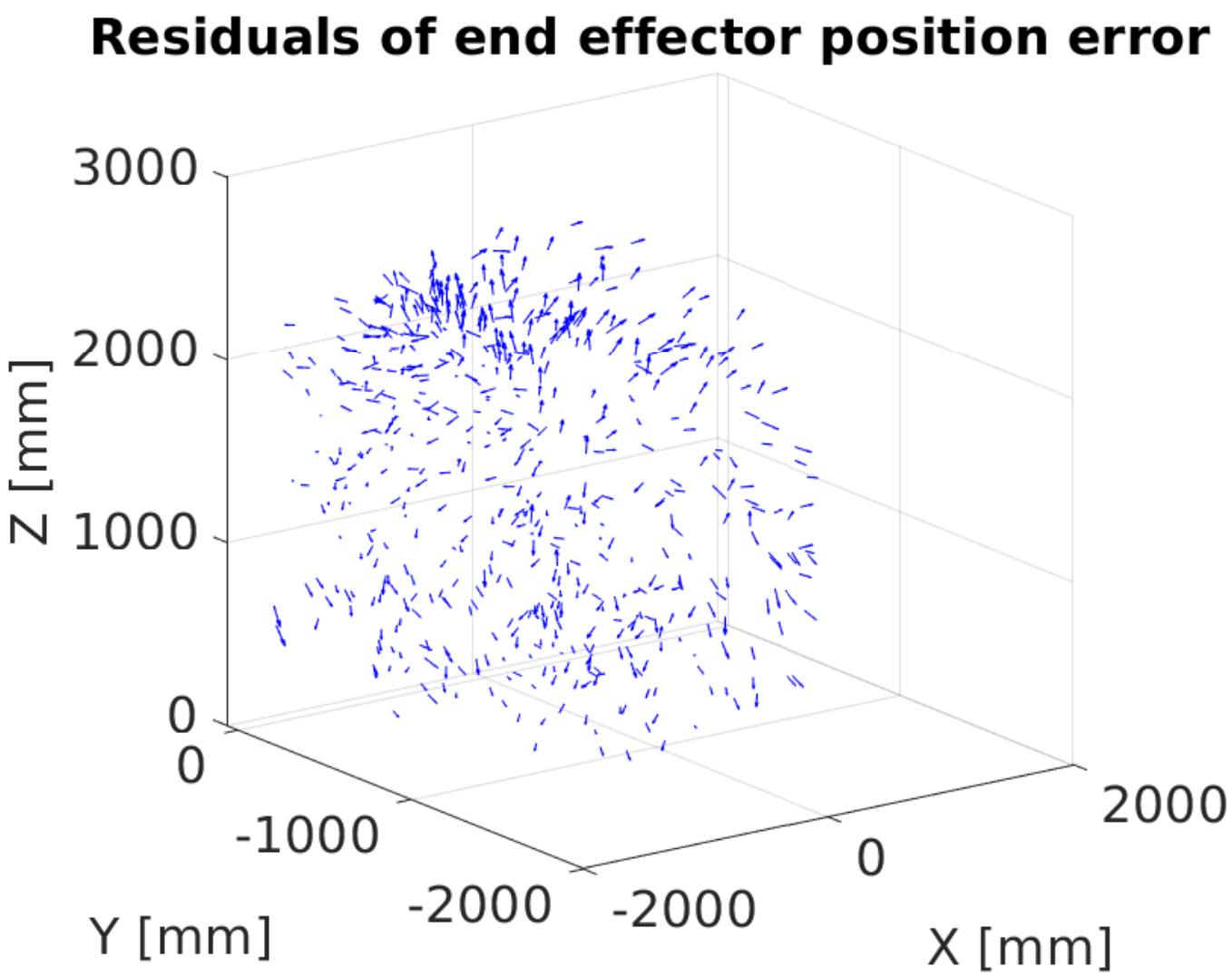}
  \end{minipage}\hfill
    \begin{minipage}[c]{0.23\textwidth}
    \includegraphics[height=9 em]{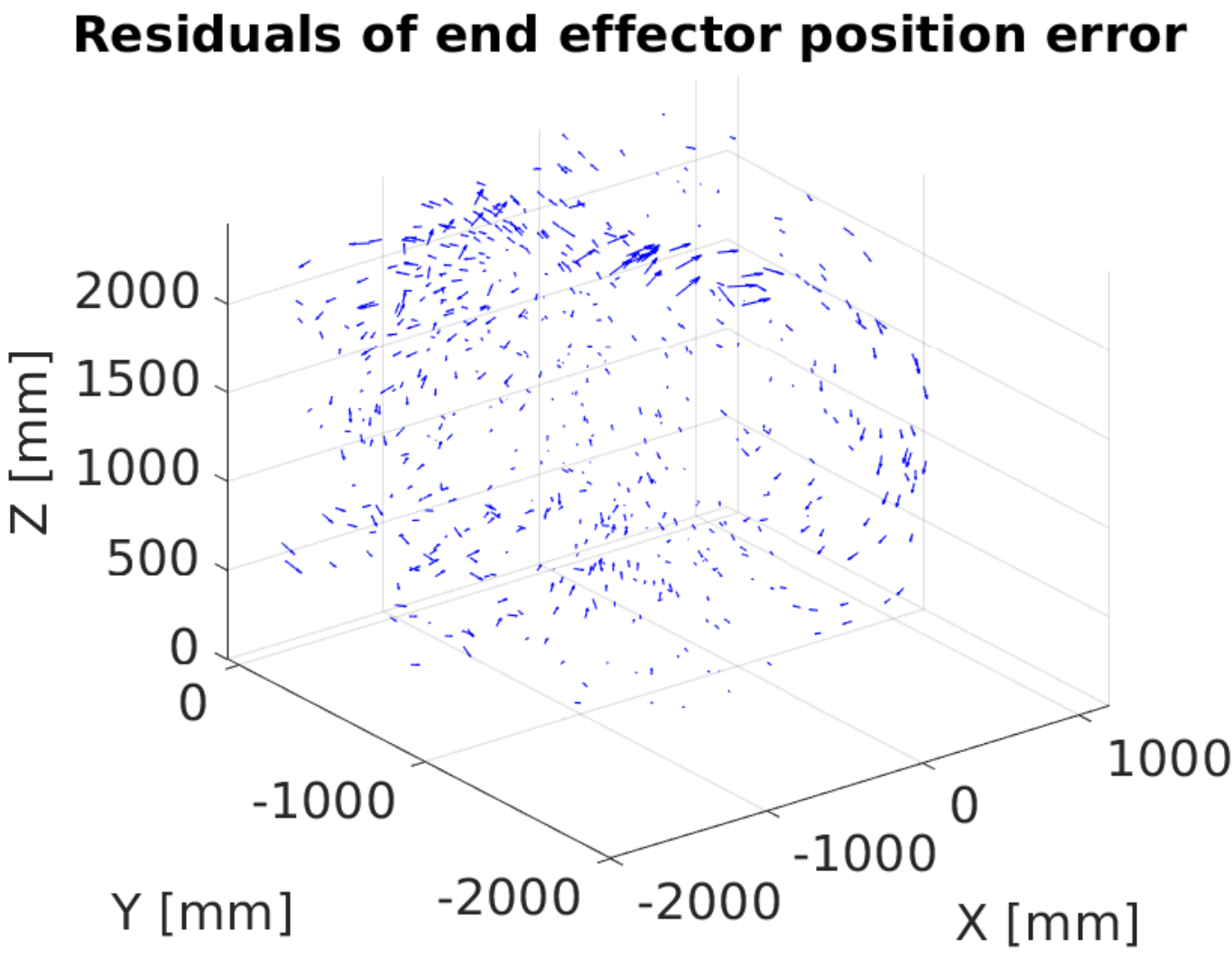} 
  \end{minipage}
  \begin{minipage}[c]{0.5\textwidth}\caption{Residual errors for individual datapoints before calibration using nominal parameters (left) and after calibration of all offsets of robotic arm including the tracker retroreflector full DH parameters (right). }
  \label{fig:leicarmsArrows}
  \end{minipage}
\end{figure}

As can be seen in Fig.~\ref{fig:leicarmsBoxplot}, left,  RMSE improved with further calibration: from $6.08~\si{mm}$ before calibration to $3.03~\si{mm}$ after retroreflector calibration and to $2.64~\si{mm}$ after all offsets calibration. To achieve the best possible solution quality, we have not split the original dataset into training and testing part in this case. Instead, we evaluated the RMSE on a separate dataset (self-contact dataset $D^{st}$) using the parameters estimated by the laser tracker calibration and compared to RMSE for nominal parameters (Fig.~\ref{fig:leicarmsBoxplot}, right). In this case, we used same parameters of the end effector for both methods and compared the error at the end of the $5$th link (see Fig.~\ref{fig:leicarmsBoxplot}). A slight improvement of RMSE can be observed also on this dataset: from $3.755~\si{mm}$ to $3.708~\si{mm}$ with the same uncalibrated end effector. The distribution of errors on the laser tracker dataset before and after calibration can be seen in Fig.~\ref{fig:leicarmsArrows}.

The corrections for retroreflector parameters (compared to nominal parameters) $\{d_{EEL1}$, $a_{EEL1}\}$ are $0.72~\si{mm}$, and $0.92~\si{mm}$, respectively, and for offset parameters $\{o_{L1}$, $o_{U1}$, $o_{R1}$, $o_{B1}$, and $o_{ELL1}\}$ are $\{-1.43$, $-0.18$, $-0.59$, $-1.03$, and $4.123\}$~\si{mrad}, respectively.

\subsection{Calibration of robot joint offsets -- self-contained approaches}
\label{sec:calibration_offsets}
We compared the quality of the right arm joint angle offsets calibration, including the end effector length, by individual approaches. Self-contact approach (Sec.~\ref{section:self_contact}),  planar constraints (using 1-3 planes; Sec.~\ref{section:planar_contact}) and pure self-observation calibration (Sec.~\ref{section:self_observation}) were compared to the results for nominal parameters and the reference parameters acquired by the laser tracker measurement device (Sec.~\ref{section:leica_calibration_results}). For the contact-based methods (self-contact, planar constraints),  we compared the results with no camera information, fixed precalibrated cameras, and cameras being part of the calibration process. Parameters subject to calibration are listed in Tab.~\ref{tab:calib_params}. Depending on the particular method, additional parameters may be subject to calibration (e.g., parameters of the plane) (see corresponding subsections in Sec.~\ref{sec:multichain}). 

\begin{figure}[!h]
    \centering
    \includegraphics[width = 0.48\textwidth]{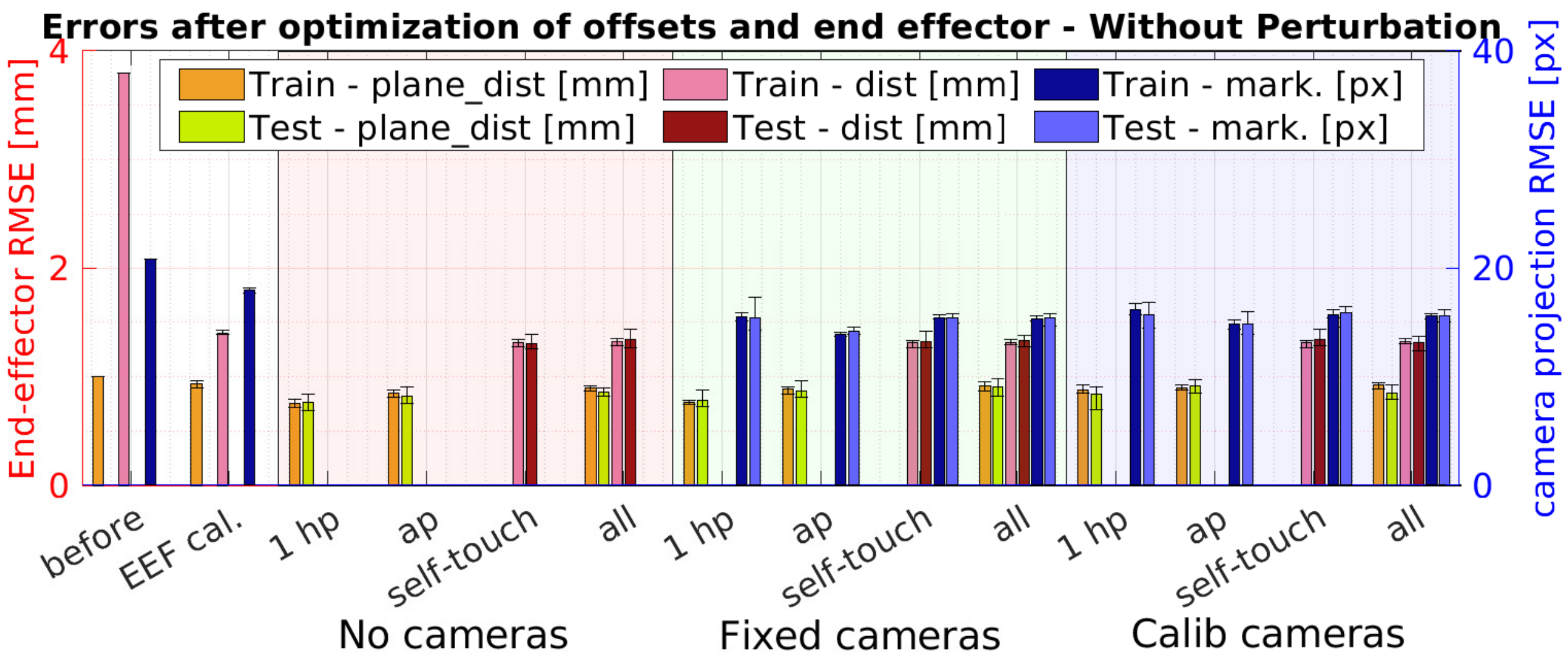}
    \caption{RMS errors -- right arm offsets and end effector calibration. Distances in 3D (left y-axis, $mm$); camera reprojection (right y-axis, $px$). \textit{before} -- nominal parameters, \textit{EEF cal.} -- after end effector calibration. For detailed legend see Fig.~\ref{fig:rmse_eef}. }
    \label{fig:rmse_offsets}
\end{figure}

RMSE after calibration of offsets and end effector length without perturbation can be seen in Fig.~\ref{fig:rmse_offsets}.
The RMS error of distance in self-contact (in $\si{mm}$) drops from $3.80~\si{mm}$ before calibration to $1.40~\si{mm}$ after end effector calibration, and to $1.31~\si{mm}$ after all offsets calibration. Distance to plane (in $\si{mm}$) drops from $1.00~\si{mm}$ before calibration to $0.93~\si{mm}$after end effector calibration, and to $0.89~\si{mm}$ after all offsets calibration without cameras ($0.91~\si{mm}$ with camera calibration and $0.92~\si{mm}$ with fixed cameras). Distance of reprojected markers in camera (in pixels) drops from $20.8~\si{px}$ before calibration to $18.0~\si{px}$ after end effector calibration to $15.3~\si{px}$ after all offsets calibration including camera calibration and to $15.6~\si{px}$ for fixed cameras calibration.

\begin{figure*}
    \centering
    \includegraphics[width = 0.85\textwidth]{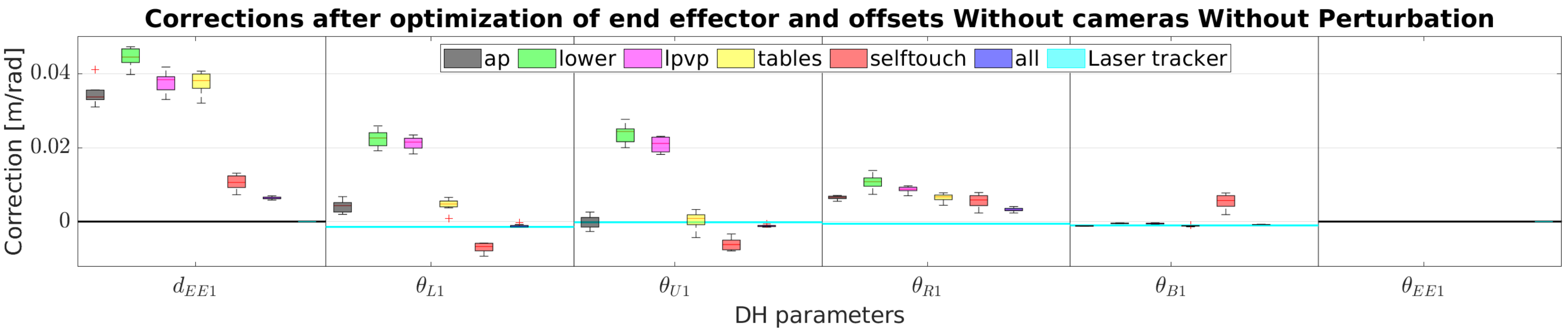}
        \includegraphics[width = 0.85\textwidth]{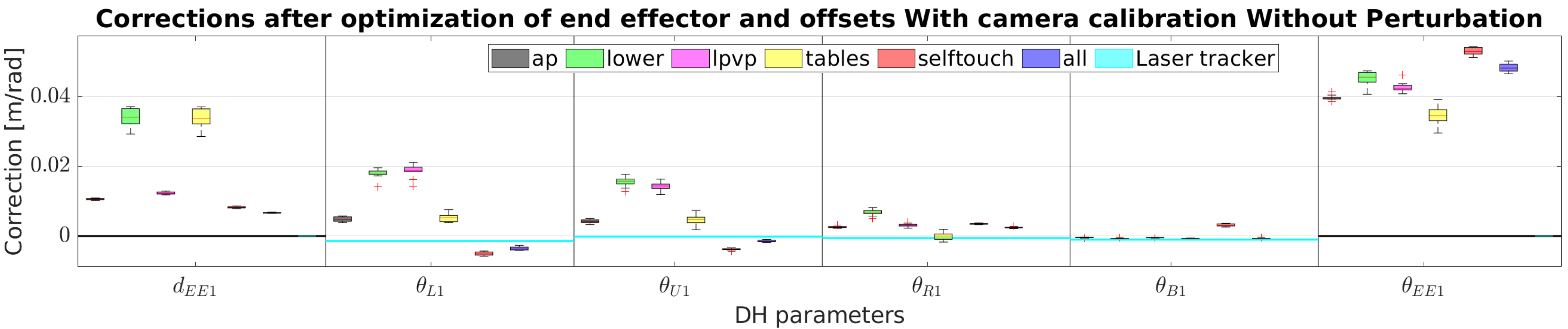}
            \includegraphics[width = 0.85\textwidth]{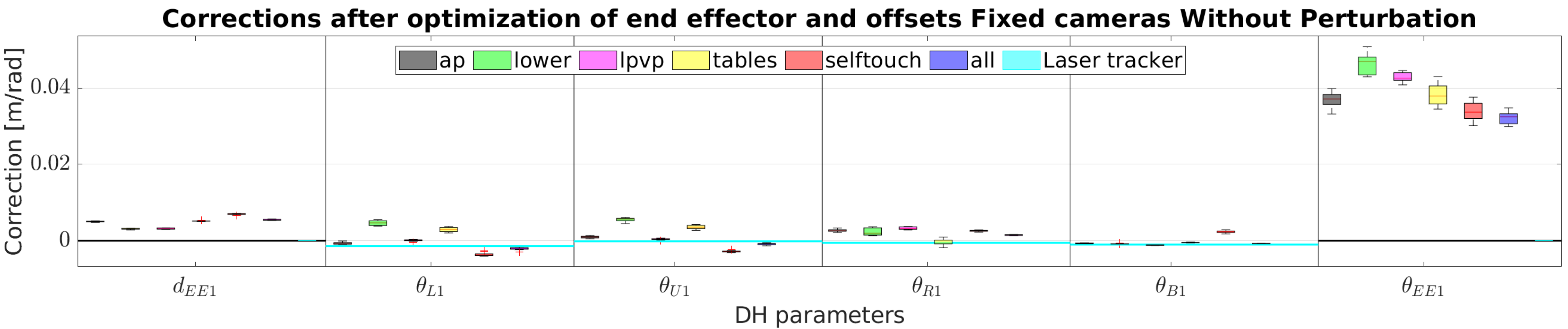}
    \caption{Corrections of offsets and end effector parameters ($d_{EE1}$, $o_{EE1}$) after calibrating end effector and offsets of right arm without cameras (using only planar constraints or self-touch), with precalibrated but fixed cameras and with precalibrated cameras which are calibrated during the experiment. Offset of the end effector cannot be calibrated without cameras and nominal values are used. Laser tracker calibration (ground truth) values in turquoise.}
    \label{fig:corr_offsets}
\end{figure*}

Corrections of the parameters to the nominal parameters are shown in Fig.~\ref{fig:corr_offsets}. Corrections of the parameters are the smallest (also having the smallest standard deviation) compared to the values achieved by laser tracker calibration for the variant \emph{all} with fixed precalibrated cameras, which means that we combined all of the chains (planar constraints, self-touch, self-observation) to optimize the offset parameters. The resulting correction for parameter $d_{EE1}$ is $5.46 \pm 0.14~\si{mm}$.
The corrections for parameters $\{o_{L1}$, $o_{U1}$, $o_{R1}$, $o_{B1}$, and $o_{EE1}\}$ are: $\{-2.10 \pm 0.38$, $-0.89 \pm 0.24$, $1.48 \pm 0.13$, $-0.76 \pm 0.06$, and $32.2 \pm 1.6\}$~\si{mrad}, respectively. The corresponding corrections from laser tracker calibration for parameters $\{o_{L1}$, $o_{U1}$, $o_{R1}$, and $o_{B1}\}$ are $\{-1.43$, $-0.18$, $-0.59$, and $-1.03\}$~\si{mrad}, respectively.

{}

\subsection{Comparison of nominal DH parameters and former calibration/calibration performed by laser tracker}
\label{sec:comparison_daily_calib}

\begin{figure}[!h]
\includegraphics[width = 1\linewidth]{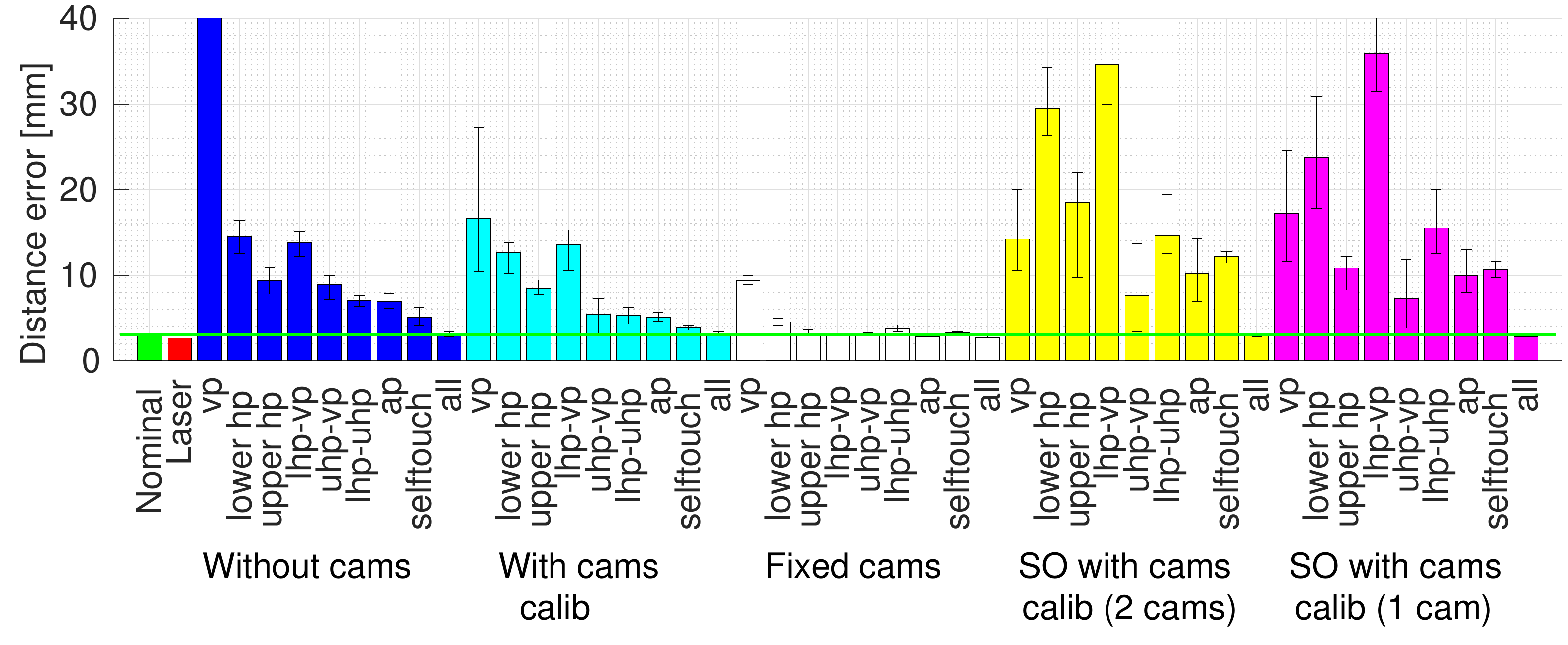}
\caption{Comparison of error on the testing dataset (\textit{$D^{tracker}$})-- dataset used for laser tracker reference calibration. Error computed towards laser tracker retroreflector data. $x$ axis -- individual datasets/methods used for finding robot kinematics parameters. We compare self-observation (\textit{Self-obs.}) with one/both precalibrated cameras (on $D^{whole}$) calibrating robot offsets based on data from individual datasets, calibration using plane constraints (horizontal plane (\textit{lower}/\textit{upper}), 2 horizontal planes (\textit{tables}), vertical plane (\textit{vp}) and all planes (\textit{ap})), self-touch calibration (\textit{self-touch}), and self-touch + planar constraints (\textit{all}). Comparison when no cameras are used (\textit{Without cams}), precalibrated (using self-observation on $D^{whole}$) but fixed cameras (\textit{Fixed cams.}), and precalibrated cameras (on $D^{whole}$) being part of the calibration (\textit{With cams calib}).}
\label{fig:finalComparison}
\end{figure}

{}

\begin{figure*}
    \centering
    \includegraphics[width = 0.85\textwidth]{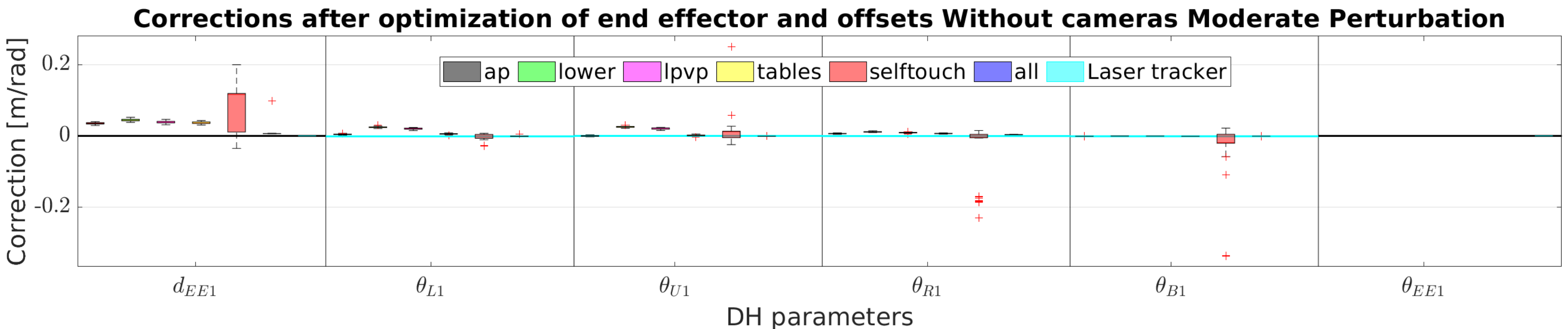}
        \includegraphics[width = 0.85\textwidth]{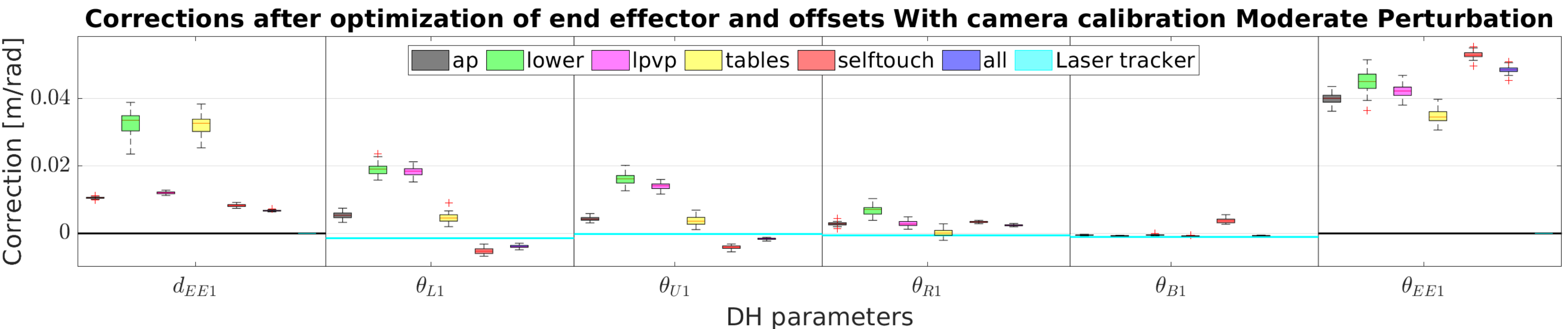}
            \includegraphics[width = 0.85\textwidth]{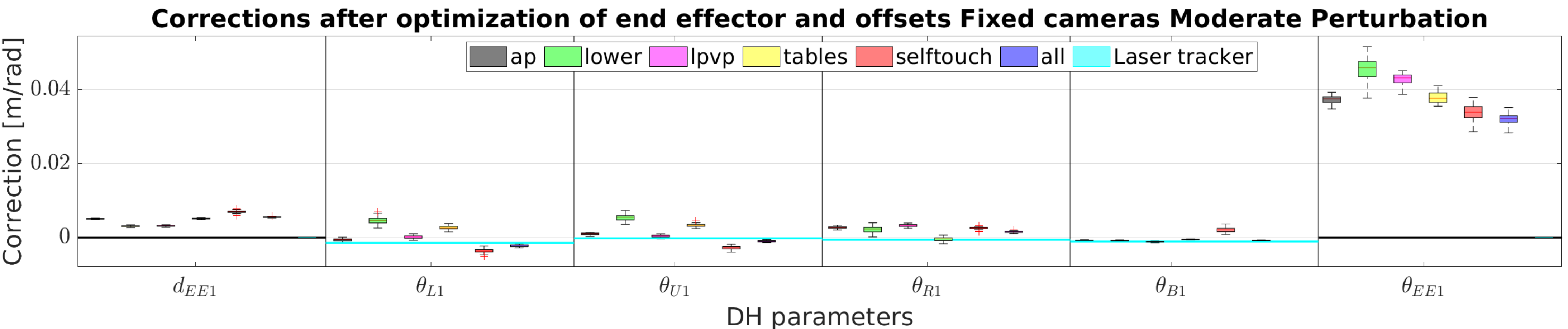}
    \caption{Corrections of offsets and end effector parameters ($d_{EE1}$, $o_{EE1}$) after calibrating end effector and offsets of right arm without cameras (using only planar constraints or self-touch), with precalibrated but fixed cameras and with precalibrated cameras which are calibrated during the experiment. Offset of the end effector cannot be calibrated without cameras and nominal values are used. Laser tracker (Leica) calibration (ground truth) values in turquoise. Initial perturbation p = 3.}
    \label{fig:corr_offsets_pert}
\end{figure*}{}

To evaluate the resulting offset parameters for the robot acquired through calibration using individual approaches, we conducted a comparison study on the laser tracker testing dataset $D^{tracker}$ (which covers the whole robot right arm workspace and not only the area where calibration using self-contained approaches was performed). To be able to perform such an evaluation, the transform to laser retroreflector, acquired from laser tracker calibration, was used for all compared calibration results. These results are shown in Fig.~\ref{fig:finalComparison}. 
Using the calibrated parameters by variant \emph{all}, we can achieve better RMSE than nominal parameters---both in the case without cameras or with fixed precalibrated cameras. The resulting RMSE is $3.06~\si{mm}$, $2.63~\si{mm}$, $3.04~\si{mm}$, $2.73~\si{mm}$  for nominal parameters, laser tracker, variant \emph{all} without cameras, and variant \emph{all} with precalibrated fixed cameras, respectively.
Adding the option of camera calibration increases the RMSE for the variant \emph{all} to $3.18~\si{mm}$ on this laser tracker testing dataset $D^{tracker}$ compared to the nominal parameters ($3.06~\si{mm}$). 

The lowest RMS errors (for some setups lower than RMSE with nominal parameters, which is $3.06~\si{mm}$) are achieved for different calibration approaches with fixed precalibrated cameras---the resulting RMSE are $9.3~\si{mm}$, $4.5~\si{mm}$, $3.8~\si{mm}$, $3.31~\si{mm}$, $2.84~\si{mm}$, and $2.73~\si{mm}$ for vertical plane (\emph{vp}), lower horizontal plane (\emph{hp}), both horizontal planes (\emph{tables}), self-contact (\emph{selftouch}), all planes (\emph{ap}), and variant \emph{all}, respectively. 

Pure self-observation (SO) highly depends on the quality of the training dataset. For the datasets from planar contact with one plane, we have only $250$ datapoints and the resulting error is $29.4~\si{mm}$ for 2 cameras and $23.7~\si{mm}$ for 1 camera. When more data points are added---all planes ($750$ datapoints) or self-contact ($566$ datapoints)---the RMSE drops to $10.2~\si{mm}$ and $12.15~\si{mm}$ for all planes and self-contact, respectively. When $D^{whole}$ is used ($1316$ datapoints), the RMSE drops below RMSE achieved for nominal parameters ($3.06~\si{mm}$): to $2.90~\si{mm}$ for 2 cameras setup and $2.82~\si{mm}$ for 1 camera setup (still higher than variant \emph{all} with fixed cameras -- $2.73~\si{mm}$).

\subsection{Calibration of robot joint offsets -- sensitivity to perturbation}
\label{sec:daily_calib_sensitivity_to_perturbation}

To evaluate the sensitivity of calibration approaches to perturbation, we run the experiment with initial perturbed parameters ($p = 3$) (see Section~\ref{section:perturbation} for details). The results can be seen in Fig.~\ref{fig:corr_offsets_pert}. For fixed precalibrated cameras we achieve very similar results as without perturbation with comparable low standard deviation.  For example, for the variant \emph{all} with fixed precalibrated cameras, the resulting correction for parameter $d_{EE1}$ is $5.46 \pm 0.14~\si{mm}$ without perturbation and $5.52 \pm 0.14~\si{mm}$with perturbation. The corrections for offset parameters $\{o_{L1}$, $o_{U1}$, $o_{R1}$, $o_{B1}$, and $o_{EE1}\}$ are: $\{-2.10 \pm 0.38$, $-0.89 \pm 0.24$, $1.48 \pm 0.13$, $-0.76 \pm 0.06$, $32.2 \pm 1.6\}$$10^{-3}~\si{rad}$ without perturbation, and $\{-2.24 \pm 0.28$, $-0.96 \pm 0.23$, $1.55 \pm 0.20$, $-0.75 \pm 0.07$, $32.1 \pm 1.3\}$$10^{-3}~\si{rad}$ with perturbation, respectively. 
Without cameras, the corrections have higher standard deviation, especially for the 'self-touch' calibration. For  all perturbation  levels,  the  performance  is  stable  (low  error
s.d.) in the case of fixed cameras.

\subsection{Calibration of all DH parameters}
\label{sec:calibration_all_dh}

\begin{figure}[!ht]
    \centering
    \includegraphics[width = 0.48\textwidth]{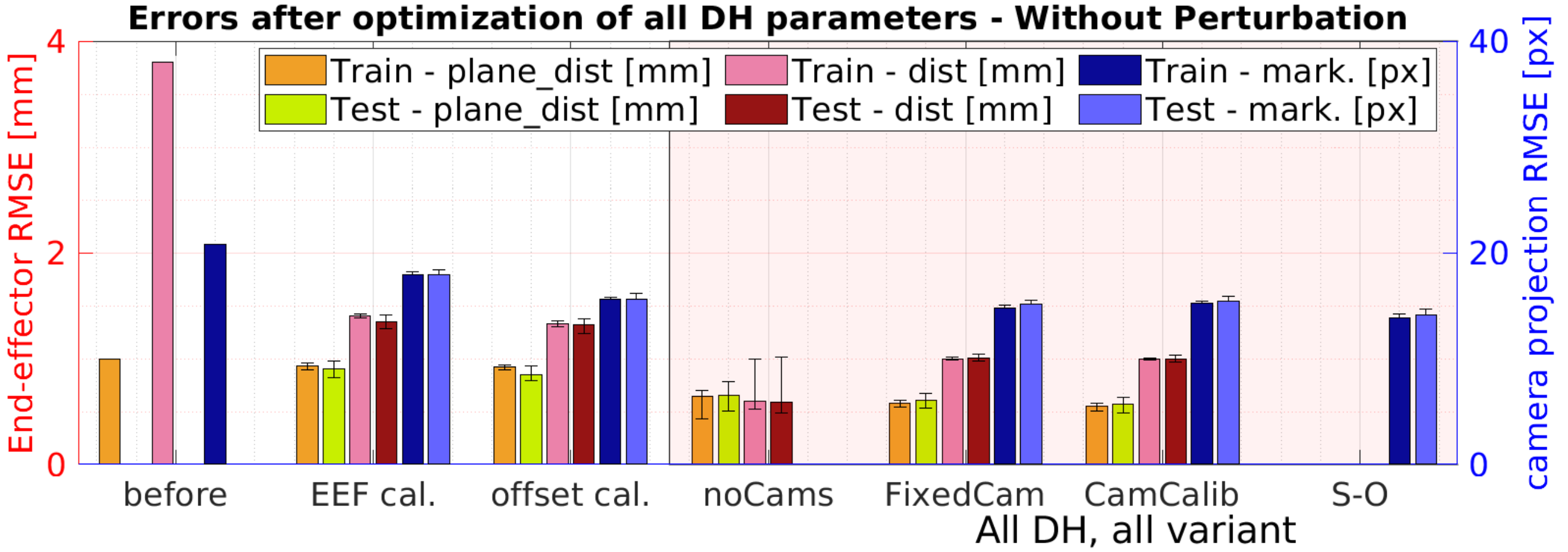}
    \caption{RMS errors -- right arm all DH parameters calibration. Distances in 3D (left y-axis, $mm$); camera reprojection (right y-axis, $px$). Results only for the variant \emph{all} using dataset $D^{whole}$. \textit{noCams} -- only planar constraints + self-touch; \textit{FixedCam} -- precalibrated but fixed cameras; \textit{CamCalib} -- precalibrated cameras with additional optimization, \textit{S-O} -- self-observation only. \textit{before} -- nominal parameters, \textit{EEF cal.} -- after end effector calibration, \textit{offset cal.} -- after calibration of robot offsets + end effector. For detailed legend see Fig.~\ref{fig:rmse_eef}.}
    \label{fig:rmse_allDH}
\end{figure}{}

\begin{figure*}[!ht]
    \centering
    \includegraphics[width = 0.9\textwidth]{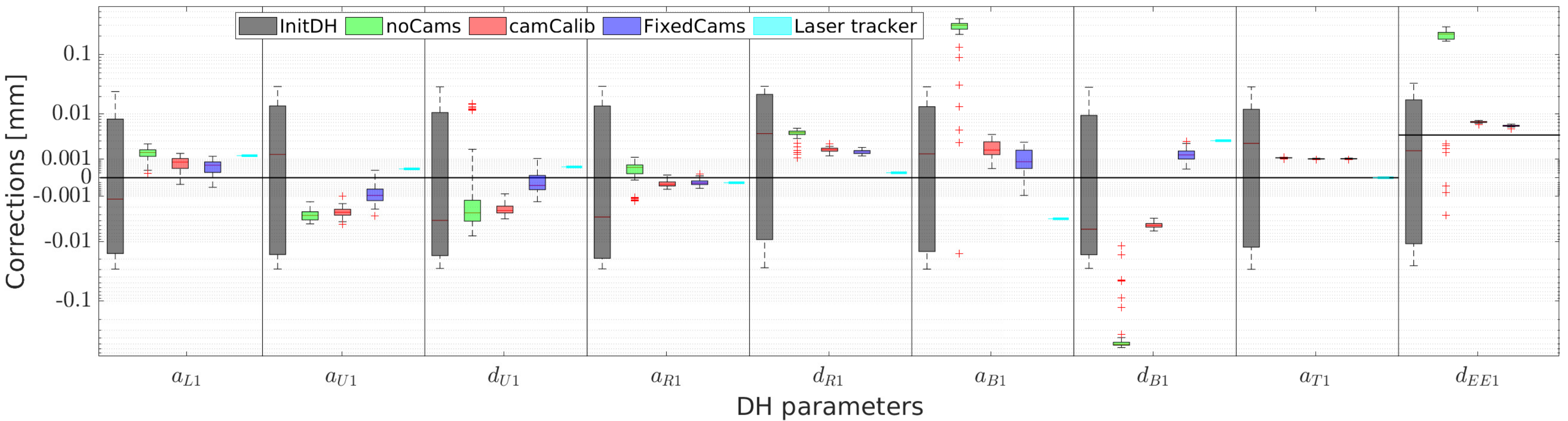}
        \includegraphics[width = 0.9\textwidth]{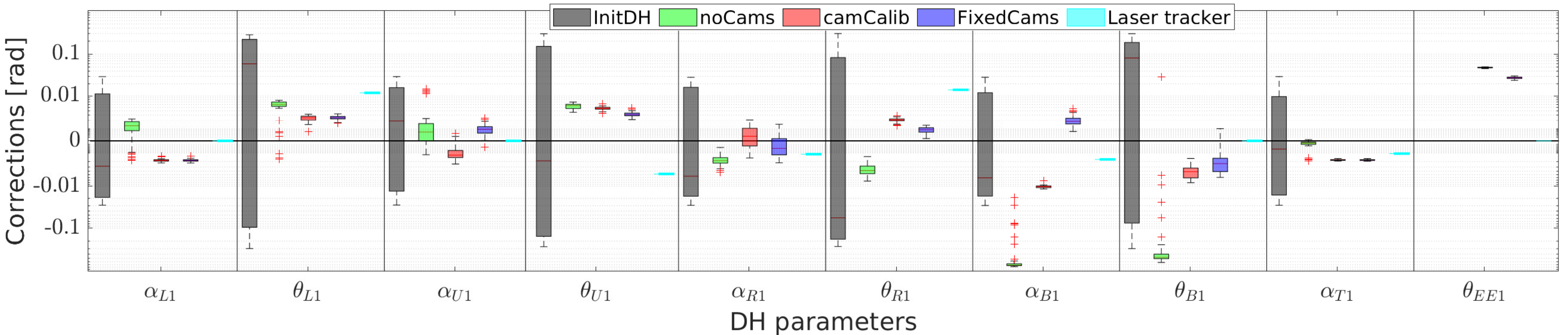}
    \caption{
    Corrections of all DH parameters for the right arm by the variant \emph{all} with initial perturbation ($p = 3$). Results without cameras (planar constraints + self-touch, \textit{noCams}), with precalibrated cameras which are calibrated during the experiment (\textit{camCalib}), and with precalibrated but fixed cameras (\textit{FixedCams}). Offset of the end effector cannot be calibrated without cameras -- nominal values are used. \textit{Laser tracker} calibration (reference) values in tyrquis. Results shown over 50 repetitions.}
    \label{fig:corr_allDH}
\end{figure*}{} 

In Figure~\ref{fig:rmse_allDH}, we show a comparison of RMSE after calibration of all DH parameters (see Table~\ref{tab:calib_params}) by the method \emph{all} for different setups (fixed cameras, calibrated cameras and no cameras). Results on both training and testing datasets are shown and compared to the case where nominal parameters are used (\emph{before}), EEF length and orientation calibration by the variant \emph{all} (EEF cal.) and right arm offset calibration by the variant \emph{all} (offset cal.). The self-contact distance, plane distance, and camera reprojection RMSE drops from $1.33~\si{mm}$, $0.92~\si{mm}$, and $15.6~\si{px}$, respectively, after offset calibration, to $1.00~\si{mm}$, $0.58~\si{mm}$, and $14.8~\si{px}$ after calibration of all DH parameters with calibrated cameras, and further to $1.00~\si{mm}$, $0.55~\si{mm}$, and $15.2~\si{px}$ after calibration with precalibrated fixed cameras. When cameras are not used (\textit{noCams}), self-contact distance further drops to $0.59~\si{mm}$ and planar distance to $0.64~\si{mm}$. In this case, the end effector offset is not calibrated and the camera reprojection error would be higher.

In Figure~\ref{fig:corr_allDH}, we show the resulting corrections of DH parameters (corrections are calculated as the difference from the nominal parameters) for the variant \emph{all} when perturbed DH parameters were used as initial value (perturbation factor $p=3$). We show initial values of perturbed DH parameters, the results of calibration by method \emph{all} in the case without cameras (using only contact information from planar constraints and self-touch), with precalibrated cameras which are calibrated during the experiment, and with precalibrated but fixed cameras. In all compared cases, we reach a lower dispersion of the final values and these values are very close to the industrial nominal DH parameters. The best results with the lowest standard deviation and closest to the industrial nominal values are consistently reached by the method when precalibrated fixed cameras are used. 
When no cameras are used, we can see that the corrections of DH parameters for variant \emph{all} are unrealistically far from the nominal parameters (e.g., for parameter $a_{B1}$ we get corrections around $0.2~\si{m}$). When also camera calibration is employed, the corresponding DH parameters corrections are reasonably close to nominal parameters and the calibration method is able to even for initially perturbed parameters (up to 10~\si{cm} difference from the nominal parameters) reach reasonable DH parameters values (all corrections are under $5~\si{mm}$).
We also show the results of the parameters from the laser tracker for comparison. End effector parameters cannot be calibrated by laser tracker thanks to its placement (see Fig.~\ref{fig:ee}). Therefore, the value is not changing from the initial parameter. For the end effector length, we visualise as the nominal parameter the correction to the initial end effector length value achieved by the former end effector calibration (Sec.~\ref{sec:ee_calibration}).

\begin{figure}[!ht]
    \centering
    \includegraphics[width = 0.5\textwidth]{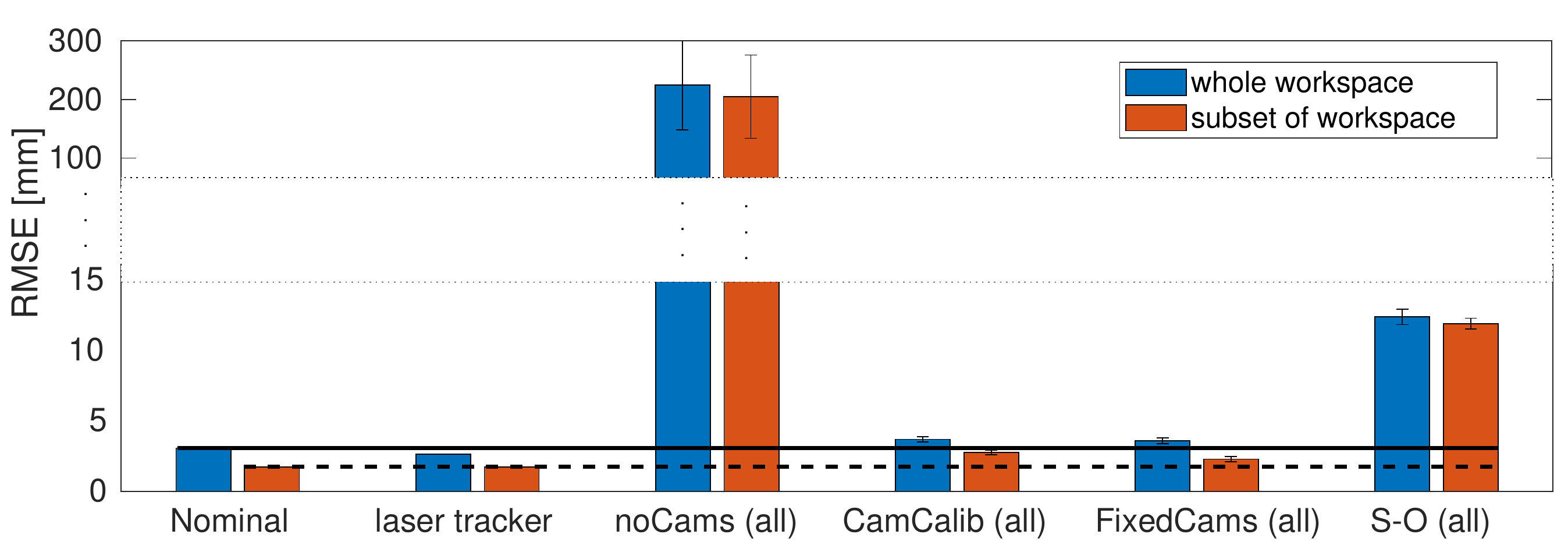}
    \caption{Comparison of RMSE for all DH calibration on the whole laser tracker dataset (\textit{$D^{tracker}$}). Results are shown for the whole laser tracker dataset (red) and for the subset of the robot workspace which corresponds to the area where other self-contact methods were trained (blue). Results for the calibration by the variant \emph{all} without cameras (\textit{noCams}), with fixed precalibrated cameras (\textit{FixedCams}) and cameras being calibrated (\textit{CamCalib}) is compared to the self-observation approach (\textit{S-O}) trained on the dataset $D^{whole}$, \textit{Nominal} parameters and the calibration by \textit{laser tracker}.}
    \label{fig:allDHRMSE_laserTracker}
\end{figure}
In Figure~\ref{fig:allDHRMSE_laserTracker}, we show the resulting RMSE  of our calibration contact-based approach on the testing laser tracker dataset $D^{tracker}$. The results for self-contact approach with fixed precalibrated cameras and with cameras calibrated as a part of the calibration approach are compared to self-observation approach where only camera information is used. All approaches are trained on the same $D^{whole}$ dataset. Comparison to the RMSE when nominal parameters and when DH parameters from laser tracker calibration are used, is provided. We show results for the whole laser tracker dataset and for the subset of the dataset which corresponds to the area where self-contact and self-observation approaches were calibrated. On the whole dataset, RMSE is $\{3.06 \pm 0$, $2.63 \pm 0$, $224 \pm 77$, $3.68 \pm 0.20$, $3.58 \pm 0.21$, and $12.36\pm0.55\}~\si{mm}$ for nominal parameters calibration (\textit{Nominal}), \textit{laser tracker}, self-contact calibration without cameras  (\textit{noCams (all)}), self-contact with cameras being part of calibration  (\textit{CamCalib (all)}), self-contact calibration with fixed precalibrated cameras   (\textit{FixedCams (all)}), and self-observation approach   (\textit{S-O (all)}), respectively. On the subset of the dataset, RMSE is $\{1.73 \pm 0$, $1.71 \pm 0$, $205 \pm 71$, $2.75 \pm 0.17$, $2.28 \pm 0.17$, and $11.88\pm0.38\}~\si{mm}$ for nominal parameters calibration (\textit{Nominal}), \textit{laser tracker}, self-contact calibration without cameras  (\textit{noCams (all)}), self-contact with cameras being part of calibration  (\textit{CamCalib (all)}), self-contact calibration with fixed precalibrated cameras (\textit{FixedCams (all)}), and self-observation approach   (\textit{S-O (all)}), respectively. We can see that including self-contact information as a part of calibration provides significantly better results than using pure self-observation approach.

\subsection{Observability analysis of individual approaches}
\label{sec:observability}

\begin{figure}[h]
    \centering
    \includegraphics[width = 0.5\textwidth]{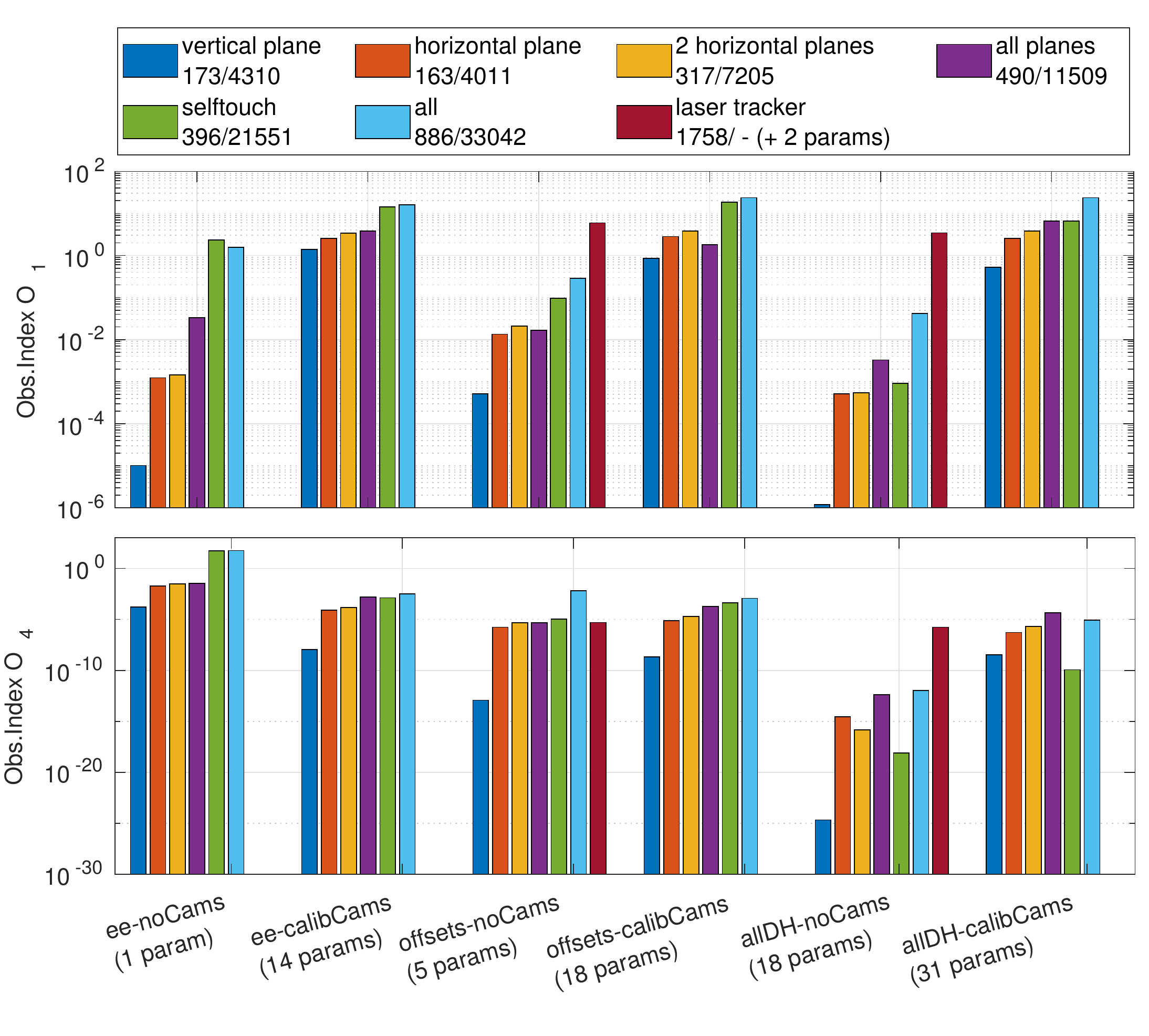}
    \caption{Observability indices $O_1$ (top) and $O_4$ (bottom). Groups on the x-axis correspond to parameters subject to calibration (\textit{ee} -- end effector length; \textit{offsets} -- $o_i$ of individual links; \textit{allDH} -- all DH parameters; \textit{noCams} / \textit{calibCams} -- without / with the ``self-observation chains''; Table~\ref{tab:calib_params} for details). Color coding of individual bars marks the calibration approach and dataset (no. data points without cameras / with cameras).}
    \label{fig:observability}
\end{figure}

We evaluated the observability indices $O_1$ and $O_4$ for the different combinations of calibration approaches and parameters subject to calibration. An overview is shown in Fig.~\ref{fig:observability} with $O_1$ on top and $O_4$ at the bottom. Please refer to  Section~\ref{section:multichain_observability} for details about how the indices are calculated. The bar groups on the x-axis correspond to the parameters subject to calibration and the number of parameters (columns of the identification Jacobian matrix \textbf{J}). Calibration using the laser tracker adds two additional parameters pertaining to the retroreflector placement (see Table~\ref{tab:calib_params}). Color coding of individual bars marks the calibration approach with the no. data points in a given dataset without cameras / with cameras (rows of \textbf{J}). Due to the variable size of \textbf{J}, comparisons are to be made with caution.
As expected, for the cases without cameras (\textit{noCams}), there is a trend that with the number of calibrated parameters increasing (from \textit{ee-noCams}, over \textit{offsets-noCams}, to \textit{allDH-noCams}), the observability indices are lower.
When self-observation as a calibration method and corresponding camera extrinsic parameters are added (\textit{calibCams}), observability indices $O_1$ in general significantly increase.
Observability index $O_4$ increases after adding camera chains to calibration for the \textit{allDH} calibration case, but drops for end effector length calibration---adding new parameters to be calibrated (columns of \textbf{J}) outweighs the effect of additional data from self-observation (rows of \textbf{J}). 
Comparing situations where the dataset and hence \textbf{J} has a similar size, we see that \textit{horizontal plane} was more effective than \textit{vertical plane} in our setup and that \textit{selftouch} slightly outperformed combinations of all planar constraints (\textit{all planes}). This is largely consistent with the RMSE errors after calibration (Fig.~\ref{fig:finalComparison}). Finally, for calibrating all DH parameters, the inclusion of cameras seems necessary.

Note that the observability analysis is also affected by measurement noise that may increase the effective rank of the Jacobian matrix. We assume that the measurement noise in this case is sufficiently small (experimental verification is presented at \cite{dataset_our}).

\section{Conclusion}
\label{section:conclusion}

 Using a dual-arm industrial robot with force/torque sensing and cameras, we presented a thorough experimental comparison of geometric kinematic calibration using ``self-contained approaches'' using sensors on the robot--- self-contact, planar constraints, and self-observation---with calibration using an external laser tracker. The main findings are summarized below.
 
 First, we studied estimation of the kinematic parameters of a new tool---a custom end effector (Section \ref{sec:ee_calibration}). To calibrate the tool length, self-contact alone proved effective; planar constraints or self-observation alone did not perform as well in isolation and improved only in synergy with one of the other approaches/datasets. Testing RMS errors of approximately 1 to 1.5 mm on the part of the workspace where calibration was performed were achieved. For orientation of the tool, the addition of cameras (self-observation) was needed. Combining different methods/kinematic chains proved effective, supported also by the observability analysis (Section~\ref{sec:observability} and Fig.~\ref{fig:observability}).
 
 Second, we analyzed the performance of ``daily calibration'': the joint angle offsets of a complete robot arm were added to the tool calibration (Section \ref{sec:calibration_offsets}). While the end effector calibration is responsible for most of the error in the workspace, further improvement from the offset calibration is achieved with all methods. To assess whether the calibration is effective only locally---on the particular dataset---or whether better kinematic parameters of the robot were learned, a comparison on an independent dataset covering the whole workspace (laser tracker calibration) was performed (Section~\ref{sec:comparison_daily_calib} and Fig.~\ref{fig:finalComparison}). Corrections to robot parameter values were also analyzed. Importantly, all self-contained approaches except for the planar constraints in isolation were able to achieve good results, sometimes outperforming the robot nominal parameters. Self-contact alone proved the most effective; the best results were achieved using a combination of datasets and methods. Self-contact was further improved by the addition of planar constraints. Self-observation alone was sensitive to the size of the dataset but was effective in combination with contact-based methods; the best results were achieved using sequential calibration: camera extrinsics were calibrated first, followed by robot arm kinematics and tool calibration. Additionally, similar results were achieved when perturbing the initial parameter estimates (Section~\ref{sec:daily_calib_sensitivity_to_perturbation}).
 
 Third, we performed a complete geometric kinematic calibration of one robot arm plus the tool (all DH parameters) using different methods (Section \ref{sec:calibration_all_dh}). Good results are achieved on the individual datasets (Fig.~\ref{fig:rmse_allDH}), but further analysis  reveals that the calibration suffered from overfitting to the particular dataset / part of the robot workspace. On an independent dataset, the performance of nominal parameters was not matched (Fig.~\ref{fig:allDHRMSE_laserTracker}). We also found that when all DH parameters are subject to calibration, the need for the synergy of different approaches increases, as testified by the unrealistic parameter corrections with the contact-only (noCams) approaches in Fig.~\ref{fig:corr_allDH} and the observability analysis (Fig.~\ref{fig:observability}, \emph{allDHnoCams}) or the self-observation only approach in Fig.~\ref{fig:allDHRMSE_laserTracker}. Like previously, the sequential calibration---first cameras, then robot kinematics---gives the best results. Although the performance of nominal or laser tracker calibration parameters on the whole workspace could not be matched (Fig.~\ref{fig:allDHRMSE_laserTracker}), the performance of the combination of self-contained approaches---also in the case of initial parameter perturbation---is reasonable (less than 4 mm error) and for a less accurate platform like a service robot may suffice.  

Fourth, the comprehensive dataset collected is made publicly available \cite{dataset_our} and can be used for additional analyses. This constitutes an additional contribution of this work.

\section{Discussion and future work}
\label{section:discussion}

Geometric kinematic calibration of industrial robots is usually performed using external metrology---measurement arms (e.g., \cite{ginani2011theoretical}) or contactless measurement systems like laser trackers \cite{ha2008kinematic,nguyen2013new,Newman2000,nubiola2013absolute}. Newman et al.~\cite{Newman2000} calibrated a Motoman P-8 robot using an SMX laser tracker, improving the RMS error from 3.595 mm to 2.524 mm. Specifically related to the setup used in this work, our platform was previously calibrated using two different methods: (1) Redundant parallel Calibration and measuring Machine (RedCaM) by  Bene{\v{s}} et al.~\cite{benevs2007experiments}, Volech et al.~\cite{volech2013concepts}, and (2) Leica laser tracker. Petr{\'\i}k and Smutn{\'y} \cite{petrik_2014_comparison} reviewed the precision of these methods using a linear probe sensor. Based on a dataset of 43 different poses with touching end effectors, they calculated the mean error as 0.67 (range 2.92) mm on CAD model, 0.54 (range 2.55) mm on Leica based calibration and 2.45 (range 9.92) mm on RedCaM based calibration. Other approaches---see \cite{nubiola2013absolute} and additional references cited therein---usually achieve sub-millimeter accuracy. It was not our goal to directly compete with these works; instead, our aim was to assess the potential of automatic self-contained kinematic calibration: using sensors on the robot and avoiding the need for external metrology. We could demonstrate that such self-contained approaches---even if the initial robot parameters are perturbed---can yield less than 4 mm position errors over the robot workspace. The accuracy increases when a combination or synergy of these approaches (e.g., self-contact and self-observation) is exploited. We chose our platform, an industrial robot, out of convenience and to have a stable enough plant that can provide stationary results. However, we see the main application area in collaborative and service robotics. These platforms are typically more lightweight, flexible, and less precise and they may be often redeployed, their kinematic configuration changed etc. At the same time, they often come with a rich set of inexpensive but increasingly powerful sensors. Cameras and means to detect physical contact are becoming common. All these factors pave the way for automatic multisensorial self-contained calibration as demonstrated here. 

We provide not only a thorough experimental investigation, but also a conceptual contribution in that we develop methods how to combine several of the methods into a single cost function. Further, unlike  Birbach et al.~\cite{Birbach2015} who claim that simultaneous calibration using all the available sensors is advantageous, we consistently found sequential calibration to perform best: camera calibration followed by robot kinematics. Conversely, calibrating robot kinematics together with camera extrinsics simultaneously was not as successful. 
Furthermore, if camera extrinsics were not precalibrated and inaccurate nominal parameters were used, the optimization often converged to physically impossible local minima.  
One known limitation of closed-loop calibration approaches---those relying on physical constraints on the end effector position or orientation---is that the set of poses is limited, which may affect the identifiability of parameters \cite{Hollerbach2016}. This holds pretty much for all self-contained calibration---self-contact or self-observation---as these also naturally constrain the set of robot configurations available. This has also been the case in this study. Better coverage of individual joint ranges within the approaches used here---in particular for the planar constraints---would further improve the results and yield better generalization to other parts of the workspace. Combining different methods / kinematic chains significantly mitigates this problem.

It should also be noted that although we used specially designed end effectors that were developed to combine self-contact (acting as a sphere) and self-observation (with flat tiles to host fiducial markers), the methods developed here have wider applicability. Contact can occur at any part of the robot provided that the link can be represented in the kinematic model; fiducial markers can be placed on any robot part or avoided altogether by tracking parts of the robot directly (e.g., \cite{Birbach2015,Fanello2014,Vicente2016}).

There are several directions for future work. First, when combining several calibration approaches into a single cost function (Sec.~\ref{section:multichain_calibration}), the errors obtained from the different components could be scaled by coefficients that are inversely proportional to their uncertainty. We attempted to acquire such coefficients using repeatability measurements, but failed to obtain estimates that would reflect the true uncertainty associated with different approaches and possibly individual data points. Additional measurements of the uncertainty of individual components (beyond Sec.~\ref{section:accuracy_components}) and their propagation would be required. Thus, all components were weighted equally in this work, but this can be changed in the future. 

Second, the methods presented here can be extended to exploit the existing sensors differently or to incorporate additional sensory modalities. For example, the two cameras in our setup were not used explicitly as a stereo head. Instead of reprojecting the end effector into the 2 image frames, one could also project the observed position of the end effector in image coordinates of both eyes (pixel $(u,v)$) to 3D space ($X^{eye}$) (similar to \cite{Fanello2014,Hirschmuller2008}). The self-contact approach would yield more than a 1-dimensional constraint in case the contact position (and hence 3 components of the pose) could be measured such as when using an artificial electronic skin \cite{Roncone_ICRA_2014, QiangLi2015, Stepanova2019}. Inertial sensors---in the robot head (\cite{Birbach2015}) or distributed on the robot body \cite{Guedelha2016,Mittendorfer2012}---could be also added. Finally, one could also calibrate both manipulators simultaneously. 

Third, for online recalibration to be performed repeatedly, the number of poses / data points needed can be reduced by employing intelligent pose selection (e.g., \cite{li2011,daney2005,zhou2014selecting,wang2018universal}). 

Fourth, the standard calibration method using non-linear least squares optimization (Levenberg-Marquardt algorithm) can be compared with filtering approaches \cite{Vicente2016,zenha2018incremental} or with methods that pose fewer assumptions on the initial model available (e.g., \cite{Lanillos2018}).

\section*{Acknowledgements}
We would like to thank Martin Hoffmann for designing and fabricating several versions of the custom end effectors, with assistance from Tom{\'a}{\v s} B{\'a}{\v c}a for 3D printing. We are also indebted to CIIRC CTU for support in using the robot and the Leica absolute tracker (Libor W{\'a}gner, Vladim{\'ir} Smutn{\'y}, and V{\'a}clav Hlav{\'a}{\v c}). Tom{\'a}{\v s} Svoboda provided general support, consultation, and pointer to the article \cite{Arun1987}. 

\section*{Funding}
This work was supported by the Czech Science Foundation (GA {\v C}R), project EXPRO (no. 20-24186X);
T.P. was supported by the European Regional Development Fund under project Robotics for Industry 4.0 (no.~CZ.02.1.01/0.0/0.0/15$\_$003/0000470) and EU H2020 SPRING Project (no.~871245).

\bibliographystyle{elsarticle-num} 
\bibliography{MotomanSelfCalib}  

\end{document}